\documentclass{article}
\usepackage{amsmath} 
\usepackage{xcolor}
\definecolor{darkgreen}{RGB}{0,100,0} 
\usepackage{graphicx}
\usepackage{lmodern}
\usepackage{PRIMEarxiv}
\usepackage{multirow}
\usepackage{mdframed}
\usepackage{todonotes}
\usepackage{enumitem}
\usepackage[table]{xcolor}
\usepackage{amsmath}  \usepackage[utf8]{inputenc}
\usepackage[T1]{fontenc}
\usepackage{diagbox}
\usepackage{xcolor}
\usepackage[most]{tcolorbox}
\usepackage{lmodern}
\usepackage{lscape}  
\usepackage{pdflscape}  
\usepackage{lscape}
\usepackage{caption}
\usepackage[utf8]{inputenc} 
\usepackage{siunitx} 
\usepackage[T1]{fontenc}    
\usepackage{hyperref}       
\usepackage{url}            
\usepackage{booktabs}       
\usepackage{amsfonts}       
\usepackage{nicefrac}       
\usepackage{microtype}      
\usepackage{lipsum}
\usepackage{fancyhdr}       
\usepackage{graphicx}       
\graphicspath{{media/}}     
\usepackage{cite} 
\usepackage{tikz}
\usetikzlibrary{positioning,fit,backgrounds}
\usepackage{subcaption}
\usepackage{graphicx}
\usepackage{lscape}
\usepackage{amssymb}
\usepackage[export]{adjustbox}
\usepackage{soul}
\usepackage{newunicodechar}
\usepackage{rotating}
\usepackage{lscape}
\usepackage{graphicx}
\usepackage{caption}
\usepackage{amssymb} 
\newunicodechar{✅}{\checkmark}

\usepackage{tikz}
\usetikzlibrary{positioning,fit,backgrounds,calc}
\tikzset{
  outer/.style={
    rectangle, rounded corners=12pt,
    draw=black!70, fill=gray!20, line width=0.8pt,   font=\small  },
  panel/.style={
    rectangle, rounded corners=12pt,
    draw=black!75, line width=0.8pt,
    inner xsep=10pt, inner ysep=10pt,
    align=left,   font=\small 
  },
  textbook/.style={panel, fill=blue!15!white,align=left},
  llm/.style={panel, fill=orange!10!white},
  tab/.style={
    rectangle, rounded corners=6pt,
    draw=black!70, line width=0.9pt,
    inner xsep=10pt, inner ysep=3pt,
    font=\small           
  },
  title/.style={align=left},     
  score/.style={
    rectangle, rounded corners=8pt,
    draw=black!60, fill=pink!35!white,
    inner xsep=12pt, inner ysep=4pt    }
}

\title{ PsychiatryBench: A Multi-Task Benchmark for LLMs in Psychiatry}
\author{
Aya E. Fouda$^\dagger$, Abdelrahamn A. Hassan$^\dagger$, Radwa J. Hanafy$^{\ddagger,\dagger}$ and  Mohammed E. Fouda$^{\dagger,*}$ \\ \\
$^\dagger$Compumacy for Artificial Intelligence solutions, Cairo, Egypt.\\
$^\ddagger$Department of Behavioural Health- Saint Elizabeths Hospital, Washington, DC, 20032, USA.\\
*Corresponding author. E-mail: fouda@compumacy.com
}

\begin{document}
\maketitle

\begin{abstract}
Large language models (LLMs) offer significant potential in enhancing psychiatric practice, from improving diagnostic accuracy to streamlining clinical documentation and therapeutic support. However, existing evaluation resources heavily rely on small clinical interview corpora, social media posts, or synthetic dialogues, which limits their clinical validity and fails to capture the full complexity of diagnostic reasoning. In this work, we introduce PsychiatryBench, a rigorously curated benchmark grounded exclusively in authoritative, expert-validated psychiatric textbooks and casebooks. PsychiatryBench comprises  eleven distinct question-answering tasks ranging from diagnostic reasoning and treatment planning to longitudinal follow-up, management planning, clinical approach, sequential case analysis, and multiple-choice/extended matching formats totaling 5,188 expert-annotated items. {We evaluate a diverse set of frontier LLMs (including Google Gemini, DeepSeek, Sonnet 4.5, and GPT 5) alongside leading open-source medical models such as MedGemma using both conventional metrics and an "LLM-as-judge" similarity scoring framework. Our results reveal substantial gaps in clinical consistency and safety, particularly in multi-turn follow-up and management tasks, underscoring the need for specialized model tuning and more robust evaluation paradigms. PsychiatryBench offers a modular, extensible platform for benchmarking and improving LLM performance in mental health applications. 
}
\end{abstract}


\section{Introduction}
\label{sec:Introduction}
{
Mental disorders impose a staggering global burden, accounting for 5.1\% of worldwide disease burden in 2019, affecting over 280 million people with depression, and claiming 703,000 lives to suicide \cite{WHO2019}. The economic toll is projected to reach \$14 trillion in the United States alone between 2024 and 2040, encompassing direct medical costs, emergency interventions, productivity losses, and premature mortality \cite{Smith2023}. These figures demand urgent action: scalable technologies that can improve diagnostic precision, expand care access, and accelerate treatment delivery are not merely aspirational; they are critical infrastructure gaps in modern healthcare systems.

Large Language Models (LLMs), a subset of artificial intelligence (AI), have emerged as a potentially transformative technology in healthcare. The integration of LLMs into psychiatric practice represents a significant technological development with substantive implications for clinical practice. Recent studies have demonstrated that models can perform meaningful tasks in mental health contexts, including detecting depression and suicide risk from text, providing supportive information through conversational agents, and generating clinical documentation that psychiatrists prefer in blind comparisons \cite{shah2025tnevalrubricevaluationprotocols, ji2021mentalbert}. Some investigations have even suggested that LLMs can surpass human physicians in expressing empathy in written clinical communication \cite{Hua2025LLMsMentalHealth}. These capabilities suggest that LLMs may offer pathways to democratize mental health assessment and support, potentially reaching populations with limited access to specialized psychiatric care.

However, the transformative potential of LLMs in psychiatry is fundamentally constrained by substantial safety and reliability concerns. A primary concern is the well-documented propensity of LLMs to generate inaccurate or nonsensical responses, a phenomenon known as ``hallucinations,'' which raises profound safety issues in psychiatry where diagnostic errors can result in serious patient harm \cite{pandey2024harnessinglargelanguagemodels}. The landscape of existing evaluation frameworks reveals critical limitations. Early, influential resources provided clinically relevant data but were limited in scale and faced significant challenges in generalizability \cite{Guo_2024}. To achieve broader evaluation scope, subsequent benchmarks increasingly relied on unverified social media data, which tends to oversimplify complex psychiatric presentations and risks conflating self-identified symptoms with formal clinical diagnoses rather than representing the diagnostic reasoning required in professional psychiatric practice \cite{low2020natural, hanafi2024comprehensiveevaluationlargelanguage}. More recent efforts have developed conversational benchmarks addressing clinical dialogue, but these frequently depend on synthetic data or employ other LLMs as evaluators, introducing methodological concerns and potential biases that compromise validity \cite{xu2025mentalchat16kbenchmarkdatasetconversational, li2024llmsasjudgescomprehensivesurveyllmbased}.

Despite these advances, a critical evaluation gap persists: existing frameworks rarely assess models on tasks that genuinely reflect the depth, complexity, and rigor of real-world psychiatric clinical reasoning and decision-making. This gap is particularly consequential because psychiatry, as a medical specialty, demands the integration of diagnostic criteria, systematic differential diagnosis, treatment algorithm selection, and dynamic case management competencies that go far beyond pattern recognition or conversational fluency.

The authoritative knowledge foundations of psychiatric practice represented in curated clinical casebooks and diagnostic resources such as the DSM-5-TR Clinical Cases \cite{APA2013} and Stahl's Essential Psychopharmacology \cite{stahl2021essential} represent decades of expert knowledge curation and validation. These resources document the complexities of differential diagnosis, the nuances of treatment planning, and the sophisticated clinical judgment required in real psychiatric practice. These foundational sources establish the gold standard against which an LLM's clinical reasoning, diagnostic acumen, safety awareness, and appropriateness for psychiatric applications must ultimately be measured\cite{aljohani2025comprehensivesurveytrustworthinesslarge}.

To address this critical evaluation gap, we introduce PsychiatryBench, a novel evaluation framework meticulously constructed from foundational psychiatric textbooks and expert-validated clinical cases, deliberately prioritizing clinical accuracy, safety, and alignment with established medical practice. PsychiatryBench is designed with a single, exclusive purpose: to serve as a rigorous, standardized tool for comparative evaluation of {\color{darkgreen}LLM's} capabilities in psychiatry. Its core aim is to provide a benchmark that assesses whether LLMs can approximate expert-level psychiatric reasoning and clinical decision-making as demonstrated by experienced psychiatrists.

We define high-stakes psychiatric applications as clinical contexts where AI errors could result in serious patient harm, including: misdiagnosis leading to inappropriate or delayed treatment, failure to recognize suicide risk, incorrect medication recommendations causing adverse effects, or inappropriate crisis management. Such applications demand rigorous evaluation standards that assess not only general language understanding but also clinical accuracy, and appropriate acknowledgment of model uncertainty and limitations.

\textbf{Scope:} PsychiatryBench is intentionally focused on adult psychiatry as a medical specialty within the broader mental health field. The benchmark targets psychiatry-specific clinical reasoning and decision-making, excluding general mental health interventions, psychotherapy, counseling techniques, and broader behavioral health domains. Child and adolescent psychiatry was deliberately excluded, as it represents a distinct subspecialty with unique developmental, diagnostic, and therapeutic considerations that would necessitate a dedicated evaluation framework. The inclusion of geriatric psychiatry reflects the clinical reality that geriatric cases are commonly encountered by general adult psychiatrists. While source cases include references to medical mimics of psychiatric conditions (e.g., thyroid disease, CNS pathology, delirium), these are not systematically or comprehensively represented across all tasks. The benchmark does not systematically assess conversational coherence (turn-by-turn dialogue management) or narrative biopsychosocial formulation (holistic case conceptualization integrating biological, psychological, and social factors), as our focus is on knowledge application and clinical decision-making rather than interactive communication or narrative synthesis.

Throughout this work, the term psychiatric reasoning is used as an umbrella term encompassing six specific, complementary dimensions of clinical competence, each targeted through dedicated task types:

\textbf{Diagnostic reasoning:} The cognitive process of synthesizing patient presentation, history, and mental status examination findings to generate and systematically differentiate among possible psychiatric disorders (assessed through Diagnosis and Classification tasks)
    
\textbf{Treatment decision-making:} Clinical judgment in selecting appropriate therapeutic interventions based on diagnosis, patient characteristics, individual vulnerabilities, and evidence-based guidelines (assessed through Treatment and Treatment Follow-Up tasks)
    
\textbf{Management planning:} Comprehensive care coordination integrating diagnostic formulation, risk assessment, therapeutic interventions, psychoeducation, and coordinated follow-up strategies (assessed through Management Plan tasks)
    
\textbf{Clinical approach:} The structured reasoning process for systematically evaluating clinical presentations, generating diagnostic hypotheses, and prioritizing clinical investigations (assessed through Clinical Approach tasks)
    
\textbf{Foundational knowledge:} Factual recall and application of psychiatric concepts, diagnostic criteria, treatment mechanisms, and clinical principles (assessed through Mental QA, MCQ, and EMI tasks)
    
\textbf{Sequential reasoning:} The ability to track evolving clinical information across multiple time points and adapt diagnostic and treatment recommendations accordingly (assessed through Sequential QA tasks).

Our contribution comprises three interconnected elements:

\begin{itemize}
    \item We construct a novel, clinically rigorous dataset curated from authoritative psychiatric casebooks, diagnostic manuals, and specialized clinical guides, ensuring that evaluation is grounded in validated, expert knowledge rather than unverified or synthetically generated data.
    
    \item We design eleven distinct and complex task types that move beyond simplified assessment to rigorously evaluate LLMs' readiness for the complex, intellectually demanding, and ethically high-stakes reality of contemporary psychiatric clinical practice.
    
    \item We provide a standardized evaluation resource specifically designed for comparative LLM assessment in psychiatry. While insights from this benchmark may secondarily inform discussions about clinical decision support systems or educational applications, PsychiatryBench's primary and exclusive purpose is rigorous model evaluation and comparison. We acknowledge that application to regulatory or policy decisions would necessitate substantial expansion to address cultural, developmental, legal, and systems-level domains, which are explicitly outside the scope of this work.
\end{itemize}

The remainder of this paper is structured as follows: Section~\ref{sec:related_work} provides a detailed review of LLM applications in psychiatry and the landscape of existing psychiatric evaluation datasets. Section~\ref{sec:The_PsychiatryBench_Benchmark} describes the design, sourcing, and annotation methodology for our novel psychiatric evaluation dataset. Section~\ref{sec:Experimental_and_Evaluation_Methodology} outlines the experimental design and evaluation methodology, including model implementation details and evaluation protocols. Section~\ref{sec:Results_Discussion} presents and interprets our empirical results. Section~\ref{sec:Limitations_and_Future_work} discusses study limitations and directions for future work. Section~\ref{sec:Conclusion} concludes by summarizing our key contributions and implications. Ethical, Copyright, and Fair Use Statements are presented in section \ref{copyright}.
}

\section{Related Work}
\label{sec:related_work}
In this section, we review the evolving landscape of LLMs in psychiatry and mental health, focusing on both their application potential and the underlying evaluation frameworks. We begin by surveying clinical uses of LLMs, ranging from diagnostics and therapy to documentation, and examining the ethical and safety concerns arising from their deployment. We then analyze the major datasets that have shaped model development in this space, highlighting limitations in clinical grounding, generalization, and task complexity. Finally, we discuss the importance of using expert-validated knowledge sources, such as psychiatric casebooks and diagnostic manuals, to develop robust, clinically meaningful benchmarks.

\subsection{The Emergence of LLMs in Mental Healthcare}
{
The integration of LLMs into healthcare represents a significant technological shift, with the potential to transform clinical practice particularly within mental health care by leveraging their remarkable capabilities in language understanding and generation \cite{Sahu2025,LIN2025100868}. The field is witnessing a rapid expansion of LLM applications across the entire psychiatric workflow. In diagnostics, LLMs demonstrate effectiveness in detecting mental health disorders and suicidal ideation from textual data, often sourced from social media, and they can translate complex diagnostic manuals such as the ICD-11 into inspectable logic programs suitable for expert review \cite{na2025surveylargelanguagemodels}.

For therapeutic interventions, conversational agents such as Woebot and Wysa, grounded in principles of cognitive behavioral therapy, offer accessible and de-stigmatized eHealth services \cite{Fitzpatrick2017,Beatty2022}. Emerging research even from non-psychiatric contexts like primary care suggests that LLMs may in some cases be perceived as more empathetic or higher-quality communicators than human clinicians in written exchanges \cite{rousmaniere2025large,Hatch2025,Kuhail2024}. In clinical documentation, LLMs have demonstrated substantial potential to improve efficiency by automating the generation of therapy and medical notes. Notably, therapists have preferred LLM-generated behavioral health notes over human-written ones in blind assessments \cite{StanfordHAI2025,shah2025tnevalrubricevaluationprotocols}.

A primary concern is the propensity of LLMs to generate inaccurate or nonsensical outputs commonly referred to as “hallucinations” which poses substantial safety concerns in clinical settings where errors carry serious consequences \cite{aljohani2025comprehensivesurveytrustworthinesslarge}. These issues necessitate stringent ethical guidelines and reinforce the importance of maintaining a “human-in-the-loop” framework in which clinician oversight remains essential.

Clinicians also express ambivalence about widespread LLM adoption, citing worries about over-reliance on opaque “black-box” algorithms, potential diminution of clinical autonomy, and biases embedded in training data related to socioeconomic status or geography. Such biases risk exacerbating existing healthcare disparities \cite{Mancini2025}. This tension between the promise of democratizing access to mental health support and the ethical imperative of ensuring patient safety highlights the urgent need for rigorous, clinically grounded evaluation frameworks that can guide the responsible integration of LLMs into psychiatric practice.
}
\subsection{The Landscape of Datasets for Psychiatric Evaluation}
The evolution of LLM evaluation in mental health can be traced through the datasets used to train and test them. Early and influential resources include clinical interview corpora like the Distress Analysis Interview Corpus (DAIC-WOZ) and its extension, E-DAIC.\cite{ali2024leveragingaudiotextmodalities} These datasets, containing transcripts of semi-structured interviews and corresponding psychiatric assessment scores (e.g., PHQ-8 for depression), have been instrumental. They enabled innovative methods such as the Language Model for Impersonation-based Questionnaire Completion (LMIQ) framework, which attempts to bridge unstructured patient narratives with quantitative metrics \cite{rousmaniere2025large}. While a significant step, these methods face challenges in generalization beyond their specific benchmarks.

A second wave of datasets has leveraged the scale of social media and public surveys. Resources like the Reddit Mental Health Posts dataset, the Depression Detection Using Text corpus, and various mental health surveys have provided vast amounts of natural language data for tasks like sentiment analysis and binary classification \cite{Jin2025}. Models like Mental BERT and Mental RoBERTa were pretrained on millions of sentences from mental health-related subreddits to improve performance in these tasks \cite{na2025surveylargelanguagemodels}. However, the clinical validity of social media data is a significant concern due to its unverified nature, oversimplification of symptoms, and the risk of reflecting trends in self-diagnosis rather than formal clinical presentations.

More recently, the field has begun to address the need for more complex, conversational benchmarks. MentalChat16K, for instance, is a benchmark for conversational mental health assistance that combines synthetic data with anonymized clinical transcripts. \cite{xu2025mentalchat16kbenchmarkdatasetconversational}. Simultaneously, researchers have also investigated automated approaches for enhancing dataset richness and annotation quality. For example, {\cite{hassan2024automatedmultilabelannotationmental} introduced a synthetic labeling framework that transforms traditional single-disorder datasets into multi-label corpora by leveraging zero-shot and prompt-driven LLM annotation strategies. Their method, exemplified by the SPAADE-DR dataset, better captures co-occurring disorders such as depression, anxiety, PTSD, ADHD, and eating disorders, thus enabling more nuanced diagnostic modeling. This development reflects the maturation of the field, as research is moving beyond simple classification toward evaluating LLMs on their ability to engage in nuanced, therapeutic dialogue. Nonetheless, the reliance on synthetic data and LLM-as-a-judge evaluation methods introduces new layers of potential bias and unreliability.}

\subsection{Foundational Knowledge Sources in Psychiatric Practice}
To ensure clinical relevance and safety, any robust evaluation of LLMs in psychiatry must be grounded in the authoritative knowledge that defines the field. The gold standard for this knowledge is found in established psychiatric textbooks and clinical guides, which provide the basis for professional training and practice. The following sources represent the bedrock of clinical knowledge that LLMs must be tested against:

\begin{itemize}[leftmargin=0.25in] 
\item \textbf{Diagnostic Manuals and Casebooks:} Books like \textit{DSM-5 Clinical Cases}\cite{barnhill2013dsm5-clinical-cases} and its successor, \textit{DSM-5-TR Clinical Cases}\cite{Barnhill2022_DSM5TRClinicalCases}, are essential companions to the official diagnostic manual. They bring diagnostic criteria to life through realistic narratives and detailed clinical discussions, illustrating the complexities of comorbidity and differential diagnosis. Similarly, texts such as \textit{100 Cases in Psychiatry}\cite{wright2017_100casespsychiatry} and \textit{Case Files Psychiatry}\cite{toy2009casefiles-psychiatry} provide high-yield clinical scenarios designed to sharpen practical decision-making skills, making them ideal for evaluating an LLM's clinical reasoning.
\item \textbf{Specialized Clinical Guides:} Deep domain expertise is captured in specialized resources. \textit{Stahl's Essential Psychopharmacology}\cite{stahl2021essential} and its case-based companions are seminal texts on the neurobiological basis and practical application of psychiatric medications, crucial for evaluating treatment-related tasks. Likewise, \textit{Geriatric Psychiatry}\cite{hategan2024geriatric} provides comprehensive knowledge on the unique diagnostic and treatment considerations for elderly populations.
\item \textbf{Structured Knowledge and Assessment Tools:} To test factual recall and application of diagnostic criteria in a standardized format, resources like \textit{DSM-5-TR Self-Exam Questions} are invaluable. They provide thousands of multiple-choice questions that mirror the format of professional board examinations, offering a structured way to measure an LLM's knowledge base.
\end{itemize}
These foundational texts, curated and validated by experts over decades, stand in stark contrast to the unverified and often transient nature of data from social media. Grounding an evaluation dataset in these sources is a deliberate choice to prioritize clinical accuracy, safety, and alignment with established medical practice.

The DSM-5-TR (Diagnostic and Statistical Manual of Mental Disorders, Fifth Edition, Text Revision; American Psychiatric Association, 2022) is the current edition. Our source materials include textbooks and casebooks aligned with both the original DSM-5 (2013) and DSM-5-TR. We refer to "DSM-5-TR" when discussing current diagnostic criteria and standards, and specify "DSM-5" only when citing pre-2022 source materials by their original titles.

\subsection{Synthesis of Prior Work and Remaining Gaps}

In reviewing the current landscape, prior work has made significant strides in applying LLMs to psychiatric and mental health contexts. 
LLMs have proven highly effective at classification tasks, such as achieving a balanced accuracy of 94.8\% for depression detection and 96.2\% for PTSD when incorporated via decision-level fusion in multimodal systems{\cite{hassan2025leveragingembeddingtechniquesmultimodal}}. 
They have also demonstrated utility in administrative and workflow-support roles within healthcare settings{\cite{Tripathi2024_JAMIA_AdminTasks}}, 
and have shown surprising capabilities in generating empathetic responses comparable to, and sometimes exceeding, those of human clinicians\cite{Ayers2023_JAMA_ChatbotEmpathy,Sorin2024_JMIR_EmpathyReview}.

These advances, however, illuminate persistent methodological and conceptual gaps. 
Much of the field’s progress relies on datasets derived from social media platforms, which, while useful for large-scale data acquisition, lack the clinical rigor required for high-stakes psychiatric applications{\cite{Rai2024_PNAS_RaceMarkers,Guo2024_MentalHealthLLMsReview}}. 
Evaluation approaches have often been restricted to simple classification or sentiment analysis tasks, which fail to capture the multi-step reasoning and clinical judgment that define psychiatric practice{\cite{Tam2024_npjDM_QUEST,Bedi2024_JAMA_SystematicReview,Chen2025_EvalLLMsAgentsHealthcare}}. 
Even with more advanced reasoning frameworks, such as those used in multimodal IQ-style benchmarks, persistent challenges remain regarding generalization and the risk of hallucination in model outputs{\cite{Bang2025_ACL_HalluLens,Kalai2025_OpenAI_WhyHallucinate,deHond2024_LancetDH_Validation}}.

This synthesis reveals a critical gap in the literature: the absence of a comprehensive benchmark that is both grounded in authoritative, expert-validated psychiatric knowledge and designed to evaluate the full spectrum of complex clinical reasoning. 
There is an urgent need to move beyond binary classification and simple question-answering tasks to assess an LLM’s ability to perform sophisticated, clinically relevant reasoning such as formulating differential diagnoses, developing multi-stage treatment plans, and engaging in sequential, case-based analysis.

The present study is designed to address these limitations directly. 
By constructing a novel dataset meticulously curated from foundational psychiatric textbooks{\cite{Barnhill2022_DSM5TRClinicalCases}}
we ensure clinical validity, safety, and alignment with professional standards. 
Furthermore, by introducing  eleven distinct {\color{darkgreen}question-answer (QA) task types} ranging from diagnostic reasoning to treatment planning and sequential analysis, 
this framework provides a rigorous, multi-dimensional evaluation of LLM performance in psychiatric contexts. 
In doing so, it moves the field beyond surface-level competence toward assessing true readiness for the complex, high-stakes reality of clinical mental healthcare.

\subsection{Benchmarks for Evaluating LLMs in Mental Health}
Recently, two notable benchmarks illustrate different approaches for mental health and psychiatry. From Empathy to Action: Benchmarking LLMs in Mental Health with MentalBench-10 and a Novel Cognitive-Affective Evaluation Approach \cite{anonymous2025from} proposed a framework that emphasizes empathy, affective alignment, and the capacity of LLMs to translate understanding into actionable guidance. MentalBench-10 focuses on cognitive-affective evaluations, combining both synthetic and user-generated dialogues to assess empathy, appropriateness of advice, and general mental health support. While this contributes valuable insights into the emotional and supportive dimensions of LLM performance, it primarily emphasizes conversational empathy and cognitive-affective alignment rather than the depth of diagnostic reasoning. Similarly, PsychBench: A comprehensive and professional benchmark for evaluating the performance of LLM-assisted psychiatric clinical practice \cite{liu2025psychbenchcomprehensiveprofessionalbenchmark} introduced a wide-ranging evaluation platform that incorporates tasks from diagnosis and treatment to professional communication. Its evaluation metrics include ICD Primary Diagnosis Accuracy, Top Choice Alignment Score (TCAS), Medication Match Score (MMS), and Recommendation Coverage Rate (RCR). Although PsychBench benefits from real-world, diverse clinical cases and validation by psychiatrists, its relatively small dataset (300 cases) restricts generalization, and it still relies partly on synthetic dialogue construction, with limited grounding in authoritative, expert-validated psychiatric sources.

{In contrast to these existing efforts, our proposed benchmark, PsychiatryBench, represents a novel contribution that emphasizes clinically grounded, expert-validated evaluation. PsychiatryBench is curated from authoritative psychiatric textbooks and casebooks, aiming to provide comprehensive diagnostic reasoning and knowledge evaluation. It introduces eleven distinct QA tasks including diagnostic reasoning, treatment planning, sequential case analysis, and longitudinal follow-up covering more than 5,188 expert-annotated items. The benchmark relies on LLM-as-judge similarity scoring but is built on an expert-validated, broad knowledge base. By grounding its dataset exclusively in clinically validated sources (e.g., DSM-5-TR Clinical Cases, Stahl's Essential Psychopharmacology), PsychiatryBench directly addresses the gaps left by prior efforts. In contrast to MentalBench-10, which emphasizes affective alignment, and PsychBench, which provides broad but synthetic coverage, PsychiatryBench advances the field by offering a benchmark that is simultaneously clinically rigorous, multi-task, and extensible.}

\section{The PsychiatryBench Benchmark: Dataset Curation and Task Design}
\label{sec:The_PsychiatryBench_Benchmark}

\begin{figure}[!ht]
    \centering

    \begin{subfigure}[t]{0.48\linewidth}
        \centering
        \includegraphics[width=\linewidth]{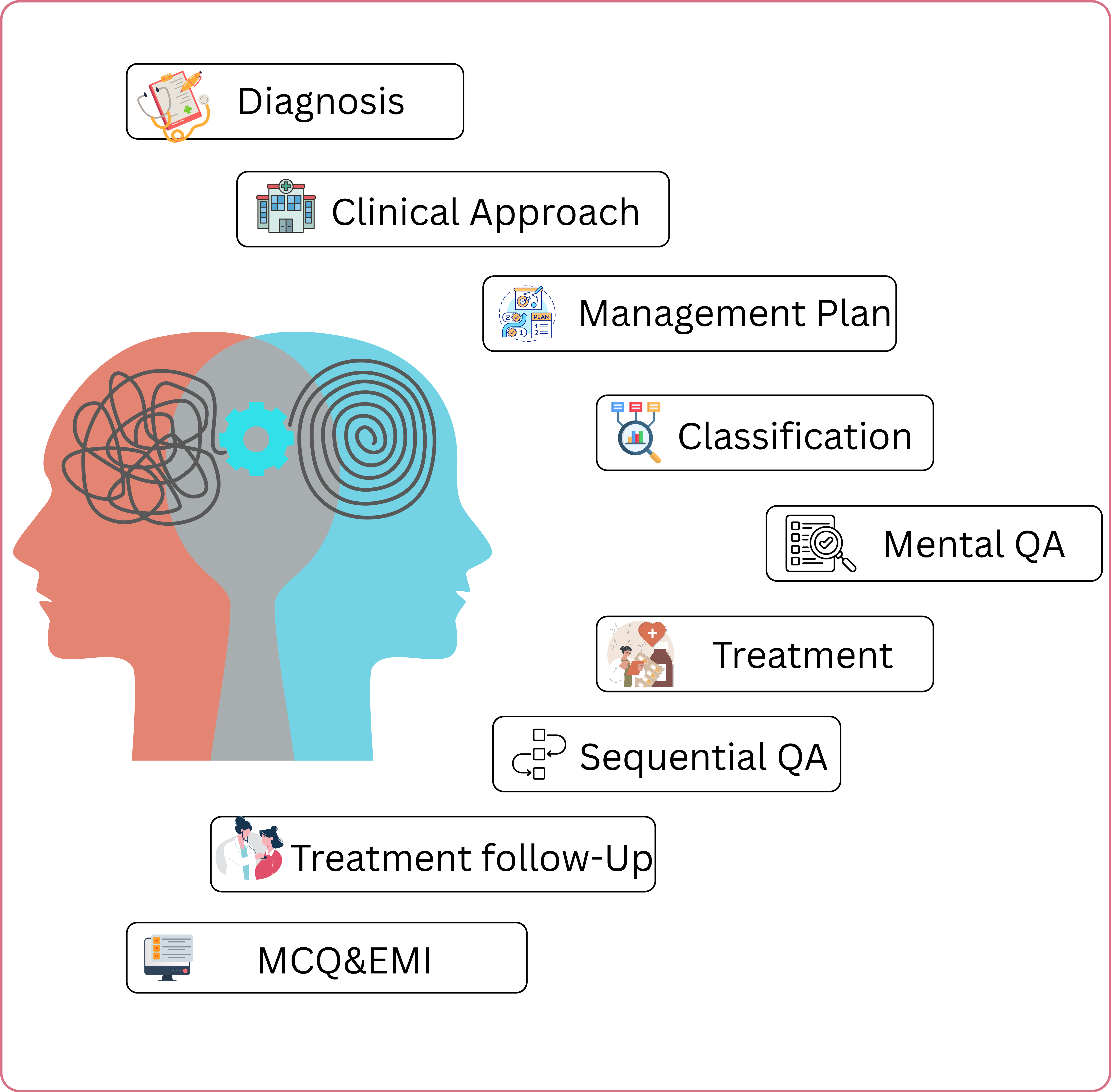}
        \caption{Overview of the full PsychiatryBench dataset, including manual extraction and filtering steps.}
                \label{fig:dataset_overview}
    \end{subfigure}
    \hfill
    \begin{subfigure}[t]{0.48\linewidth}
        \centering
        \includegraphics[width=\linewidth]{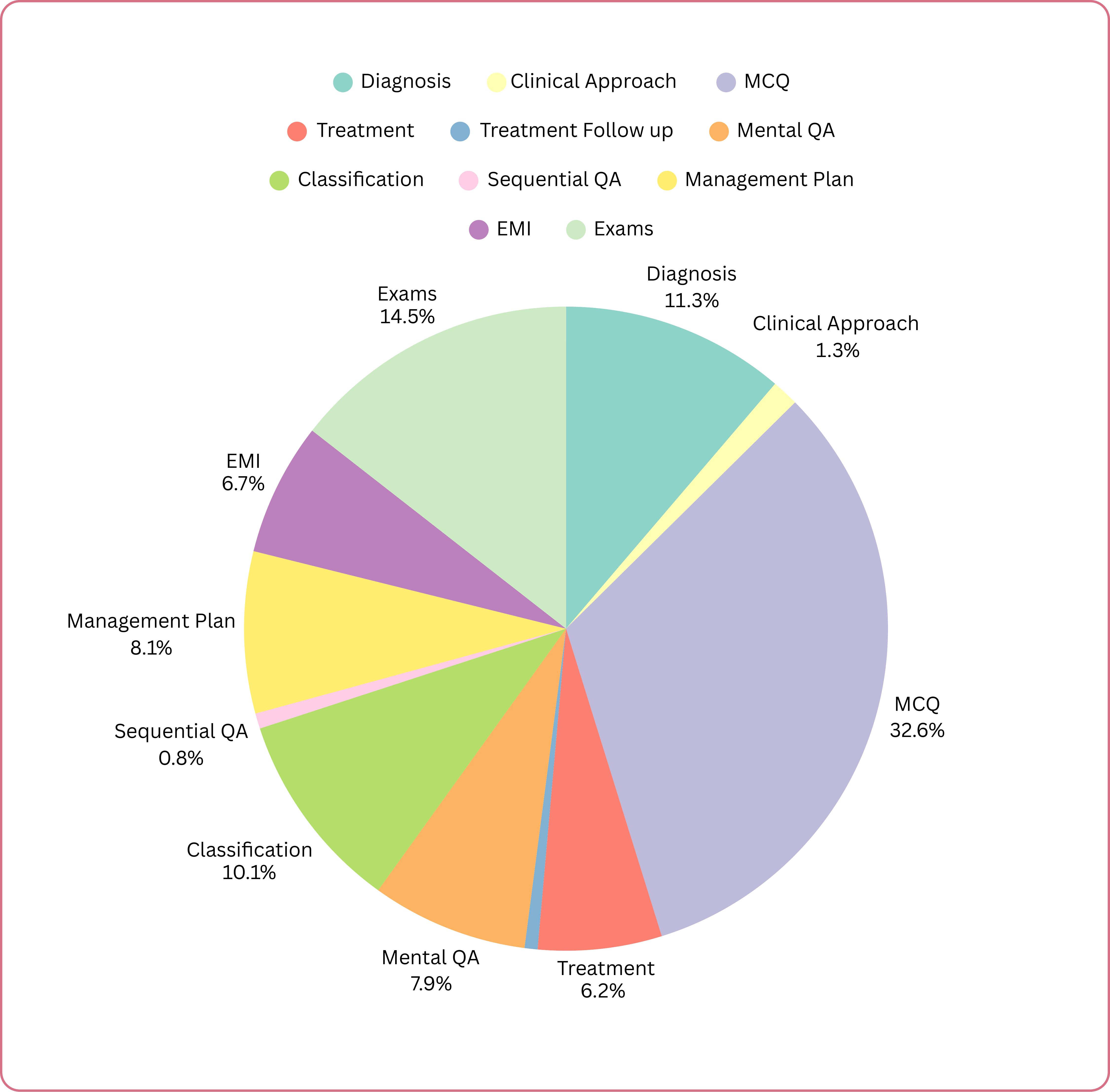}
        \caption{Structure of the selected subset (S-dataset) used for LLM evaluation.}
        \label{fig:s_dataset}
    \end{subfigure}
    \caption{Visual representation of the PsychiatryBench dataset development pipeline and its final evaluation subset.}
    \label{fig:dataset_comparison}
\end{figure}

The dataset created in this study was manually curated to address the unique challenges of psychiatric reasoning, with a focus on a diverse range of clinically relevant tasks. Sourced from authoritative psychiatry textbooks and expert-validated clinical resources, the dataset comprises natural language questions paired with expert-formulated answers. These questions are grounded in real-world psychiatric scenarios. They are designed to evaluate the reasoning, decision-making, and knowledge application abilities of LLMs in both clinical and educational contexts.
It is important to clarify the benchmark's intended clinical setting and acuity level. PsychiatryBench is primarily designed as a research and pedagogical benchmark to support model development and the systematic evaluation of psychiatric reasoning tasks. As such, its current focus is on adult outpatient psychiatric reasoning and not severely acute clinical contexts.

This emphasis on outpatient scenarios, where structured diagnostic and treatment reasoning can be most effectively assessed, means that the benchmark does not comprehensively represent high-acuity, emergency, or inpatient psychiatry scenarios (such as acute mania requiring admission, delirium, or imminent suicidality). We acknowledge that these contexts represent distinct and critical aspects of clinical care. This focus is a deliberate scoping choice, and the exclusion of these high-acuity presentations is further discussed as a limitation in Section \ref{sec:Limitations_and_Future_work}.

Figure \ref{fig:dataset_comparison} provides a visual summary of the composition across  
task categories, while Table \ref{tab:book-coverage} outlines the quantitative distribution of samples per task type. Together, they offer a comprehensive overview of the dataset structure and task diversity used in model evaluation.

The manual collection process ensured that the dataset maintains high-quality, medically accurate content, aligning with standardized diagnostic frameworks such as the DSM-5 and ICD-10, as well as clinical practice guidelines from organizations like the National Institute for Health and Care Excellence (NICE) and the American Psychiatric Association (APA).

\subsection{PsychiatryBench Development Pipeline}
\begin{figure}[!ht]
    \centering
\includegraphics[width=1\linewidth]{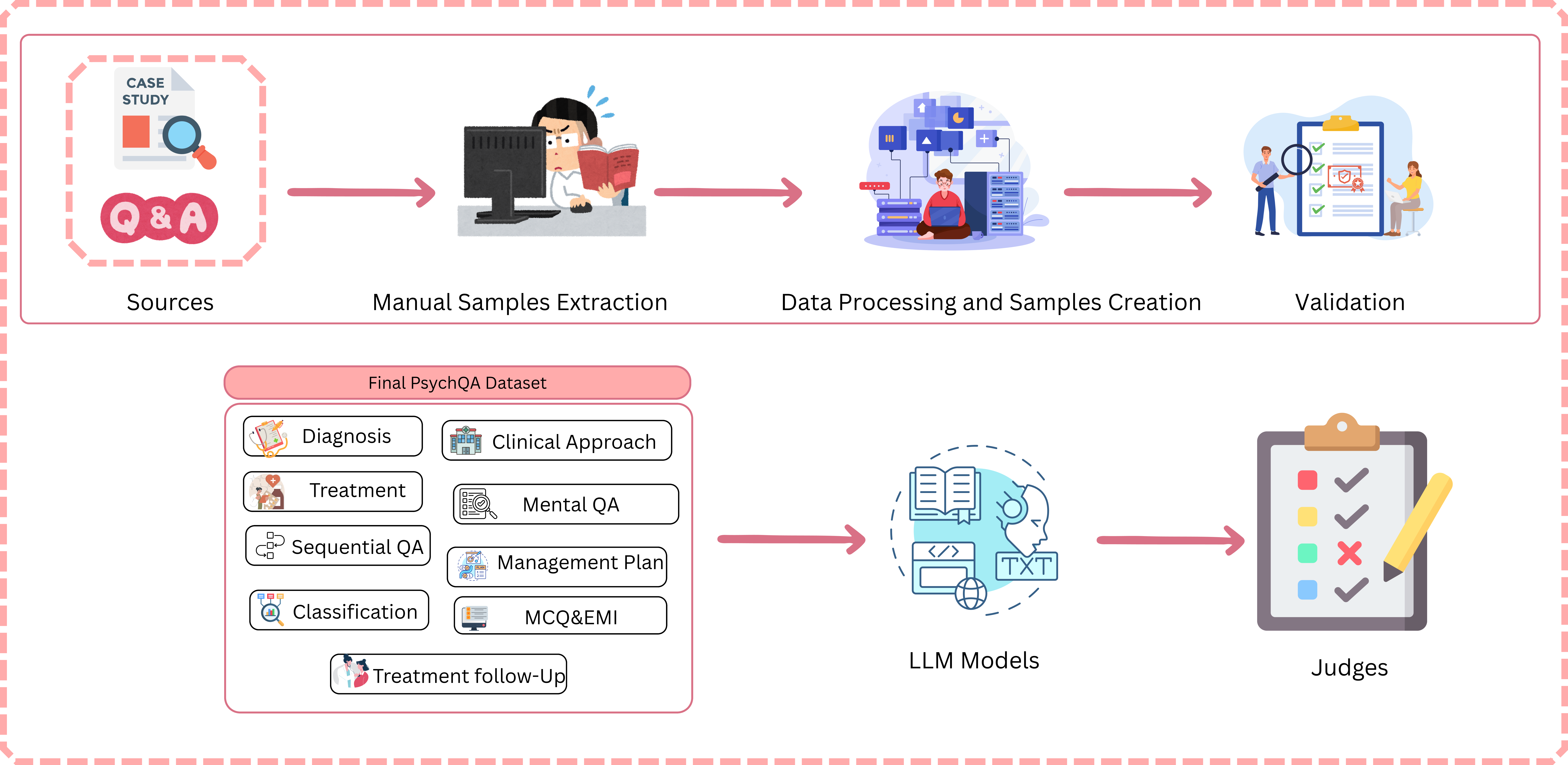}
    \caption{Workflow for the PsychiatryBench study: manual extraction and processing of clinical QA samples followed by LLM evaluation across eleven task types.}
    \label{fig:Flowchart}
\end{figure}

{
Our methodology, shown in Figure~\ref{fig:Flowchart}, begins with systematic manual extraction of content from standard evidence-based psychiatry textbooks. Subject matter experts thoroughly review and select clinical material with high educational and diagnostic value, vibrant case studies, descriptive patient scenarios, and disorder-specific treatment information. From this material, a diverse set of question-answer (QA) pairs is constructed to represent real-world diagnostic reasoning. The QA items are crafted across eleven core task types: Diagnosis, Treatment, Mental QA, Sequential QA, Management plan, Treatment Follow-up ,Clinical Approach, Classification, Multiple Choice Questions (MCQ), exams, and Extended Matching Items (EMI).

\begin{landscape}
\begin{table}[!ht]
  \centering
  \caption{{A summary of the psychiatry textbooks and case-based resources used for the dataset, showing the specific tasks covered by each source. A check mark (✅) indicates coverage of a task.}
}
  \label{tab:book-coverage}
  \normalsize
  \begin{tabular}{llllllllll}
    \hline
    Books & Diagnoses & Treatment & \shortstack{Treatment\\Follow‐Up}
          & Classification & \shortstack{Management\\Plan}
          &\shortstack{ Clinical \\Approach} & \shortstack{Mental QA }
          & \shortstack{Sequential\\ QA}
          & MCQs \\
    \hline
   Geriatric psychiatry \cite{hategan2024geriatric}            & ✅ & ✅ &   &   & ✅ &   &   &   & ✅ \\
Case Files Psychiatry \cite{toy2009casefiles-psychiatry}             & ✅ & ✅ &   & ✅ & ✅ & ✅ & ✅ &   & ✅ \\
Clinical Psychology Casebook \cite{choo2019clinical-psychology-casebook}      &   &   &   & ✅ &   &   &   &   &   \\
DSM-5 Clinical Cases \cite{barnhill2013dsm5-clinical-cases}              & ✅ & ✅ & ✅ &   &   &   &   &   & ✅ \\
DSM-5-TR Self-Exam Questions\cite{muskin2023dsm5tr-self-exam}      &   &   &   &   &   &   &   &   & ✅ \\
100 Cases in Psychiatry    \cite{wright2017_100casespsychiatry}        & ✅ & ✅ &   & ✅ & ✅ &   & ✅ &   &   \\
Core Clinical Cases   \cite{clark2011_core-clinical-cases-psychiatry}             & ✅ &   &   & ✅ &   &   &   & ✅ &   \\
Casebook in Abnormal Psychology \cite{brown2016_casebook-abnormal-psychology}    &   &   & ✅ & ✅ &   &   &   &   &   \\
Revision MCQ and EMI  \cite{puri2011_revision-MCQs-EMIs}             &   &   &   &   &   &   &   &   & ✅ \\
Psychiatry case vignettes  \cite{pall2002_underground-vignettes}        & ✅ &   &   & ✅ & ✅ &   & ✅ &   &   \\
Clinical Cases in Psychiatry \cite{altamura2019_clinical-cases-psychiatry}      &   &   & ✅ & ✅ &   & ✅ &   &   & ✅ \\
Stahl’s Psychopharmacology Cases \cite{cooper2021_stahl-case-studies3}   &   &   & ✅ &   &   &   &   &   &   \\
Stahl’s Essential Cases  \cite{radonjic2024_case-studies5}          & ✅ &   &   & ✅ &   &   &   &   & ✅ \\ \hline
  \end{tabular}
\end{table}

\vspace{1cm}

\begin{table}[!htt]
  \centering
  \caption{Distribution of available datasets categorized by clinical task type}
  \label{tab:number-of-datasets}
  \small
  \begin{tabular}{lc}
    \hline
Clinical Task                     & Number of samples \\ \hline
Diagnoses                         & 467                \\
Treatment                         & 258                \\
Treatment Follow-Up               & 27                 \\
Classification                    & 418                \\
Management Plan                   & 337                \\
Clinical Approach                 & 56                 \\
Mental QA                         & 326                \\
Sequential Question Answering     & 32                 \\
Multiple Choice Questions         & 1353               \\
Extended Matching Items           & 277                \\
Extended Matching Items Separated & 1037               \\
Exams                             & 600                \\ \hline
\textbf{Total}                    & \textbf{5188}      \\ 
\end{tabular}%
\end{table}

\end{landscape}

Following construction, the QA items are passed through a rigorous filtering process designed to ensure clarity, clinical validity, and educational utility. Ambiguous, overly simple, or redundant items are removed, and the remaining questions are manually reviewed for content balance and diagnostic diversity. These filtered QA items form the foundation of the PsychiatryBench dataset. In this phase, questions are sorted by task type and formatted consistently, allowing for structured downstream evaluation. A comprehensive data cleaning step is performed next, standardizing answer styles, resolving typographic issues, and ensuring uniform metadata. The result is a refined, high-quality QA dataset that tests not only factual recall but also diagnostic reasoning, treatment planning, and sequential clinical decision-making in psychiatry.

To evaluate the performance of LLMs, we apply each model to the finalized PsychiatryBench dataset. Each model is prompted to generate answers across all eleven selected task types. Importantly, in addition to serving as QA generators, LLMs are also used as evaluators or judges, scoring the generated responses based on accuracy, completeness, and clinical relevance. This dual use of LLMs helps automate and scale the evaluation process while maintaining consistency across judgments. A structured rubric is applied to produce a final score for each model, capturing performance across various clinical dimensions. This evaluation framework enables a robust and scalable benchmark for assessing how well LLMs understand, reason through, and respond to complex psychiatric questions, grounded in authentic medical education content.
}

{Psychiatry Bench evaluates these dimensions across its eleven task types. We acknowledge that some of our source texts, such as DSM-5-TR Clinical Cases and 100 Cases in Psychiatry, incorporate elements of the biopsychosocial model and functional assessment in their discussions. However, the benchmark tasks primarily emphasize diagnostic accuracy\textbf{}, treatment decision-making, and knowledge application. Notably, we do not systematically assess broader integrative or systemic dimensions such as narrative biopsychosocial formulation (holistic case conceptualization integrating biological, psychological, and social factors), detailed functional assessment, or systems-level collaboration (e.g., multidisciplinary team planning), nor do we assess interactive communication or conversational coherence.}

{
\subsection{Clinical Task Design and Implementation:}
{
The overall objective of capturing the multifaceted reasoning processes essential to psychiatric practice served as a guide for the creation and execution of the PsychiatryBench clinical tasks. Each task type was designed to assess a distinct aspect of clinical competency, including sequential reasoning, longitudinal management, treatment decision-making, and diagnostic formulation. Real clinical cases were converted into structured question–answer pairs that emulate real-world diagnostic and therapeutic reasoning, using only reliable, expert-validated psychiatric sources. The inclusion of medical history and mental state examination (MSE) elements varies across cases, reflecting the structure of the original source texts rather than deliberate design choices. This variability was retained to preserve the authenticity and instructional style of the reference materials. This section outlines the conceptual rationale, data generation procedures, and task-specific frameworks used to operationalize these competencies, thereby providing a foundation for the systematic, clinically grounded evaluation of LLMs in psychiatry. For additional transparency, representative sample items from multiple PsychiatryBench task types are provided in Appendix~\ref{sec:Samples}.
}

\subsubsection{Diagnosis}
This subset contains 467 expert-annotated question-answer pairs centered on diagnostic reasoning. The task focuses on determining psychiatric diagnoses based on rich clinical vignettes that simulate real-world case presentations. Each item includes a clinical vignette and a diagnostic query, requiring inference of the most appropriate psychiatric diagnosis. These scenarios often incorporate elements such as the patient’s psychiatric and medical history, presenting symptoms, findings from the MSE, and physical examination.
The cases, representing 123 unique diagnostic questions and 295 distinct clinical histories, require models to identify either a likely diagnosis or generate a differential diagnosis, drawing on a nuanced understanding of symptom patterns and temporal progression. Examples are drawn from six authoritative sources, including 100 Cases in Psychiatry and Case Files Psychiatry. This task challenges models to interpret and synthesize complex narratives and apply diagnostic reasoning aligned with formal criteria from standard classifications like the DSM-5-TR, making it critical for evaluating systems intended for clinical decision support and medical education.
\subsubsection{Treatment}
This subset comprises 258 expert-annotated clinical question-answer pairs focused on treatment decision-making. The task centers on identifying appropriate treatment strategies based on clinical scenarios that include a detailed case history, a MSE, physical examination ,and a treatment-related question. The expert-curated answers reflect real-world psychiatric care and cover a wide range of disorders and treatment modalities, including pharmacological interventions (e.g., prescribing antidepressants, antipsychotics, mood stabilizers), psychotherapeutic approaches (e.g., cognitive behavioral therapy, psychoeducation, motivational interviewing), and acute care level decisions such as hospitalization when risk severity or treatment resistance requires intensive monitoring and intervention. Sourced from texts like 100 Cases in Psychiatry and Geriatric Psychiatry, this task tests a system’s ability to apply evidence-based psychiatry and personalize care according to clinical guidelines.
\subsubsection{Treatment Follow-Up}
This dataset comprises 27 clinical question-answer pairs focused on the longitudinal management of psychiatric patients after an initial treatment decision has been made. Each scenario includes a psychiatric case history, a primary follow-up question, and often one or more follow-up reports (e.g., new symptoms, lab results), with a subset of 6 entries extending to a second-tier follow-up. These items evaluate a model’s capacity to support ongoing clinical care by interpreting medication effects, identifying emerging side effects, assessing adherence, or recommending adjustments to the treatment plan. The task simulates ongoing psychiatric care, testing a model's ability to adapt recommendations over time and handle the incomplete, evolving information that is critical for chronic care planning.
\subsubsection{Classification}
The classification component was developed to evaluate the ability of LLMs to categorize psychiatric disorders based on clinical descriptions. It includes 418 psychiatric case descriptions, each annotated with both a broad diagnostic category and a specific disorder label. Clinical case samples were manually extracted from authoritative casebooks and annotated with preliminary labels. To standardize terminology, two reference lists were created from the DSM-5-TR: one for high-level diagnostic categories (e.g., "mood disorders") and another for specific disorders (e.g., "bipolar I disorder"); see Appendix~\ref{subsec:Class}A for the full lists. A mapping process was then applied to align each sample with both the validated category-level and disorder-level labels, enabling multi-label classification. Both lists were checked for clinical accuracy by a licensed psychiatrist.

\subsubsection{Management Plan}
This dataset consists of 337 case-based question-answer pairs that test a model’s ability to generate a structured and contextually appropriate management plan. Each scenario provides a detailed clinical history, often supplemented with MSE and physical examination findings, followed by a question like "How would you manage this case?". The answers outline holistic, step-by-step care strategies that integrate diagnostic clarification, risk assessment, therapeutic planning (pharmacological and psychological), psycho-education, social interventions, and follow-up.The task emphasizes comprehensive care planning and management decision-making beyond single diagnoses or treatment choices, reflecting the complex care coordination required in real-world psychiatric settings.

\subsubsection{Clinical Approach}
The development of this subset began with extracting patient case scenarios from psychiatric reference books. Since these source vignettes were not accompanied by ready-made questions, it was necessary to transform them into evaluable QA pairs. For this purpose, these cases were used to generate open-ended clinical reasoning questions using the Gemini 2.5 Pro model, which was instructed to formulate questions requiring multi-step interpretive reasoning. Subsequently, each generated question underwent manual verification to ensure its clinical relevance. This subset contains 56 expertly crafted scenarios designed to evaluate a model’s ability to emulate the diagnostic thought process of a psychiatrist. Unlike tasks that ask for a final diagnosis, this component focuses on process-oriented reasoning, assessing how clinicians should gather information, rule out critical disorders, prioritize differential diagnoses, and choose initial interventions. The questions emphasize the rationale behind clinical decisions, such as identifying red flags or determining the most appropriate investigative step. This task is essential for assessing how well LLMs can replicate the nuanced cognitive workflow of trained psychiatrists. Unlike tasks that ask for final diagnoses or treatments, this task emphasizes process-oriented clinical reasoning how clinicians gather information, rule out critical disorders, prioritize differential diagnoses, and choose appropriate initial interventions.
\subsubsection{Mental QA}
This dataset includes 326 expert-annotated question-answer pairs that test foundational psychiatric knowledge through the definition and clarification of core concepts, syndromes, and clinical terms. Questions, sourced primarily from the Case Files and Case Vignettes series, ask models to define terms like "bizarre delusions," explain concepts such as "thought withdrawal," or describe pharmacological classes. This task targets the factual psychiatric knowledge and terminology aligned with the DSM-5-TR and formal training curricula, enhancing conceptual grounding and promoting explainability in systems intended for clinical education or knowledge base construction.
\subsubsection{Sequential Question Answering}
This task simulates dynamic, case-based diagnostic reasoning through a dataset of 32 clinical vignettes, each accompanied by a structured sequence of interrelated questions. These questions were designed to reflect the natural progression of a psychiatric evaluation, encompassing five core domains of clinical reasoning: (1) differential diagnosis, (2) supporting evidence, (3) etiological factors, (4) treatment options, and (5) prognosis. To extend the evaluative scope of the dataset, we introduced an additional sixth question classification based on DSM-5-TR, which is not conventionally included in standard psychiatric textbooks. This item was purposefully developed to facilitate diagnostic classification, enabling each vignette to be mapped to a unique label representing the preferred diagnosis. This multi-turn structure provides a rigorous framework for assessing a model’s capacity for longitudinal reasoning, contextual retention, and adaptive clinical judgment across sequential interactions. For example, in a vignette describing a 10-year-old boy presenting with poor eye contact, delayed speech, and repetitive behaviors, the model is guided through the reasoning sequence: identifying possible differential diagnoses (e.g., Autism Spectrum Disorder vs. Communication Disorder), outlining supporting evidence from the history, exploring potential etiological factors, proposing treatment options, and discussing prognosis. The final classification question then requires the model to determine the preferred DSM-5-TR diagnosis, which in this case would be Autism Spectrum Disorder.

\subsubsection{Question Formats and Exam Simulations}
This task group encompasses a variety of structured assessment formats commonly used in psychiatric board examinations and educational settings to assess knowledge recall, clinical recognition, and treatment matching. The development process for all items involved manually curating questions from reputable psychiatry resources and implementing a dedicated preprocessing step to remove all accompanying explanatory text, ensuring an unbiased evaluation format.

\textbf{Standard Multiple-Choice Questions (MCQ)}

This component consists of 1353 standard multiple-choice questions, each with 4-5 answer options. These items reflect the structured testing style common in psychiatric board examinations and are designed to function as clinical knowledge checks grounded in patient histories and symptom-based reasoning. For instance, a question might describe a patient with specific movements and a family history of psychiatric illness, requiring the identification of a gene location associated with a particular disease. As noted in the introductory description for this section, a dedicated preprocessing step was implemented. Many of the original source items were accompanied by detailed explanations or rationales, which, while pedagogically useful, could introduce bias in LLM evaluation. To prevent this, all explanatory text was removed, ensuring only the question stem and options were retained for an unbiased evaluation. Furthermore, all answer formats were standardized from various original formats (e.g., Roman numerals) into a consistent A/B/C/D labeling scheme, and ambiguous or duplicate questions were filtered to maintain quality.

\textbf{Extended Matching Items (EMI)}
This task is designed to assess deeper clinical reasoning by requiring models to match multiple clinical vignettes to a single, shared list of 8-15 answer options (the "theme"). This format tests pattern recognition and fine-grained differentiation across closely related disorders or treatments. The dataset contains 277 whole EMI clusters. To support more granular evaluation and fine-tuning, the full EMI format was also disaggregated into 1037 standalone "EMI Separated" question items. Each separated item retains its original header and option list for context but functions as an independent unit, simplifying scoring and improving input granularity during training.
}

\textbf{Exam Simulations}
This component includes curated exam-style datasets designed to simulate standardized testing conditions and replicate the structure of real psychiatric board exams. This set contains 600 questions split into two parts for evaluation: 300 EMI questions and 300 MCQs. These items are designed to assess factual recall and decision-making across a complete test, allowing for the evaluation of a model's performance in a sustained, exam-like scenario.

\section{Experimental and Evaluation Methodology}
\label{sec:Experimental_and_Evaluation_Methodology}
{This section outlines the experimental framework used to evaluate the performance of LLMs across the aforementioned tasks {(as defined in Section~\ref{sec:The_PsychiatryBench_Benchmark})}, {which simulate/reflect real-world clinical workflows}. We describe the models employed, including both general-purpose LLMs and domain-specialized medical models. {The evaluation methodology (prompting, parsing, and metrics) is summarized here; details of the benchmark construction are provided in Section~\ref{sec:The_PsychiatryBench_Benchmark}.} The evaluation methodology {combines} key implementation details such as prompting and output parsing, automatic metrics, and LLM-as-a-judge assessments to capture performance on accuracy, reasoning depth, and clinical alignment. By structuring our experiments around tasks that simulate/reflect realistic clinical scenarios and rigorous evaluation criteria, we aim to produce insights into the capabilities and limitations of current LLMs in mental health domains.}

\subsection{Language Models under Evaluation}

{In this section, we detail the LLMs evaluated throughout our study, covering both general-purpose and domain-specialized systems. The goal is to assess their capabilities in handling various clinically relevant tasks {(see Section~\ref{sec:The_PsychiatryBench_Benchmark} for task definitions)}, such as diagnosis, treatment planning, and diagnostic reasoning. By comparing general LLMs with medical-domain-specific models, we aim to highlight the performance trade-offs and advantages of each category. This dual perspective provides a clearer understanding of how LLMs can be optimized or selected for different healthcare applications.}

\subsubsection{General Models}
{

\begin{table}[!ht]
\centering
\caption{Summary of LLMs used in our experiments}
\resizebox{\textwidth}{!}{%
\begin{tabular}{lllll}
\toprule
\textbf{Model} & \textbf{Size (Parameters)} & \textbf{Source} & 
\begin{tabular}[c]{@{}l@{}}\textbf{API Service}\\\textbf{Provider}\end{tabular} & 
\begin{tabular}[c]{@{}l@{}}\textbf{Release (Year}\\\textbf{\& Quarter)}\end{tabular} \\
\midrule
Gemini 2.5 Pro Preview 03-25 \cite{google2024geminiapi} & Proprietary & Closed-source & Google's Gemini API & {2025 Q1} \\
Gemini 2.5 Flash Preview 04-17 Thinking \cite{google2024geminiapi} & Proprietary & Closed-source & Google's Gemini API & {2025 Q2} \\
Gemini 2.5 Flash Preview 04-17 \cite{google2024geminiapi} & Proprietary & Closed-source & Google's Gemini API & {2025 Q2} \\
Gemini 2 Flash \cite{google2024geminiapi} & Proprietary & Closed-source & Google's Gemini API & {2025 Q1} \\
Claude Sonnet 4 \cite{sonnet2024} & Proprietary & Closed-source & Anthropic API & {2025 Q2} \\
{Claude Sonnet 4.5 \cite{sonnet2024}} & {Proprietary} & {Closed-source} & {Anthropic API} & {2025 Q3} \\
{Claude Sonnet 4.5 Thinking} \cite{sonnet2024} & {Proprietary} & {Closed-source} & {Anthropic API} & {2025 Q3} \\
{GPT 5 Medium Thinking \cite{openai2025gpt5}} & {Proprietary} & {Closed-source} & {OpenAI API} & {2025 Q3} \\
DeepSeek-R1 \cite{DeepSeek2024} & 671B & Open-source & DeepSeek API & {2025 Q1} \\
DeepSeek Chat \cite{DeepSeek2024} & 67B & Open-source & DeepSeek API & {2024 Q4} \\
LLaMA 3.3 70B \cite{dubey2024llama} & 70B & Open-source & Together API & {2024 Q4} \\
QWQ 32 \cite{qwen2024} & 32B & Open-source & Nvidia API & {2025 Q1} \\
Qwen 3 32B \cite{qwen2024} & 32B & Open-source & Deepinfra API & {2025 Q2} \\
{GPT-OSS \cite{dubey2024llama}} & {120B} & {Open-source} & {Together API} & {2025 Q3} \\
\bottomrule
\end{tabular}%
}
\label{tab:models}
\end{table}

Table ~\ref{tab:models} presents a concise overview of the LLMs evaluated in the study. It categorizes each model by name, parameter size, source type (open or closed), and the API service provider used to access them.The models include both proprietary and open-source systems. Proprietary models, such as various versions of Google's Gemini (Gemini 2.5 Pro Preview, Gemini 2.5 Flash "Thinking," and Gemini 2 Flash), Anthropic's Claude series (Claude Sonnet 4, 4.5, and 4.5 Thinking), and OpenAI's GPT 5 Medium Thinking, are closed-source and there is no official public data on model size. As their architectures and training details are not publicly available, they are accessed through their respective API services. Models with 'Thinking' in their name (e.g., GPT 5 Medium Thinking) may be referred to with a (T) suffix for brevity. 

On the other hand, the study includes open-source models like DeepSeek-R1 (671B), DeepSeek Chat (67B), LLaMA 3.3 70B, QWQ 32 (32B), Qwen 3 32B, and GPT-OSS (120B). These models offer more transparency and are made available through platforms such as the DeepSeek API, Together API, Nvidia API, and Deepinfra API. These models have sizes ranging from 32B-671B parameters.  
Overall, the table illustrates a diverse selection of LLMs sourced from both major tech companies and open research communities, providing a broad basis for comparative experimentation and analysis.

}

\subsubsection{Medical Models}
\label{medical_models}

\begin{table}[]
\centering
\caption{Summary of Medical Models used in our experiments}
\label{tab:med_model}
\begin{tabular}{lclll}
\hline
\textbf{Model}                                                                 & \textbf{Size (Parameters)} & \textbf{Source}   & \textbf{API Service Provider} & \begin{tabular}[c]{@{}l@{}}\textbf{Release (Year}\\\textbf{\& Quarter)}\end{tabular}
\\ \hline
Llama3-OpenBioLLM-70B \cite{openbiollm2024}                                    & 70B                        & Open-source       & Hugging Face                  & 2024 Q2                             \\
Palmyra-Med-70B \cite{palmyra_med_2024}                                        & 70B                        & Open-source       & Hugging Face                  & 2025 Q1                             \\
Llama-MedX v3.2 \cite{llama_medx_v3_2_2025}                                    & 70B                        & Open-source       & Hugging Face                  & 2025 Q1                             \\
JSL-MedLlama-3-8B-v2.0 \cite{jsl_medllama3_8b_v2_0_2025}                       & 8B                         & Open-source       & Hugging Face                  & 2025 Q2                             \\
MedGemma \cite{medgemma2025}                                                   & 27B                        & Open-source       & Hugging Face                  & 2025 Q2                             \\
II-Medical-8B \cite{2025II-Medical-8B}                                         & 8B                         & Open-source       & Hugging Face                  & 2025 Q3                             \\
Med-PaLM (v2) \cite{medpalm_v2_2024}                                           & Proprietary                & Closed-source     & Google Cloud Vertex AI        & 2023 Q1                             \\ \hline
\end{tabular}
\end{table}

{To ensure a well-rounded and insightful evaluation of LLMs in the context of psychiatry, we selected a diverse group of medical-focused models that reflect a range of architectures, parameter scales, and accessibility levels, as shown in Table \ref{tab:med_model}. This curation was designed to balance frontier performance with practical considerations such as openness, reproducibility, and ease of deployment.

Our model set includes high-capacity open-source systems like Llama3-OpenBioLLM-70B, Palmyra-Med-70B, Llama-MedX v3.2, and MedGemma, which leverage recent advances in foundational models and are optimized for biomedical and clinical tasks. We also included JSL-MedLlama-3-8B-v2.0, a smaller model aimed at providing strong performance while remaining lightweight and adaptable. These open-source models are readily accessible via platforms such as Hugging Face and are suitable for both academic research and fine-tuned clinical deployments.

In addition, we explicitly report each model’s Release (Year \& Quarter) to contextualize performance within the rapidly evolving landscape of medical-oriented LLMs. Because model capabilities improve significantly over short development cycles, situating each system within its temporal release window enables more meaningful comparisons. This temporal framing highlights trends such as the recent surge of high-capacity open-source biomedical models, while also clarifying how older, proprietary systems like Med-PaLM (v2) continue to serve as reference benchmarks. Including release timing therefore supports a more accurate interpretation of model maturity, recency, and relevance for clinical and educational applications.

In contrast, we also evaluated Med-PaLM (v2), a proprietary model developed by Google DeepMind. While not open-source, Med-PaLM represents a high-performing benchmark for medical reasoning and is accessible through Google Cloud’s Vertex AI platform. Its inclusion provides an important comparison point for understanding the trade-offs between commercial-grade deployment and the flexibility of open systems.

By including both open and closed models of varying sizes, we aim to assess not only raw accuracy on psychiatric tasks but also broader factors such as infrastructure requirements, cost, adaptability, and transparency, key considerations for real-world clinical and educational use. A summary of these models, along with their sources and access points, is provided in Table~\ref{tab:med_model}.

\subsection{Evaluation Framework}
To assess the performance and alignment of model outputs, we employ a set of evaluation metrics designed to quantify agreement, consistency, and quality across different prompts and models. These metrics allow us to systematically compare outputs, identify patterns in model behavior, and measure improvements across prompt variations. In particular, we focus on agreement counts between model pairs to capture alignment and shared reasoning under different prompting conditions.

\subsubsection{Accuracy}

Accuracy is the proportion of correctly classified samples among the total number of samples.
It provides an overall measure of correctness; however, it may be misleading in the presence of
class imbalance, as it can be biased toward the majority class.

\begin{equation}
\text{Accuracy} = \frac{\text{Number of correct predictions}}{\text{Total number of predictions}}
\end{equation}

\subsubsection{Extended Matching Item Evaluation}
In the EMI set, each question consists of multiple subquestions, all of which must be answered based on shared response options. To ensure accurate data evaluation, the dataset was assessed in two distinct formats: first, as a complete EMI set (full-format evaluation), and second, by treating each subquestion as an independent query with its evaluation (separated format).

{In the full EMI set format, scoring follows the Partial Scoring System (PCS), which assigns partial credit based on the proportion of correctly answered subquestions in a set. In contrast, the separated subquestion format evaluates each subquestion independently using a binary scoring system. For both formats, any instance of multiple answers being given for a single question is treated as incorrect and scored as zero. the evaluation The evaluation is done as follows:}

\paragraph{1. Partial Scoring System (PCS) - Full EMI Set Evaluation}

This evaluation method measures how well the model performs on grouped EMI questions, where each set consists of multiple related subquestions. The scoring system allows for partial credit, ensuring that models are not penalized excessively for a few incorrect subanswers.

The score for the \( i \)-th EMI set is computed as the proportion of correct subquestions:
\begin{equation}
EMI\_Score_i = \frac{1}{S_i} \sum_{j=1}^{S_i} c_{ij}
\label{eq:emi_score}
\end{equation}
{ where \( S_i \) is the number of subquestions in the \( i \)-th EMI set, and \( c_{ij} \in \{0, 1\} \) is a binary indicator denoting whether the \( j \)-th subquestion in the \( i \)-th set is answered correctly.}

The overall EMI accuracy is then calculated as the mean score across all EMI sets:
\begin{equation}
EMI\_Accuracy = \frac{1}{N} \sum_{i=1}^{N} EMI\_Score_i
\label{eq:emi_accuracy}
\end{equation}
{where \( N \) is the total number of EMI sets.}

\paragraph{2. Individual Subquestion Evaluation - Separated EMI}

This approach evaluates each subquestion independently, allowing for standard binary scoring per subquestion. It offers a fine-grained view of model performance without being affected by the structure of the EMI sets.

The overall accuracy is then computed as:
\begin{equation}
\text{Subquestion\_Accuracy} = \frac{1}{T} \sum_{k=1}^{T} a_k
\label{eq:subq_accuracy}
\end{equation}
where \( T \): Total number of individual subquestions across all EMI sets and equals $\sum_i S_i$, and \( a_k \in \{0, 1\} \): Binary indicator of correctness for the \( k \)-th subquestion when evaluated independently.

\paragraph{Answer Integrity Rule - Penalty for Multiple Answers}

In both evaluation methods, if the model produces multiple answers for a single question (whether set-based or individual), the answer is considered invalid and marked as incorrect. This ensures clarity, fairness, and aligns with human assessment standards.

\begin{equation}
c_{ij} = 0 \quad \text{if multiple answers are provided}
\label{eq:multi_answer_penalty}
\end{equation}

\textbf{Common Correct:}  
This measures the number of questions answered correctly in both formats:
\begin{equation}
\text{CommonCorrect} = \sum_{i=1}^{N} \mathbf{1}\left(F_i = 1 \land S_i = 1\right)
\end{equation}
where  $F_i$ and $S_i$ are indicator variables for \textit{Full format} response and \textit{Separated format} response on question $i$, respectively. $F =1$ if the model answers question $i$ correctly in the Full format and 0 otherwise. $S_i = 1$ if the model answers question $i$ correctly in the Separated format and 0 otherwise. $\land$: The logical AND operator. { And, \( \mathbf{1}(\cdot) \) is the indicator function that returns 1 if the condition inside is true and 0 otherwise.}

\textbf{Common Incorrect:}  
This measures the number of questions answered incorrectly in both formats:
\begin{equation}
\text{CommonIncorrect} = \sum_{i=1}^{N} \mathbf{1}\left(F_i = 0 \land S_i = 0\right)
\end{equation}

\textbf{Overall Consistency:}  
This score reflects how often the model yields the same result (correct or incorrect) across both formats:
\begin{equation}
\text{Consistency} = \frac{1}{N} \sum_{i=1}^{N} \mathbf{1}\left(F_i = S_i\right)
\label{eq:overall_consistency}
\end{equation}

\textbf{Format Agreement Rate:}  
This metric calculates the proportion of questions for which the model was correct in both formats, relative to all cases where the model was correct in at least one:
\begin{equation}
\text{Agreement} = \frac{C}{C + O_F + O_S}
\end{equation}
{ where \( C \) is the number of the correct questions, questions the correct questions in both formats, \( O_F \) are the correct questions the correct questhetions only in Full format, and \( O_S \) are the correct questions the correct questions only in the separated format.}

\textbf{Format Divergence Rate:}  
This captures how often the model's correctness depends on format. A high divergence suggests inconsistent reasoning between question structures:
\begin{equation}
\text{Divergence} = \frac{O_F + O_S}{N}
\end{equation}

These consistency metrics are crucial for evaluating the robustness and fairness of models, especially in educational and clinical settings where presentation format should not influence correctness.

\subsubsection{Classification Evaluation}

Since the classification task is multi-label in nature, each instance may be associated with multiple correct labels. To accurately assess model performance in this setting, we employed two complementary evaluation metrics: \textbf{Subset Accuracy} and \textbf{Weighted-Average F1-Score}. These metrics together offer a balanced view rewarding exact correctness while also accounting for partial matches and class imbalance \cite{wu2020multilabelclassificationhammingloss}.

\paragraph{Subset Accuracy (Exact Match Ratio)}  
Subset Accuracy is a stringent metric that only considers a prediction correct if the predicted label set exactly matches the true label set no missing or extra labels are allowed. This makes it particularly valuable when precision in full-label prediction is critical. It answers the question: "How often did the model get everything exactly right?"

\begin{equation}
\text{Subset Accuracy} = \frac{1}{N} \sum_{i=1}^{N} \mathbf{I}(\hat{Y}_i = Y_i)
\end{equation}
where \(N\) is the total number of samples, \(Y_i\) is the set of true labels for the \(i\)-th sample,
 \(\hat{Y}_i\) is the set of predicted labels,
\(\mathbf{I}(\cdot)\) is the indicator function, and equal to 1 if \(\hat{Y}_i = Y_i\), and 0 otherwise.

\paragraph{Weighted-Average F1-Score (F1-weighted)}  
While Subset Accuracy requires perfect prediction, the F1-weighted score provides a more flexible and balanced view of model performance. It evaluates how well the model balances \textit{precision} (avoiding false positives) and \textit{recall} (avoiding false negatives) for each label independently. Then, it aggregates these per-label F1-scores into a single number using a weighted average, where more common labels contribute more to the final score.

First, the F1-score for each label \(l\) is calculated as the harmonic mean of precision (\(P_l\)) and recall (\(R_l\)):

\begin{equation}
F_{1,l} = 2 \times \frac{P_l \times R_l}{P_l + R_l}
\end{equation}

Then, the overall weighted-average F1-score across all labels \(\mathcal{L}\) is computed as:

\begin{equation}
\text{F1-weighted} = \sum_{l \in \mathcal{L}} w_l \cdot F_{1,l}
\end{equation}

Where \(\mathcal{L}\) is the set of all labels, and \(w_l\) is the proportion of samples in which label \(l\) appears in the ground truth.

This metric ensures that performance on frequent labels contributes more heavily, helping to mitigate distortions caused by class imbalance.}

\subsubsection{LLM as a judge}

To quantitatively evaluate the output quality of model-generated answers against ground truth references, we employed an automated approach leveraging an LLM as an expert judge \cite{gu2024surveyllmasajudge}. Specifically, Llama 3 70B was configured for this evaluation task. The core objective was to compute a similarity score, ranging from 0 to 100, representing the degree of substantive congruence between the model-generated response (referred to as the \textit{Candidate Answer}) and the reference ground truth (referred to as the \textit{Reference Answer}). Importantly, the evaluation was designed to focus on the core clinical content, including reasoning, correctness, and completeness of the response while disregarding differences in writing style or phrasing Appendix \ref{subsec:Ev_Judging}.

Recognizing the sensitivity of LLM evaluations to prompt design, we systematically compared two distinct evaluation prompts generated by GPT-4.5 and Gemini 2.0 Pro. The GPT-4.5 prompt employed a rigid point-based scoring rubric with specific weights allocated to different content components (full prompt in appendix figure ~\ref{fig:Prompt-generated-GPT-4.5}), while the Gemini-generated prompt adopted a logic-driven, few-shot learning approach emphasizing holistic understanding of clinical reasoning and semantic meaning  (full prompt in appendix figure ~\ref{fig:generated-Gemini-for-Llama-evaluation.}). After empirical testing with LLaMA 3.3 70B as the judge model, we selected the Gemini-generated prompt for our final evaluation pipeline. This decision was driven by its superior consistency and reliability across diverse psychiatric tasks, with the logic-driven, example-based approach eliciting more reliable similarity scores compared to the rigid, point-based system.

To identify the most reliable judge model, we conducted a comparative evaluation using three candidate LLMs: LearnLM (J1), LLaMA 3.3 70B (J2), and GPT-4o mini (J3). As detailed in  Appendix \ref{subsec:Ev_Judging} and Table 10, LLaMA 3.3 70B consistently assigned scores with higher discernment and sensitivity across diverse model outputs, demonstrating both consistency in ranking and meaningful differentiation of clinical quality. Consequently, LLaMA 3.3 70B was selected as our primary evaluator and remained constant throughout all experiments to eliminate evaluator-induced bias.

All generated outputs for open-ended tasks were evaluated using this single, standardized framework. Each output-reference pair was scored using the same prompt-generated similarity rubric to ensure methodological consistency across the entire benchmark. Individual similarity scores $S_i \in [0, 100]$ obtained for each task instance i are then aggregated to compute the final evaluation score by taking the arithmetic mean across all evaluated instances.

Once individual similarity scores \( S_i \in [0, 100] \) are obtained for each task instance \( i \), we compute the \textbf{final evaluation score} by taking the arithmetic mean across all evaluated instances:
\begin{equation}
\text{Final Score} = \frac{1}{N} \sum_{i=1}^{N} S_i
\end{equation}
where \( N \) is the number of datasets used. This final score serves as a robust metric for quantifying the overall content-level alignment between model-generated outputs and expert-authored references across multiple clinical domains.

\subsection{Implementation Details}
We designed PsychiatryBench to ensure fair, consistent, and reproducible evaluation across models. This section outlines key implementation choices, including how prompts were standardized and how model outputs were parsed. These steps were essential for reliable cross-model comparison given the varied response behaviors of modern LLMs.

\subsubsection{Prompting Strategy}

{
All models were evaluated using a standardized zero-shot prompting approach. While the same prompt format was
applied uniformly across all models to ensure fairness in comparison, prompt design was tailored to each specific task
type within PsychiatryBench. For example, diagnostic reasoning tasks employed structured clinical case prompts,
whereas factual knowledge tasks such as Mental QA used direct definition queries. This task-specific prompting allowed
us to preserve the integrity and intent of each evaluation domain while maintaining cross-model consistency.

Across the benchmark, every task was framed as a combination of prompt prediction and prompt evaluation, ensuring
that models were not only generating responses but also being assessed on their ability to interpret and follow clinical
or knowledge-based instructions precisely.

We conducted two dedicated guidance sessions to refine the prompt templates used for each category. These sessions
explored alternative phrasing, instruction clarity, and structural variations. After comparative analysis, we selected
the prompt configuration that produced the highest performance while maintaining conceptual validity. Detailed
descriptions of the guidance sessions, including the candidate prompt variations, are provided in Appendix ~\ref{sec:Prompt_Templates}.

For evaluation, certain tasks employed an LLM-as-a-judge framework to assess answer quality, particularly for open-
ended reasoning or multi-step justification tasks. This approach ensured scalable and consistent scoring while reducing
manual annotation demands. The judging rubric, reliability checks, and model parameters used in this process are
detailed in Appendix ~\ref{sec:Prompt_Templates}.
}

\subsubsection{Parsing Model Outputs}

The process of interpreting and extracting meaningful information from LLM outputs emerged as a critical and non-trivial component of our methodology. While it may appear straightforward for tasks involving simple responses such as multiple-choice questions (MCQs) or numerical ratings the actual output behavior of large models introduced substantial complexity.

Advanced reasoning-oriented models, such as DeepSeek Chat , Gemini Thinking and Sonnet 4.5 Thinking, often diverged from strict output formats. Rather than outputting a clean label or value, they frequently embedded the answer within lengthy justifications or explanations. This behavior required parsing not just the final answer, but also understanding the structure of the response and retaining the embedded reasoning what we refer to as \textit{"printing the reasoning".}

Consequently, our methodology included customized parsers and heuristics tailored to each model’s output style, allowing us to isolate the relevant information while preserving contextual evidence to support qualitative evaluation.

\section{Results \& Discussion}
\label{sec:Results_Discussion}
{ This section presents a multi-faceted analysis of the performance of 15 leading LLMs on the PsychiatryBench benchmark. Our evaluation, summarized in Table~\ref{tab:results}, provides a granular view of each model's capabilities across eleven distinct diagnostic reasoning tasks. Complementing this, Figure~\ref{fig:performance_plot} offers a macroscopic perspective, plotting each model's average performance against its release date, with bubble size representing the model's parameter count. This integrated approach allows for a thorough discussion of overarching performance trends, the impact of model scale, the value of domain-specific training, and the efficacy of different inference strategies.

{

\subsection{Performance of Medical Models}
\label{subsec:Evaluation_of_medical}

In this section, we evaluate the medical models introduced in Section~\ref{medical_models} to establish a clinically meaningful baseline. Table~\ref{tab:med_result} reports the performance of several specialized medical LLMs alongside multiple frontier models. II-Medical-8B attains (76.8\%) on both Management Plan and Mental QA, indicating well-balanced performance across tasks. OpenBioLLM achieves (75.1\%) on Management Plan and a higher (82.2\%) on Mental QA. Med-Palmyra obtains similarly competitive results, with (78.2\%) and (83.2\%), respectively. In contrast, JSL-MedLlama performs substantially worse, scoring (61.0\%) on Management Plan and (65\%) on Mental QA, falling well below the 75\% threshold and therefore being excluded from subsequent specialized analyses. Med-PaLM~2 was also excluded due to deployment constraints that require formal Google approval for medical use cases, which prevented local fine-tuning and evaluation.

The underlined bold entries in Table~\ref{tab:med_result} denote the best-performing medical models within this subset, rather than the best overall scores. This distinction is important because several general-purpose frontier LLMs evaluated elsewhere in the manuscript achieve markedly higher performance on PsychiatryBench, surpassing even the strongest medical models. Accordingly, the underline identifies the top results within the medical-model category only, not the global optimum across all models considered.

We further evaluated the medical language models using two complementary benchmark datasets, Management Plan and Mental QA, each targeting distinct aspects of clinical competence. Together, these datasets yield a balanced assessment of practical clinical decision-making and domain-specific psychiatric knowledge, providing a comprehensive characterization of medical LLM performance.

\begin{table}[!ht]
\caption{Evaluation of medical and frontier LLMs on clinical diagnosis and follow-up tasks using benchmark datasets. The \textbf{underlined bold} values indicate the best score specifically among the listed medical models.}
\label{tab:med_result}
\centering
\begin{tabular}{lcc}
\toprule 
\textbf{Models} & \textbf{Management Plan} & \textbf{Mental QA} \\ 
\midrule
II-Medical-8B & 76.8 & 76.8 \\
OpenBioLLM & 75.1 & 82.2 \\
MedGemma & \ul{\textbf{81.7}} & \ul{\textbf{87.1}} \\ 
Med-Palmyra & 78.2 & 83.2 \\
JSL-MedLlama & 61.0 & 65.0 \\
\midrule

Gemini 2.5 Flash & 81.1 & 83.5 \\ 
Gemini 2.5 Flash (T) & 82.4 & 79.3 \\ 
Deepseek-chat & 83.5 & 82.7 \\
\bottomrule 
\end{tabular}
\end{table}

Among all medical LLMs examined, MedGemma attains the strongest overall results, with (81.7\%) on Management Plan and (87.1\%) on Mental QA. These clinical benchmarks also help contextualize the notably stronger performance of general-purpose frontier LLMs. For example, Deepseek-chat reaches (83.5\%) on Management Plan, and Gemini 2.5 Flash (T) achieves (82.4\%), both outperforming MedGemma. While the medical models exhibit strong biomedical knowledge and structured reasoning, the frontier LLMs demonstrate superior coherence, adaptability, and open-ended reasoning capabilities that are particularly valuable for psychiatric assessment. In our benchmarking, we chose MedGemma to be representative to medical models in the benchmarking against other SoTA LLMs. 

}
\subsection{Overall Performance Landscape and a Clear Trajectory of Improvement}

The results unequivocally demonstrate a stratified performance landscape and a rapid, positive trajectory of advancement in LLM capabilities for diagnostic reasoning. As detailed in Table~\ref{tab:results}, a distinct top tier of models has emerged. GPT 5 Medium (T) stands out as the premier model with the highest average score of (84.5\%), closely followed by Sonnet 4.5 (T) (83.7\%). These models consistently deliver state-of-the-art performance, securing the best or second-best scores across the majority of tasks, particularly those involving complex clinical judgment like Diagnosis, Treatment, and management plan.

Conversely, a clear performance gap separates these leaders from earlier and smaller models. Gemini 2.0 Flash (74.6\%) and GPT-oss (72.3\%) anchor the lower end of the performance spectrum. Figure~\ref{fig:performance_plot} visually articulates this evolutionary trend, revealing a strong positive correlation between model release date and average performance. For instance, average scores improve by over +8 points between Gemini 2.0 Flash (74.6\%) and the Sonnet 4.5 series (83.5\% –83.7\%), and by more than +10 points relative to baselines such as GPT-oss (72.3\%). The jump of +4.0 points from Sonnet 4 (80.5\%) to GPT 5 Medium (T) (84.5\%) in a matter of months underscores the accelerated pace of innovation. This progression suggests that continuous refinements in model architecture, training methodologies, and data curation are yielding substantial and measurable improvements in clinical applicability.

\subsection{Task-Specific Performance Analysis: Identifying Strengths and Persistent Challenges}

A horizontal analysis of Table~\ref{tab:results} reveals consistent patterns in task difficulty across models. Modern LLMs demonstrate remarkable proficiency in tasks that require synthesizing contextual information and generating structured, long-form clinical reasoning. This is most evident in Sequential QA, where Sonnet 4.5 (T) achieved a near-perfect score of (96.2\%). Similarly, strong performance in the Clinical Approach task topped at (90.2\%) by Sonnet 4.5 shows that contemporary models are adept at constructing coherent diagnostic and management pathways from complex psychiatric vignettes.

Conversely, two task categories remain persistent challenges, exposing the limits of current LLM capabilities. First, the Classification of Specific Disorders task proved to be the most difficult across the benchmark, with even top-tier models like GPT 5 Medium (T) achieving only (0.52/45.0) F1-score/subset-accuracy. This reflects the inherent difficulty of multi-label classification in psychiatry, where overlapping symptoms and comorbidities blur categorical boundaries. Second, the Extended Matching Items (EMI) task, which demands discriminating among numerous clinically similar options, also showed broad performance variability. While models like GPT 5 Medium (T) scored (89.1\%), Gemini 2.0 Flash recorded a much lower score of (75.5\%) in this task. Separately, on the Exams task, Gemini 2.0 Flash had one of the lowest results at (69.9\%), underscoring broad difficulties in structured assessment tasks. These findings highlight that while LLMs have mastered structured narrative reasoning, fine-grained clinical discrimination remains a frontier for further optimization.

\subsection{Specialized vs. Generalist Models: The Case of MedGemma}

A compelling narrative within our results is the performance of the domain-specialized model, MedGemma. Despite its moderate size, MedGemma achieves an impressive average score of (78.5\%), placing it on par with large-scale generalist models like Gemini 2.5 Pro (80.2\%) and DeepSeek-R1 (80.4\%). This performance is not uniform; rather, it is concentrated in areas that directly benefit from its specialized training on biomedical and clinical texts.

As highlighted in Table~\ref{tab:results}, MedGemma's distinct advantage is evident in its strong F1/Subset Accuracy scores in the highly granular Classification of Specific Disorders task (0.69/45.0), where it outperforms nearly all generalist counterparts. Furthermore, it achieved one of the highest results in the knowledge-intensive MCQ task (87.4\%), reflecting superior factual retention and domain recall.

However, this specialization comes with a trade-off, as MedGemma was less competitive in broader, open-ended reasoning tasks such as Management Plan (81.7\%) and Sequential QA (87.7\%) compared to top-tier generalist models like Sonnet 4.5 (T) (96.2\%) and GPT 5 Medium (T) (96.1\%). This suggests a dichotomy where generalist models excel at fluid, multi-step reasoning and contextual integration, while specialized models remain superior in precision-driven, knowledge-based clinical classification and recall.

\begin{landscape}
\begin{table}[t]
\caption{Performance comparison of 15 selected LLMs across all PsychiatryBench tasks. Classification tasks are evaluated using F1 Score and Subset Accuracy (F1/A), while other tasks are assessed based on the similarity between model responses and reference answers, as judged by an LLM evaluator. (T) denotes Thinking models. Green, blue, and red highlight the best, second-best, and lowest scores per column, respectively.}

\label{tab:results}
\resizebox{\linewidth}{!}{%
\begin{tabular}{lcccccccccccc|c}
\hline
\multirow{2}{*}{Model} &
\multirow{2}{*}{\shortstack{Diagnosis}} &
\multirow{2}{*}{\shortstack{Treatment}} &
\multirow{2}{*}{\shortstack{Treatment\\Follow-Up}} &
\multicolumn{2}{c}{Classification} &
\multirow{2}{*}{\shortstack{Management\\Plan}} &
\multirow{2}{*}{\shortstack{Clinical\\Approach}} &
\multirow{2}{*}{\shortstack{Mental\\QA}} &
\multirow{2}{*}{\shortstack{Sequential\\QA}} &
\multirow{2}{*}{\shortstack{MCQ}} &
\multirow{2}{*}{\shortstack{Exams}} &
\multirow{2}{*}{\shortstack{EMI}} &
\multirow{2}{*}{\shortstack{AVG.}} \\
\cline{5-6}
& & & & \shortstack{Categories} & \shortstack{Specific} & & & & & & & & \\
\hline
\begin{tabular}[c]{@{}l@{}}DeepSeek -R1\end{tabular} &
  84.2 & 79.9 & 78.7 & 0.72/54.0 & 0.50/44.0 & 84.4 & 90.0 & 88.0 & 89.0 & 83.2 & 79.2 & 85.8 & 80.4  \\
\begin{tabular}[c]{@{}l@{}}DeepSeek -chat\end{tabular} &
  82.7 & 81.9 & 79.4 & 0.63/31.0 & 0.41/23.0 & 83.5 & 90.1 & 82.7 & 88.9 & 75.8 & 71.4 & 79.8 & 76.7 \\
\begin{tabular}[c]{@{}l@{}}Gemini 2.0 Flash\end{tabular} &
  83.1 & 77.4 & 66.5 & 0.63/31.0 & 0.46/31.0 & 78.7 & 86.7 & 82.7 & 90.0 & 75.6 & 69.9 & 75.5 & 74.6 \\
\begin{tabular}[c]{@{}l@{}}Gemini 2.5 Pro\end{tabular} &
  86.0 & 80.5 & 86.5 & 0.60/26.0 & 0.42/31.0 & 84.1 & 89.2 & 87.0 & {\color{darkgreen} 96.9} & 81.7 & {\color{darkgreen} 80.4} & 87.9 & 80.2  \\
\begin{tabular}[c]{@{}l@{}}Gemini 2.5 Flash\end{tabular} &
  78.0 & 81.0 & 80.2 & 0.68/47.0 & 0.44/36.0 & 81.1 & 88.6 & 83.5 & 89.2 & 79.5 & 73.1 & 84.5 & 77.6 \\
\begin{tabular}[c]{@{}l@{}}Gemini 2.5 Flash (T)\end{tabular} &
  85.0 & 82.9 & 82.6 & 0.60/26.0 & 0.37/14.0 & 82.4 & 87.5 & { 79.3} & 89.7 & 79.4 & 78.4 & 86.1 & 77.5 \\
\begin{tabular}[c]{@{}l@{}}Llama 3.3 70B\end{tabular} &
  83.6 & 82.3 & 75.0 & 0.68/46.0 & 0.47/34.0 & 82.1 & 89.7 & 84.2 & { 81.4} & 77.1 & 66.5 & { 72.2} & 75.8 \\
QWQ-32 &
  85.9 & 82.1 & 83.7 & 0.65/38.0 & 0.45/33.0 & 82.9 & 89.2 & 86.0 & 86.2 & 78.4 & 73.4 & 79.8 & 78.1 \\
\begin{tabular}[c]{@{}l@{}}QWEN 3 32B\end{tabular} &
  82.1 & 81.7 & 80.9 & 0.58/25.0 & { 0.35/14.0} & 80.3 & 90.0 & 90.6 & 89.7 & { 74.2} & { 65.3} & 73.1 & 75.1 \\
MedGemma &
  79.4 & 84.2 & 79.2 & { 0.55/34.0} & {\color{darkgreen} 0.69/45.0} & 81.7 & 90.0 & 87.1 & 87.7 & { 87.4} & 69.2 & 72.5 & 78.5 \\
GPT-oss &
  { 70.8} & { 67.8} & { 64.5} & 0.68/53.0 & 0.46/39.0 & { 72.2} & { 73.5} & 86.3 & 93.4 & 74.6 & 72.2 & 78.4 & { 72.3} \\
GPT 5 Medium (T) &
  88.2 & { 88.0} & { 90.0} & {\color{darkgreen} 0.75/59.0} & { 0.52/45.0} & { 87.9} & { 90.1} & { 90.7} & 96.1 & 87.1 & 79.8 & {\color{darkgreen} 89.1} & {\color{darkgreen} 84.5} \\
Sonnet 4 &
  85.8 & 84.5 & 83.5 & 0.69/45.0 & 0.47/34.0 & 84.8 & 90.0 & 89.4 & 89.6 & 81.6 & 77.0 & 83.9 & 80.5 \\
Sonnet 4.5 &
  { 89.6} & 86.6 & 89.3 & 0.70/43.0 & 0.44/30.0 & 87.0 & {\color{darkgreen} 90.2} & 81.8 & 95.6 & 83.0 & 75.6 & 84.9 & 81.5 \\
\begin{tabular}[c]{@{}l@{}}Sonnet 4.5 (T)\end{tabular} &
  {\color{darkgreen} 89.9} & {\color{darkgreen} 88.4} & {\color{darkgreen} 90.4} & { 0.73/55.0} & 0.40/33.0 & {\color{darkgreen} 88.6} & 87.5 & {\color{darkgreen} 92.5} & { 96.2} & {\color{darkgreen} 89.2} & { 80.0} & { 88.9} & { 83.7} \\ \hline
\end{tabular}%
}
\end{table}
\end{landscape}

\begin{figure}[!ht]
    \centering
    \includegraphics[width=1\linewidth]{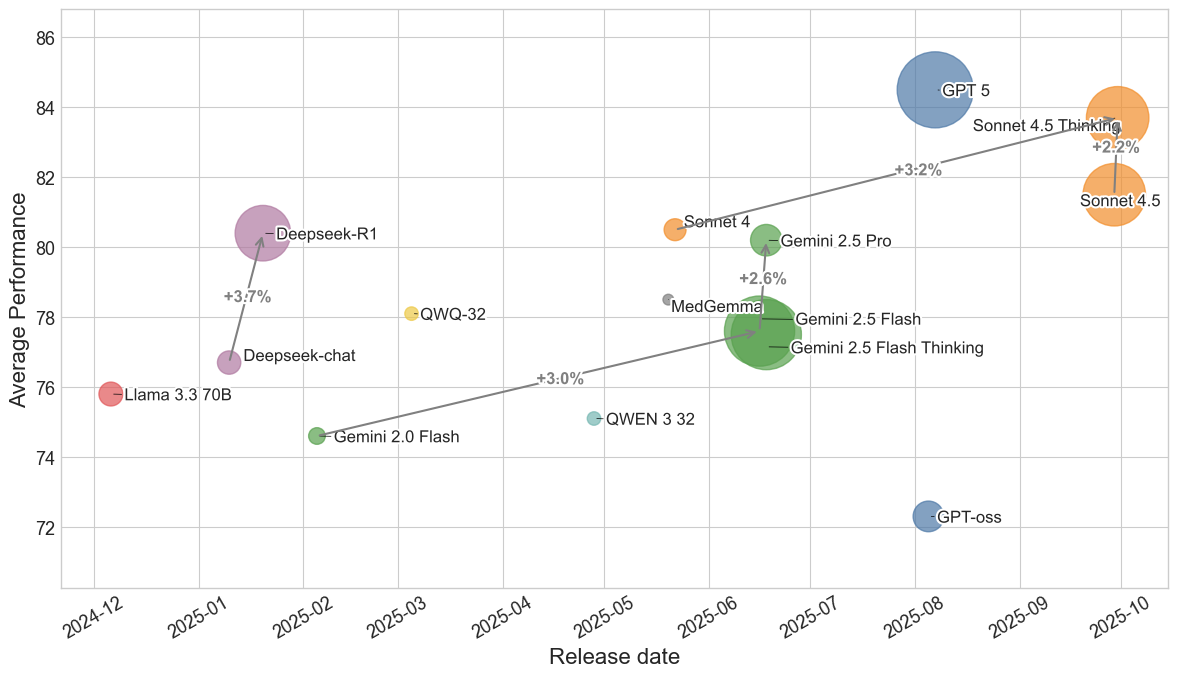}
    \caption{Each bubble shows a model, positioned by release date and average performance. Bubble size denotes parameter count, and color represents model family. Larger, newer models generally achieve higher performance.}
    \label{fig:performance_plot}
\end{figure}

\subsection{The Role of Inference Strategies: "Thinking" vs. Standard Modes}

Our study's inclusion of "Thinking" variants for the Gemini 2.5 Flash and Sonnet 4.5 models reveals that the benefit of more deliberative, multi-step inference is highly architecture-dependent. For the Anthropic models, this strategy yielded a significant performance dividend. Sonnet 4.5 (T) not only surpassed its standard counterpart with an average score of (83.7\%) versus (81.5\%) but also achieved the highest performance across several of the most cognitively demanding tasks: Diagnosis (89.9\%), Treatment (88.4\%), Treatment Follow-Up (90.4\%), and Mental QA (92.5\%). This indicates that for the Sonnet architecture, the “Thinking” mode effectively activates deeper and more structured clinical reasoning pathways.

In contrast, this advantage was not evident for the Google Gemini models. The average performance of Gemini 2.5 Flash (T) (77.5\%) was negligibly lower than that of its standard counterpart, Gemini 2.5 Flash (77.6\%). Moreover, the 'Thinking' variant's performance was mixed: it underperformed in Mental QA (79.3\%) compared to the standard model's (83.5\%), but it scored higher in Diagnosis (85.0\%) versus the standard model's (78.0\%), indicating an inconsistent benefit from the additional inference step.

\subsection{The Impact of Model Scale and Release Date}

The scatter plot in Figure~\ref{fig:performance_plot} provides critical insights into the relationship between model size (parameter count), recency, and overall performance. A clear upward trend emerges: larger models depicted by larger bubbles tend to cluster in the higher performance range. The leading results achieved by Sonnet 4.5 (T) and GPT 5 Medium (T), with average scores of (83.7\%) and (84.5\%) respectively, exemplify how scale combined with architectural refinement drives superior clinical reasoning outcomes.

However, size alone does not guarantee success. The performance of DeepSeek-R1, a large-scale model with an average score of  (80.4\%), remains strong yet is surpassed by more recent and structurally optimized models like Sonnet 4.5 (T) and GPT 5 Medium (T). This pattern underscores that architectural innovation, training corpus diversity, and post-alignment strategies can yield greater performance gains than mere parameter expansion.

The overall trajectory visible in Figure~\ref{fig:performance_plot} confirms this trend: newer architectures are consistently leveraging scale more effectively, converting size into meaningful improvements in reasoning accuracy, clinical adaptability, and task generalization.

{
\subsection{Model Family Performance}
Our analysis reveals distinct performance signatures across model families, reflecting different architectural priorities and the effectiveness of their training. The frontier generalist models from Anthropic and OpenAI set the benchmark, with Sonnet 4.5 (T) (83.7\%) and GPT-5 Medium (T) (84.5\%) demonstrating state-of-the-art capabilities driven by recent architectural innovations.

In contrast, other families highlight key limitations. The Gemini models, while strong with Gemini 2.5 Pro at 80.2\%, show that advanced inference modes are not universally beneficial; its 'Thinking' variants yielded inconsistent gains, unlike the significant boosts seen in the Sonnet series. Similarly, DeepSeek-R1 (80.4\%) illustrates that massive scale alone does not guarantee leadership, as it is surpassed by more structurally optimized models.

Finally, specialized medical models like MedGemma (78.5\%) confirm a classic trade-off: their domain-specific pretraining provides a measurable edge in knowledge-intensive tasks but limits the cognitive flexibility needed for nuanced psychiatric reasoning. These patterns confirm that the frontier of performance is now defined more by sophisticated architectural design and alignment than by raw parameter count alone.

\subsection{Cross-Task Consistency and Stability Analysis}
\label{subsec:cross_task_consistency}

Beyond evaluating raw performance on individual tasks, an important dimension of Model reliability is the degree to which (LLMs) maintain stable behavior across the diverse task categories within PsychiatryBench. Given that the benchmark spans eleven clinically heterogeneous tasks ranging from structured classification to open-ended sequential reasoning cross-task consistency offers a deeper view of each model’s robustness, generalizability, and susceptibility to task-specific variability.

An inspection of the scores in Table ~\ref{tab:results}  reveals substantial differences in cross-task stability among models. General-purpose frontier models such as Sonnet 4.5 (T) and GPT 5 Medium (T) demonstrate the highest degree of uniformity, maintaining strong performance across both highly structured tasks (e.g., classification of specific disorders) and open-ended 
reasoning tasks (e.g., sequential QA and clinical approach). This stability 
reflects the broad generalization capabilities of recent frontier models, 
supported by large-scale heterogeneous training data and refined inference 
strategies that preserve coherence across task formats. These models exhibit 
narrow performance fluctuations between their strongest and weakest tasks, 
indicating that their reasoning processes remain stable regardless of prompt 
structure or clinical domain.

In contrast, specialized medical models such as MedGemmathe  and 
Med\_Palmyra display a more uneven performance profile. Their strengths 
are concentrated in knowledge-intensive tasks such as mental QA and the 
classification of specific disorders, where domain-specific pretraining yields 
clear advantages. However, these models are noticeably less consistent on tasks 
requiring multi-step reasoning, contextual linking, or flexible narrative 
generation, such as sequential QA and clinical approach. The resulting 
performance gaps highlight a specialization-versus-generalization trade-off: 
while medical models excel when tasks directly match their training distribution, 
They are more sensitive to variations in clinical reasoning format or task 
structure.

Lower-performing models, exemplified by JSL\_MedLlama, exhibit the 
widest instability across tasks. Large drops in performance between knowledge 
tasks and contextual reasoning tasks suggest that these models struggle to 
maintain coherent reasoning pipelines when task demands shift. Such volatility 
reduces their suitability for clinical applications where stability across 
problem types is a prerequisite for reliability.

Overall, the cross-task analysis underscores an important conclusion: 
the strongest models are not those that achieve isolated peaks on 
specific tasks, but those that maintain consistent performance across the full 
spectrum of psychiatric reasoning challenges. Frontier generalist models achieve 
the highest stability, while specialized models provide strong but localized 
competence. This distinction reinforces the need for multifaceted evaluation 
frameworks such as PsychiatryBench, which are capable of exposing not just peak 
performance but also the breadth and reliability of clinical reasoning 
capabilities.

}

{
\subsection{Extended Matching Item (EMI): Full vs.\ Separated Evaluation}
\label{subsec:Emi-full-vs-separated}

Based on the overall EMI-Full scores in Table~\ref{tab:results}, we restrict our format comparison to the five best-performing models on the EMI tasks: DeepSeek-R1, Gemini 2.5 Pro, Gemini 2.5 Flash (T), Sonnet 4.5 (T), and GPT-5. For these models, we compare performance on the original extended-set format (EMI-Full) and a Separated evaluation in which each subquestion is scored independently. To ensure a fair item-by-item comparison, all metrics are computed on the overlapping subset of \(N = 144\) EMI sets (\(T = 514\) subquestions) that are structurally compatible across formats, with responses case-normalised (e.g., ''a''/''A'') and scored under the same Answer Integrity Rule. A detailed description of this overlapping subset and the scoring procedure is provided in Appendix~\ref{subsec:Emi}.

Across this overlapping subset, the full-format results confirm that the selected models are already strong on EMI-Full. On the full 277-set benchmark, average PCS lies in a narrow band (approximately (85.8\%)--(89.1)\%), and when restricted to the 144 overlapping sets, full-format PCS remains high with a similar ordering: DeepSeek-R1 is clearly weaker at (72.0\%), whereas Gemini 2.5 Flash (T) and Sonnet 4.5 (T) attain mid-70s to mid-80s PCS, and Gemini 2.5 Pro and GPT-5 reach (87.5\%) and (87.2\%), respectively. Gemini 2.5 Pro thus provides the strongest full-format performance on the overlapping subset, closely followed by GPT-5 and Sonnet 4.5 (T). Under the Separated evaluation, subquestion-level accuracy typically falls in the low-to-high (80\%) range, with DeepSeek-R1 improving from (72.0\%) PCS (full) to (83.6\%) accuracy (separated), and Gemini 2.5 Flash (T) increasing from (75.7\%) PCS to (88.1\%) accuracy. In contrast, Gemini 2.5 Pro shifts only slightly from 87.5\% PCS to (86.0\%) accuracy, and GPT-5 from (87.2\% to 87.8\%), indicating that these models are already near their ceiling in the full format. Overall, Gemini 2.5 Pro achieves the highest full-format PCS on the overlapping subset, while Gemini 2.5 Flash (T) attains the highest separated accuracy (88.1\%) and separated PCS (89.0\%).

The consistency metrics summarised in Appendix~\ref{subsec:Emi} provide additional insight into how stable each model’s behaviour is across formats. Overall Consistency is highest for Gemini 2.5 Pro (95.7\%) and GPT-5 (93.8\%), with correspondingly low Format Divergence values of 0.043 and 0.062, indicating that these models change their correctness status between formats on only about 4--6\% of overlapping subquestions. Gemini 2.5 Flash (T) and Sonnet 4.5 (T) achieve similarly strong accuracies but show somewhat higher divergence (around 0.13 and 0.20), suggesting more item-level sensitivity to format changes, while DeepSeek-R1 exhibits the lowest consistency (78.3\%) and highest divergence (0.217) among the five models. Taken together, these score patterns indicate that the top EMI models identified in Table~\ref{tab:results} generalise well across full-set and separated-subquestion formats, with Gemini 2.5 Pro and GPT-5 appearing particularly robust, and Gemini 2.5 Flash (T) benefiting the most from decomposing Extended Matching Items into separated subquestions.

}

\section{Limitations and Future Work}
\label{sec:Limitations_and_Future_work}
Although this study offers valuable information on the capabilities of LLMs for mental health tasks, it is important to recognize several limitations that may affect the interpretation and generalizability of our findings. In addition to the future directions that can help achieve higher performance.

\subsection{Limitations}

While PsychiatryBench represents a significant step towards the clinically grounded evaluation of LLMs in psychiatry, its development and scope were shaped by several inherent limitations. These constraints, which our work seeks to begin addressing, stem primarily from the nature of the available source material and the practicalities of evaluation at scale.

\subsubsection{Data Source and Textbook Limitations}

The foremost limitation is that our reliance on published textbooks introduces a trade-off between authoritative knowledge and real-world clinical practice. The vignettes in these resources are, by design, well-structured and pedagogically focused to illustrate specific diagnostic principles. They may not fully capture the ambiguity, noise, and unstructured nature of raw clinical data, such as real-time patient dialogues or fragmented electronic health record (EHR) notes. Consequently, while PsychiatryBench robustly tests an LLM's reasoning against expert-curated knowledge, performance may not perfectly generalize to the complexities of live, unfiltered clinical interactions. 

A related point is that the distribution of psychiatric disorders within PsychiatryBench does not mirror their real-world epidemiological prevalence. This is a direct consequence of our source materials; psychiatric textbooks often over-represent rare, complex, or "classic" cases to serve specific pedagogical goals. Therefore, models evaluated on PsychiatryBench are tested on their ability to handle a teaching-oriented distribution of cases, which differs from the case mix they would encounter in a general clinical setting. Additionally, PsychiatryBench's case distribution emphasizes moderate-acuity outpatient presentations, with limited representation of acute crisis scenarios such as severe psychosis requiring emergency intervention, active suicidal intent with plan and means, or manic episodes necessitating involuntary hospitalization.

\subsubsection{Western Psychiatric Bias}

A significant limitation of PsychiatryBench is its reliance on predominantly Western psychiatric sources, particularly American textbooks such as DSM-5-TR Clinical Cases and Stahl's Essential Psychopharmacology. While these resources represent gold-standard, expert-validated knowledge within Western psychiatric practice, they inherently reflect Western conceptualizations of mental illness, symptom presentation, diagnostic categories, and treatment approaches. Psychiatric presentations, the validity of diagnostic constructs, and the appropriateness of treatment recommendations can vary significantly across cultures. Help-seeking behaviors, symptom expression, and therapeutic expectations are all influenced by cultural context, yet our current benchmark evaluates LLMs primarily on their alignment with Western psychiatric frameworks. Consequently, strong performance on PsychiatryBench does not necessarily indicate an LLM's readiness for deployment in culturally diverse or non-Western clinical settings.

\subsubsection{Evaluation Methodology Limitations}

Our evaluation methodology for open-ended tasks relies on an LLM-as-a-judge framework. While this approach offers the scalability and consistency required to evaluate thousands of responses across numerous models, it is not a perfect substitute for human expert evaluation. LLM judges can exhibit inherent biases and may not capture the full clinical nuance apparent to a practicing psychiatrist. A significant methodological limitation concerns the potential circularity of our approach: the model selected as the evaluator (Llama 3 70B) was also included as one of the models under evaluation in our test set. We mitigated this by grounding the evaluation in expert-validated reference answers, meaning the judge’s task was one of comparing outputs to a “gold standard” rather than making de novo clinical judgments.
However, a full-scale validation by a panel of human psychiatrists was beyond the scope of this work and remains a crucial future direction. LLM-as-judge evaluation also introduces potential bias due to overlapping reasoning frameworks. In particular, when Llama 3 70B serves as both evaluator and evaluated model, a direct circularity arises that may artificially inflate agreement metrics, representing an acknowledged methodological limitation.
Strong performance on the expert-validated psychiatric knowledge and reasoning tasks used by PsychiatryBench does not necessarily indicate clinical preparedness. Additional validation, such as prospective clinical trials, evaluations of conversational coherence in patient contacts, assessments of cultural competence across varied groups, and strict safety monitoring in real-world clinical settings, would be necessary for real-world clinical deployment.

A further limitation is the unavoidable risk of training data contamination. The authoritative, publicly available textbooks and casebooks used to construct PsychiatryBench are likely included in the pre-training corpora for many of the large-scale models evaluated. This may inflate performance scores, particularly on tasks measuring factual recall or knowledge recognition. We believe the benchmark's value is nonetheless retained, as its primary focus is on novel task formats that test the application of knowledge and multi-step clinical reasoning rather than simple rote memorization.

\subsubsection{Safety Limitations}
{
The current evaluation framework does not systematically assess several safety-critical behaviors essential for the responsible use of LLMs in psychiatric contexts. Models may inadvertently reinforce delusional beliefs by favoring conversational coherence over clinical accuracy, thereby exacerbating pathological thinking. For instance, LLMs may also mishandle suicidal ideation by failing to recognize acute risk or escalate appropriately, a shortcoming with potentially life-threatening consequences. Additional concerns include the tendency toward anthropomorphism using language such as “I’m here for you,” which can blur therapeutic boundaries and foster unhealthy attachments, as well as sycophantic responses that prioritize user satisfaction over evidence-based reasoning, potentially minimizing symptoms or validating unsafe preferences. These behaviors reflect forms of value misalignment that extend far beyond factual correctness and are not captured by our current LLM-as-a-judge scoring approach, which focuses primarily on semantic agreement with expert references. As a result, clinicians and ethicists have raised concerns that reliance on opaque, uncalibrated systems may erode clinical judgment, amplify existing biases, and undermine fairness in mental healthcare, underscoring the urgent need for evaluation frameworks that directly interrogate safety, alignment, and boundary-maintaining behavior rather than indirect proxies of correctness.

}

\subsubsection{Additional Limitations}

Additional limitations include: the variable representation of medical mimics (e.g., thyroid disease, delirium, CNS pathology) that was not systematically designed to test for medical rule-outs; the intentional limitation to adult psychiatry, excluding child and adolescent psychiatry which represents a distinct subspecialty; and the primary focus on diagnostic reasoning and treatment planning, with limited assessment of broader integrative skills such as comprehensive biopsychosocial formulation, functional assessments, or multidisciplinary collaboration planning.

These limitations highlight the very challenges our work was designed to confront. The constraints on scale, data distribution, and evaluation methods are direct consequences of the field's lack of high-fidelity psychiatric benchmarks. By navigating these issues and creating a transparent, clinically-grounded resource, PsychiatryBench establishes an essential baseline. It serves not only as an evaluation tool but as a foundational scaffold upon which future research can build to address these very limitations, paving the way for safer and more reliable AI in mental healthcare.

\subsection{Future Work}
\label{sec:futurework}

Future work should build directly on the limitations identified above and focus on strengthening the clinical realism, cultural breadth, methodological rigor, and safety of LLM evaluation in psychiatry. These directions aim to advance the development of more comprehensive, reliable, and clinically defensible AI systems that reflect the complexities of real-world psychiatric practice.

\subsubsection{Expanding Clinical Data Representativeness}

Addressing the limitations associated with textbook-derived vignettes requires the incorporation of more diverse and ecologically valid clinical data sources. Future versions of PsychiatryBench should integrate de-identified electronic health records, naturalistic patient–clinician dialogues, and real-time clinical notes that capture the ambiguity, noise, and fragmentation characteristic of authentic psychiatric encounters. This expansion would allow for evaluation settings that more closely approximate the unpredictability and variability of real-world practice. Furthermore, the dataset should broaden its diagnostic coverage to better align with real-world epidemiology by including a larger proportion of highly prevalent disorders, as well as more medically complex presentations. Enriching the dataset with emergency and crisis-level scenarios such as active suicidal intent, severe psychosis, or acute mania requiring involuntary intervention would further enhance ecological validity. Scalable data curation pipelines that combine automated extraction tools with expert psychiatric oversight would facilitate the responsible and accurate expansion of clinically representative cases.

\subsubsection{Improving Cross-Cultural Validity}

To address the Western psychiatric bias inherent in the current benchmark, future work should incorporate case materials and diagnostic frameworks from a broader range of cultural and geographic contexts. Integrating guidelines and casebooks from Asia, Africa, Latin America, and the Middle East would promote greater cultural representativeness and reduce reliance on Western-centric models of mental illness. Expanding the benchmark to include culturally variable idioms of distress, distinct help-seeking patterns, and diverse symptom presentations would help assess whether models can avoid applying ethnocentric assumptions or misinterpreting culturally mediated clinical information. Collaborations with international psychiatric organizations and culturally diverse clinical institutions will be essential for acquiring representative source material. Additionally, the development of evaluation metrics that explicitly assess cultural competence would allow for systematic examination of whether models accurately and sensitively interpret psychiatric presentations across varying cultural contexts.

\subsubsection{Improving Evaluation Methodology and Reducing Model Circularity}

Future methodological work should focus on strengthening the reliability and validity of LLM-as-a-judge evaluation frameworks. Incorporating panels of licensed psychiatrists for periodic benchmarking would provide a clinical anchor point for evaluating whether automated scoring methods align with expert judgment. Establishing adjudication systems that combine human oversight with LLM-based evaluators may further reduce the biases associated with relying on a single model for judgment. To mitigate circularity, future iterations of the benchmark should ensure strict separation between the models serving as evaluators and those being evaluated, potentially through the development of judge models specifically designed for assessment tasks. Enhancing the evaluation methodology to include step-by-step reasoning audits, diagnostic chain-of-thought analyses, and temporal generalization tests based on post–knowledge cutoff data would offer a more robust assessment of clinical reasoning capabilities. Further research is also needed to quantify and mitigate the effects of training data contamination, including temporal ablation analyses and the use of synthetic or newly developed case vignettes that can test true generalization rather than memorization.

\subsubsection{Safety-Critical Evaluation and Value Alignment}

Advancing the safety evaluation of LLMs is essential for their use in psychiatric contexts. Future work should develop comprehensive evaluation tasks that explicitly test models’ handling of clinically sensitive scenarios. These tasks must assess whether models can avoid reinforcing delusional content, appropriately recognize and respond to suicidal ideation, maintain clear therapeutic boundaries without anthropomorphic self-representation, and prioritize clinical accuracy over user agreement in situations where user preferences conflict with evidence-based practice. Developing evaluation metrics that systematically examine these behaviors will enable a more rigorous assessment of safety vulnerabilities that extend beyond factual correctness. Long-term directions should include the integration of clinically informed red-teaming efforts, formal ethical oversight, and the creation of value alignment benchmarks that test whether models can recognize the limits of their competence, decline unsafe tasks, and default to risk-minimizing behaviors during ambiguous or high-risk situations. Strengthening safety evaluation in this manner will be critical for ensuring that LLMs support rather than undermine clinician judgment and patient well-being.

\subsubsection{Broadening Clinical Scope and Integrative Task Coverage}

Future expansions of PsychiatryBench should extend its clinical breadth to encompass domains currently underrepresented or absent. Incorporating cases involving medical mimics such as endocrinological disorders, neurological disease, and delirium would enable evaluation of integrated medical–psychiatric reasoning, which is essential in many real-world settings. Further work should introduce a comprehensive suite of cases from child and adolescent psychiatry, a subspecialty with distinct diagnostic frameworks, developmental considerations, and contextual factors. Beyond diagnosis and treatment planning, the benchmark should also evaluate broader integrative capabilities, including biopsychosocial formulation, functional assessment, and the capacity to incorporate psychosocial stressors into clinical reasoning. Evaluating longitudinal care abilities, such as tracking symptom evolution, anticipating relapse risk, and adjusting treatment plans over time, would align more closely with the realities of ongoing psychiatric care. Incorporating these dimensions will allow future versions of PsychiatryBench to capture the full complexity of psychiatric practice and provide a more comprehensive foundation for evaluating LLMs intended for clinical use.

\section{Conclusion} 
\label{sec:Conclusion}
{
PsychiatryBench represents a significant step toward establishing clinically grounded, domain-specific standards for evaluating LLMs in mental healthcare. By integrating thousands of expert-curated QA pairs derived from authoritative psychiatric texts, alongside complex multi-step clinical vignettes, the benchmark captures a broad spectrum of diagnostic, therapeutic, and longitudinal reasoning skills. Our analysis reveals that advanced models can demonstrate strong psychiatric knowledge and generate coherent clinical interpretations, yet these strengths are unevenly distributed across task types. Models continue to struggle with ambiguity, multi-label diagnostic reasoning, context-dependent interpretation, and scenarios requiring high-fidelity clinical judgment. These limitations highlight the risks of deploying general-purpose models in sensitive psychiatric settings without rigorous, domain-tailored evaluation. PsychiatryBench thus fills a critical gap by providing a structured, transparent, and reproducible framework for assessing the depth and reliability of psychiatric reasoning in contemporary LLMs.

At the same time, this benchmark serves as a foundation for future innovations in AI-driven mental healthcare. Improving the cultural breadth, ecological validity, and modality diversity of PsychiatricBench will be essential to creating more comprehensive assessments that reflect the realities of global clinical practice. Equally important will be the development of safety-oriented evaluations that specifically test crisis response, boundary maintenance, hallucination robustness, and value alignment dimensions that are central to real-world psychiatric care but absent from most current benchmarks. Continued interdisciplinary collaboration between psychiatrists, clinical researchers, computational scientists, and ethicists will be necessary to ensure that future models not only perform well on structured tasks but also adhere to the ethical and safety standards expected in mental health contexts. By establishing a strong evaluative foundation and outlining clear directions for progress, PsychiatryBench aims to catalyze the creation of AI systems that meaningfully support clinicians, enhance patient care, and advance the safe and equitable integration of AI in psychiatric settings.

}

\color{black}

\section{Ethical Statement, Copyright, and Fair Use Statement}
\label{copyright}
The PsychiatryBench dataset was developed using content manually curated from publicly available, expert-authored psychiatric textbooks and casebooks. All material was selected and annotated by subject matter experts with a clinical background to support research in safe and effective psychiatric reasoning by AI models. No personal, identifiable, or patient-derived data were used in the construction of this dataset. The dataset does not contain any clinical notes, transcripts, or sensitive medical information from real individuals.

This research adheres to ethical guidelines regarding the use of educational content for research purposes. The dataset is intended solely for non-commercial research, academic, and educational uses that aim to improve the safety, reliability, and clinical alignment of language models in mental health applications. All references to psychiatric diagnoses, treatments, and case scenarios are drawn from simulated or publicly available educational materials, not real patient interactions.

All content in the PsychiatryBench dataset is derived from publicly available and widely used psychiatric educational resources for academic research and benchmarking. The inclusion of questions and scenarios from books such as DSM-5-TR Clinical Cases, Stahl’s Psychopharmacology, and others is done under the principle of fair use, specifically for:

\begin{itemize}[leftmargin=0.25in] 
    \item Transformative use: The material has been substantially modified, structured, and reformatted into a new dataset format for the distinct purpose of evaluating AI systems.
    \item Non-commercial research: The dataset is intended solely for research and academic purposes, without any commercial exploitation.
    \item Limited scope and effect on market: Only selected, representative samples were used, not the entirety of any single source, and use of the dataset does not substitute for or compete with the original textbooks.
\end{itemize}

Users of the dataset are responsible for ensuring compliance with local copyright laws. They should obtain appropriate permissions if intending to reuse substantial parts of the original material for commercial or distribution purposes.

\section*{Acknowledgements}
The authors thank Abdelrahaman Ali for his help in data collection and for doing some evaluations.

\section*{Declarations}
\subsection*{Author Contributions}
A. E. F. : Conceptualization; Methodology; Benchmark framework development; Data collection and curation; Drafting the manuscript.
A. A. H. :  Benchmark framework development; Data collection and curation; Drafting the manuscript.
R. J. H.: Supervision; Psychiatry domain expertise; Design of diagnostic and therapeutic tasks; Validation of benchmark tasks; Writing review \& editing.
M. E. F.: Conceptualization; Methodology;  Supervision; Project administration; Funding acquisition; Writing review \& editing; Final approval of the manuscript. All authors reviewed the final manuscript, provided critical feedback, and approved the submission.

\subsection*{Funding}
This study received no funding.

\subsection*{Conflict of interest}
The authors declare no conflict of interest.

\subsection*{Data Availability}
All data supporting the findings of this study are provided within the article and its Supplementary Information files. 

\subsection*{Code Availability}
The code used for the analysis in this study is available upon reasonable request.

\bibliographystyle{unsrt}  
\bibliography{references}  

\newpage
\setcounter{section}{0}
\setcounter{figure}{0}
\setcounter{table}{0}
\section{Appendix}
\label{sec:Appendix}
{\

\subsection{Extended Matching Item and Separated Extended Matching Item}
\label{subsec:Emi}

A methodological point worth noting is how we handled EMI items that appeared in both the full-set format and the separated-subquestion format. Although the original EMI dataset contained 277 extended matching sets, not all of these sets were represented in a fully comparable way across the two evaluation formats. In practice, some EMI sets appeared only in the full-format runs or only in the separated runs, and in a few cases the subquestion structure or answer keys were not perfectly aligned between formats. In addition, response options in the model outputs sometimes differed only in letter case (e.g., ''a'' vs.\ ''A''); to avoid spurious mismatches, all responses and answer keys were case-normalized before scoring. To avoid introducing bias from missing, structurally mismatched, or trivially misaligned items, we therefore restricted our consistency and format-sensitivity analysis to the subset of EMI sets that were present and structurally compatible in both conditions. This yielded a shared subset of \(N = 144\) EMI sets with a total of \(T = 514\) overlapping subquestions, for which we could compute correctness in the full format (\(F_i\)) and the separated format (\(S_i\)) using the same answer key and the same Answer Integrity Rule.

Focusing on this overlapping subset has two important implications. First, it strengthens the internal validity of the comparison between formats: every metric Total Correct (Full / Separated), Common Correct, Common Incorrect, Total Incorrect (Full / Separated), Average Partial Credit Scores (PCS) in both formats, Accuracy (Separated), Overall Consistency, Format Agreement Rate, and Format Divergence Rate is computed on exactly the same pool of underlying subquestions (Tables~\ref{tab:EMI_P2} and~\ref{tab:EMI_P1}). This ensures that any observed differences are attributable to differences in presentation format and model behavior, rather than to differences in item composition across conditions. Second, this restriction necessarily means that our format-comparison results should be interpreted as applying to the shared core of the EMI dataset rather than to all 277 sets. The excluded items are not assumed to be easier or harder, but they fall outside the strict intersection required for a one-to-one comparison. Overall, this design choice reflects a trade-off between breadth and fairness: by sacrificing some coverage of the original EMI pool, we obtain a cleaner and more conservative estimate of how stable each model's performance is across formats.

A second methodological choice concerns the selection of models included in the comparative analysis. We initially evaluated a larger pool of 15 models on the EMI-Full setup and then restricted our detailed consistency and format-sensitivity analysis to the top-performing systems on this task: GPT-5, Sonnet 4.5 (T), Gemini 2.5 Flash (T), Gemini 2.5 Pro, Gemini 2.5 (T), and DeepSeek-R1. This decision reflects both practical and scientific considerations. From a practical standpoint, these are precisely the models that are most likely to be considered for deployment in educational or clinical assessment contexts; understanding their robustness to format changes is therefore the most consequential. From a scientific standpoint, concentrating on the strongest systems allows us to probe whether high aggregate performance on EMI-Full (as quantified by PCS over all 277 sets) necessarily translates into stable behavior under the more granular Separated evaluation. The resulting comparison is thus not intended as an exhaustive benchmark of all available models but as a targeted analysis of the current leading systems under a unified evaluation framework (PCS-based EMI-Full scores and subquestion-level Separated scores on the 144-set overlap.

The numerical results for this overlapping subset, as reported in Tables~\ref{tab:EMI_P2} and~\ref{tab:EMI_P1}, reveal several notable patterns. On the original 277-set EMI-Full benchmark, average PCS is already high for all selected models, ranging from 85.8\% for DeepSeek-R1 up to 89.1\% for GPT-5, with intermediate values for Gemini 2.5 Flash (86.1\%), Gemini 2.5 Pro (87.9\%), and Sonnet 4.5 (T) (88.9\%). When we restrict attention to the 144 overlapping sets, the same ordering largely persists: DeepSeek-R1 drops to (72.0\%) PCS in the full format on this subset, whereas Gemini 2.5 Flash (T) reaches (75.7\%), Sonnet 4.5 (T) (84.7\%)--85.8\%), Gemini 2.5 Pro (87.5\%), and GPT-5 (87.2\%). This pattern suggests that the overlapping subset is broadly representative of the difficulty profile of the full EMI pool, rather than being skewed toward unusually easy or hard items: the strongest models remain strong, and the weaker one (DeepSeek-R1) still lags behind in full-format PCS.

Second, direct comparison of the full and separated formats shows that top models are generally robust to the shift from extended matching sets to item-by-item subquestions, but not completely insensitive to it. In the separated condition, subquestion-level accuracy on the overlapping subset ranges from (83.6\%) for DeepSeek-R1, through (86.0\%) for Gemini 2.5 Pro and (87.8\%) for GPT-5, up to (88.1\%) for Gemini 2.5 Flash (T), with Sonnet 4.5 (T) in the mid-80s (around 86\%). Average PCS in the separated format follows a similar pattern: (84.9\%) for DeepSeek-R1, (86.2\%) for Gemini 2.5 Pro, (87.8\%) for GPT-5, and (89.0\%) for Gemini 2.5 Flash (T), again with Sonnet 4.5 (T) very close to the top tier. For some models (notably DeepSeek-R1 and Gemini 2.5 Flash), separated accuracy and PCS are slightly higher than their full-format counterparts on the same overlapping items, suggesting that decomposing complex EMIs into atomic subquestions can help format-sensitive models avoid cross-item interference and better exploit local item cues. For others (Gemini 2.5 Pro, GPT-5, and Sonnet 4.5 (T)), separated performance is extremely close to full-format performance on the overlap, with only modest gains or small trade-offs, consistent with a more stable internal representation of the underlying medical knowledge.

Third, the consistency-oriented metrics make it possible to distinguish genuinely format-stable models from those that merely achieve similar aggregate scores by different patterns of errors. Overall Consistency defined as the proportion of overlapping subquestions that are either correct in both formats or incorrect in both, varies substantially across models. DeepSeek-R1 shows the lowest consistency at (78.3\%), whereas Gemini 2.5 Flash (T) achieves (86.7\%), Sonnet 4.5 (T) around (80\%--92\%), Gemini 2.5 Pro (95.7\%), and GPT-5 (93.8\%). The corresponding Format Divergence Rates span almost an order of magnitude: 0.217 for DeepSeek-R1, 0.133 for Gemini 2.5 Flash (T), around 0.20 for Sonnet 4.5 (T), 0.062 for GPT-5, and only 0.043 for Gemini 2.5 Pro. In other words, Gemini 2.5 Pro and GPT-5 change their correctness status between formats on only about 4--6\% of overlapping subquestions, while DeepSeek-R1 does so on more than 20\%. Format Agreement measured at the level of the chosen option letter, irrespective of correctness is high across the board (from (85.4\%) for Gemini 2.5 Flash (T) up to (95.2\%) for Gemini 2.5 Pro), but the gap between agreement and correctness-based consistency reveals how often models ``agree with themselves'' on a wrong answer versus maintaining both correctness and choice across formats.

Taken together, these findings suggest that strong performance on EMI-Full does not guarantee complete robustness to changes in question format. However, the leading models, especially Gemini 2.5 Pro and GPT-5, display a relatively high degree of format invariance, combining high full-format PCS (87.5\%--89.1\%) with strong separated accuracies (86.0\%--87.8\%) and very low divergence (0.043--0.062). From an assessment perspective, this is encouraging: it implies that for top-tier models, extended matching sets and separated-item formats may be used somewhat interchangeably without dramatically altering performance profiles, provided that scoring rules and answer integrity constraints are held fixed. At the same time, the non-trivial divergence observed for models such as DeepSeek-R1 (21.7\% divergence) and, to a lesser extent, Gemini 2.5 Flash (T) and Sonnet 4.5 (T), highlights the importance of specifying and standardising evaluation formats when comparing models or using them in educational settings.

Finally, these analyses underscore the value of designing evaluation protocols that explicitly test format sensitivity, rather than relying solely on aggregate scores in a single presentation style. Future work could extend the present approach in several directions: by incorporating a broader range of models (including weaker baselines and smaller open-source systems), by exploring additional EMI transformations (e.g., varying distractor sets, randomising option labels, or altering clinical context), and by examining how format sensitivity interacts with domain shifts and language variation. Such extensions would provide a more complete picture of how large language models internalise and apply domain knowledge and of the conditions under which their apparent competence on structured medical examinations remains stable or begins to fracture under seemingly minor changes in task framing.

\begin{table}[]
\caption{Comparison of EMI full-set vs. separated formats for top models (144 EMI sets, 452 subquestions).}
\label{tab:EMI_P2}
\begin{tabular}{ccccccc}
\hline
Models / Data        & \begin{tabular}[c]{@{}c@{}}Total Correct \\ (Full)\end{tabular} & \begin{tabular}[c]{@{}c@{}}Total Correct \\ (Separated)\end{tabular} & \begin{tabular}[c]{@{}c@{}}Correct \\ in Both\end{tabular} & \begin{tabular}[c]{@{}c@{}}Incorrect \\ in Both\end{tabular} & \begin{tabular}[c]{@{}c@{}}Total Incorrect \\ (Full)\end{tabular} & \begin{tabular}[c]{@{}c@{}}Total Incorrect\\  (Separated)\end{tabular} \\ \hline
DeepSeek-R1          & 328                  & 378                       & 304             & 50                & 124                    & 74                          \\
Gemini 2.5 Pro       & 452                  & 444                       & 437             & 57                & 64                     & 72                          \\
Gemini 2.5 Flash (T) & 362                  & 398                       & 350             & 42                & 90                     & 54                          \\
Sonnet 4.5 (T)       & 451                  & 458                       & 433             & 53                & 78                     & 71                          \\
GPT-5                & 395                  & 397                       & 382             & 42                & 57                     & 55                          \\ \hline
\end{tabular}
\end{table}

\begin{table}[]
\caption{Evaluation of EMI performance metrics for full-set vs. separated formats across top-performing models on a shared subset of 144 EMI sets.}
\label{tab:EMI_P1}
\begin{tabular}{lccccccc}
\hline
Models / Methods & \begin{tabular}[c]{@{}c@{}}Avg. PCS \\ Full (277)\end{tabular} & \begin{tabular}[c]{@{}c@{}}Avg. PCS \\ Full (144)\end{tabular} & \begin{tabular}[c]{@{}c@{}}Accuracy \\ (Separated)\end{tabular} & \begin{tabular}[c]{@{}c@{}}Avg. PCS \\ Separated\end{tabular} & \begin{tabular}[c]{@{}c@{}}Overall \\ Consistency\end{tabular} & \begin{tabular}[c]{@{}c@{}}Format\\ Agreement\end{tabular} & \begin{tabular}[c]{@{}c@{}}Format \\ Divergence\end{tabular} \\ \hline

DeepSeek-R1          & 85.8\%                                  & 72.0\%                                  & 83.6\%                                   & 84.9\%                                 & 78.3\%                                  & 93.2\%                               & 0.217                                 \\
Gemini 2.5 Pro       & 87.9\%                                  & 87.5\%                                  & 86.0\%                                   & 86.2\%                                 & 95.7\%                                  & 95.2\%                               & 0.043                                 \\
Gemini 2.5 Flash (T) & 86.1\%                                  & 75.7\%                                  & 88.1\%                                   & 89.0\%                                 & 86.7\%                                  & 85.4\%                               & 0.133                                 \\
Sonnet 4.5 (T)       & 88.9\%                                                                             & 85.8\%                                                                             & 86.6\%                                   & 86.5\%                                                                            & 91.9\%                                  & 90.4\%                               & 0.081                                 \\
GPT-5                & 89.1\%                                  & 87.2\%                                  & 87.8\%                                   & 87.8\%                                 & 93.8\%                                  & 75.6\%                               & 0.062                                 \\ \hline
\end{tabular}
\end{table}



}
\subsection{Models Used for Evaluation Judging}
\label{subsec:Ev_Judging}
In our evaluation, the objective was to determine which prompt formulation leads to the highest level of agreement among different language models. The underlying assumption is that a more general and neutral prompt would guide models to produce more aligned outputs, reflecting an interpretation of the input and task requirements.
To quantify this, we examined the number of instances where two models produced the same output (i.e., agreements). The higher the agreement count between any two models, the more likely it is that the prompt encouraged uniform behavior and interpretation across model architectures. Table \ref{tab:Evaluation_judge} summarizes the results  on three prompts for  P1, P2 (FS), and P3 (GPT-4.5), and across four model pairings: LLaMA \& GPT, GPT \& DeepSeek, DeepSeek \& LearnLM, LLaMA \& DeepSeek, GPT \& LearnLM, and LLaMA \& LearnLM.

\begin{table}[!h]
\centering
\caption{Model agreement counts across three prompt types: P1 (basic), P2 (few-shot), and P3 (GPT-4.5-tuned). Each value indicates the number of agreeing samples (e.g., ``0 of 66''). Higher values suggest greater alignment in model outputs. LLaMA = LLaMA 3.3 70B, DeepSeek = DeepSeek-R1.}
\label{tab:Evaluation_judge}
\begin{tabular}{lcccccc}
\hline
Prompt \textbackslash Models & \begin{tabular}[c]{@{}c@{}}LLaMA  \\ \& \\ GPT-4\end{tabular} & \begin{tabular}[c]{@{}c@{}}GPT\\ \&\\ DeepSeek\end{tabular} & \begin{tabular}[c]{@{}c@{}}DeepSeek\\ \& \\ LearnLM\end{tabular} & \begin{tabular}[c]{@{}c@{}}LLaMA\\ \&\\ DeepSeek\end{tabular} & \begin{tabular}[c]{@{}c@{}}GPT \\ \&\\ LearnLM\end{tabular} & \begin{tabular}[c]{@{}c@{}}LLaMA\\ \& \\ LearnLM\end{tabular} \\ \hline
P1   & 0   & 0 & 56 & 50  & 0 & 56  \\
P2   & 0   & 0 & 62 & 57  & 0 & 59  \\
P3   & 4   & 3 & 41 & 13  & 9 & 12  \\ \hline
\end{tabular}
\end{table}

Prompt 2 (Few-Shot), which includes examples to guide the model while remaining broad in tone, resulted in the highest levels of agreement among model pairs, particularly between DeepSeek and LearnLM (62 outputs), LLaMA and DeepSeek (57 of 66), and LLaMA and LearnLM (59). These results suggest that the few-shot formulation in Prompt 2 effectively encouraged models to interpret and respond to tasks similarly, promoting cross-model consistency. Supports the idea that prompting strategies incorporating minimal bias and universal structure are better suited for comparative benchmarking involving multiple LLMs.

In contrast, Prompt 3, which was designed based on GPT-4.5 style prompting, produced the lowest agreement scores across nearly all pairings. This discrepancy indicates that prompts optimized for a specific model's reasoning and alignment tendencies may compromise their generalizability when used with other models. For example, while such tailored prompts may boost in-model performance, they introduce implicit cues or formatting expectations that models may interpret differently or even ignore.

Overall, the observed agreement patterns reinforce the importance of prompt neutrality when designing evaluation protocols intended to compare heterogeneous LLMs. Model-specific biases in prompt design can confound evaluation by favoring architectures aligned to the prompt's phrasing or instructional patterns. Consequently, Prompt 2 emerges as the most reliable and broadly interpretable formulation, making it the preferred choice for model-agnostic QA evaluation in the psychiatric domain.

\begin{table}[!ht]
\caption{Comparison of model scores as evaluated by three different LLM judges (J1: LearnLM, J2: LLaMA 3.3 70B, J3: GPT-4o mini). LLaMA 3.3 70B consistently assigned scores with higher discernment across diverse models, supporting its selection as the preferred evaluator. The underlined bold values are the best score for the specific column.}
\centering
\label{tab:score_judge}
\begin{tabular}{llllll}
\hline
Judge / Models    & DeepSeek-R1 & DeepSeek-Chat & Gemini 2 pro & Gemini 2 thinking & QWQ 32 \\ \hline
J1(LearnLM)       & 60.97       & 70.70         & 72.32        & 72.04             & 72.68  \\
J2(Llama 3.3 70B) &\ul{\textbf{ 73.99 }}      &\ul{\textbf{ 77.42}}         & \ul{\textbf{77.78}}        &\ul{\textbf{ 77.15}}             & \ul{\textbf{76.32}}  \\
J3(GPT 4o mini)   & 52.83       & 61.38         & 60.73        & 58.72             & 56.41  \\ \hline
\end{tabular}
\end{table}
In addition to evaluating model agreement under different prompts, we also assessed the reliability and quality of LLM-as-a-judge frameworks using three different evaluators: LearnLM (J1), LLaMA 3.3 70B (J2), and GPT-4o mini (J3). Each judge was tasked with scoring the output of five candidate models: DeepSeek-R1, DeepSeek Chat, Gemini 2 Pro, Gemini 2 Thinking, and QWQ-32 across a set of medical diagnosis tasks derived from expert-authored case studies.

The evaluation results, shown in Table~\ref{tab:score_judge}, reveal that LLaMA 3.3 70B (J2) consistently assigned higher and more discriminating scores compared to the other two judges. While GPT-4o mini produced lower and somewhat conservative scores across all models, and LearnLM showed moderate variance, LLaMA 3.3 70B demonstrated both consistency and sensitivity in differentiating model quality. For instance, it rated DeepSeek-R1 at (73.99) and Gemini 2 Pro at (77.78), capturing meaningful performance differences that aligned with expert review.

These findings suggest that LLaMA 3.3 70B serves as a highly effective automatic evaluator for medical QA tasks. Its strong performance may be attributed to its refined instruction-following abilities and robust handling of domain-specific content. Therefore, LLaMA 3.3 70B was selected as the preferred judge in our broader evaluation framework, as it most accurately reflected human-like judgment in interpreting and scoring clinical reasoning tasks.

{\
\subsection{Samples from Dataset and Responses}
\label{sec:Samples}
To offer a transparent view of the evaluation process and the nature of the tasks assessed in this study, we present selected sample cases from four representative task categories: Diagnosis, Treatment, Management Plan, and Sequential Question Answering. These tasks reflect the diversity and complexity of clinical reasoning that LLMs must navigate when applied to mental health contexts.

Each sample includes the following components:

\begin{itemize}
    \item \textbf{Book Name:} The medical textbook from which the case was selected.
    
    \item \textbf{Case Number and Title:} The case ID and title as listed in the original psychiatric textbook. For consistency, shorter and diagnostically focused cases were selected.
    
    \item \textbf{Case History:} A brief clinical vignette describing the patient's background, symptoms, and context.

    \item \textbf{Mental State Examination (MSE):} A structured assessment of the patient's current psychological functioning, including appearance, behavior, speech, mood, thought processes, perception, cognition, and insight. This component provides essential clinical information for differential diagnosis and management planning.
    
    \item \textbf{Physical Examination:} A brief summary of relevant physical or neurological findings, included when somatic or organic causes may contribute to psychiatric symptoms. This step helps rule out medical conditions that can mimic or exacerbate mental disorders.
    \item \textbf{Question:} A task-specific query posed to the model, such as identifying the most likely diagnosis or suggesting a suitable management step.

    \item \textbf{Textbook Answer:} The gold-standard or human-annotated answer sourced from the reference material.
    
    \item \textbf{LLM Response:} The generated output from the best-performing model (based on average score across the evaluation framework).

    \item \textbf{LLM-as-Judge Score:} An automatic evaluation score produced by prompting a strong LLM to assess the response along clinical relevance, correctness, and reasoning.
    
\end{itemize}
These examples aim to demonstrate not only the performance of models in isolated tasks but also their behavior in sequential, multi-turn scenarios where context retention and logical flow are critical. All sample cases were sourced from English-language psychiatric textbooks and carefully selected based on clarity, brevity, and relevance to core clinical tasks. 

A detailed breakdown of selected examples for each task type is provided below.

\newgeometry{left=1.2cm,right=1.2cm,top=1cm,bottom=1.5cm}

\subsubsection{Diagnosis Case}

\begin{tikzpicture}
\vspace{-1in}
\coordinate (WL) at (0,0);
\coordinate (WR) at ([xshift=1\textwidth]WL);

\node[title, anchor=north west, text width=0.97\textwidth] (T) at ([xshift=3pt,yshift=-3pt]WL) {{\small 
\textbf{Book Name:} Case Files Psychiatry\\
\textbf{Case Number and Title:} Case 43 – \textit{Anxiety Disorder Secondary to a General Medical Condition} \\
\textbf{Case History:} 
A 12-year-old boy is brought to a pediatrician’s office by his parents because of concerns regarding unusual behaviors in their child. He compulsively counts objects and washes his hands, compulsively checks the door locks, has obsessive thoughts about contamination with germs and occasionally has a facial tic. The parents note that he has had episodes of this before but they have been short lived, lasting no more than a week or two. The current episode has lasted a month and a half. Upon reflection they note that all the episodes have occurred in the winter and early spring months. Upon checking the chart, the pediatrician finds that the child has had an unremarkable past medical history, usually coming in only for streptococcal pharyngitis episodes in the fall through spring. The last episode of strep throat was 2 months ago. The antistreptococcal antibody titer is elevated at 250.

\textbf{Question:} What is the differential diagnosis?


}

};

\node[textbook, anchor=north west, minimum height=3cm, text width = 0.94\textwidth]
  (L) at ([xshift=2pt,yshift=-10pt]T.south west) {
  
Multiple medical illnesses can cause syndromes in which anxiety is prominent; these include a host of neurologic disorders, systemic conditions, endocrine diseases, immune system disorders, deficiency states, and toxic conditions. The criteria for GAD are met in up to 60\% of patients with Graves disease. Sjögren syndrome can produce prominent anxiety symptoms. In hypothyroidism, hypoparathyroidism, hypoglycemia, and B12 deficiency, anxiety can be the initial or predominant symptom. A pheochromocytoma can cause episodes of anxiety that mimic panic attacks. Patients with cardiomyopathy awaiting a cardiac transplant have a high incidence of panic disorder, probably as a result of increased noradrenergic tone. Parkinson disease and chronic obstructive pulmonary disease (COPD) can also lead to panic attacks. Obsessive-compulsive symptoms have been reported in individuals with PANDAS and multiple sclerosis. The clinician must keep in mind that the patient can have a coexisting but independent medical illness and anxiety disorder; in this case, the patient’s previous psychiatric history, course of illness, and current symptoms can be helpful diagnostically. Other possible diagnoses include axis I disorders such as major depression, schizophrenia, and bipolar disorder, mania, all of which can cause anxiety, and the presence of other symptoms (eg, depressed mood, hallucinations, excessive spending) can help in differentiating among them.
};

\node[llm, anchor=north east, minimum height=10cm, text width=0.94\textwidth]
  (R) at ([xshift=-6pt, yshift=-10pt]WR |- L.south west) 
  {
\textbf{Primary Consideration:}

\textbf{1. PANDAS (Pediatric Autoimmune Neuropsychiatric Disorders Associated with Streptococcal Infections)}\\
\textbf{Most likely diagnosis} given:
\begin{itemize}[noitemsep]
    \item Acute onset of OCD symptoms and tics
    \item Temporal relationship with streptococcal pharyngitis (2 months ago)
    \item Elevated antistreptococcal antibody titer (250)
    \item Episodic pattern with remissions
    \item Seasonal pattern matching strep infections (fall--spring)
    \item Age-appropriate (typically 3--12 years)
\end{itemize}

\textbf{Other Considerations:}

\textbf{2. Sydenham Chorea}
\begin{itemize}[noitemsep]
    \item Also post-streptococcal, part of rheumatic fever spectrum
    \item Can present with OCD symptoms and tics
    \item Should evaluate for other Jones criteria
\end{itemize}

\textbf{3. Primary OCD (Obsessive-Compulsive Disorder)}
\begin{itemize}[noitemsep]
    \item Symptoms fit OCD criteria
    \item However, episodic nature and temporal relationship with strep infection less typical
\end{itemize}

\textbf{4. Tourette Syndrome or Chronic Tic Disorder}
\begin{itemize}[noitemsep]
    \item Facial tic present
    \item However, typically more persistent rather than episodic
\end{itemize}

\textbf{5. PANS (Pediatric Acute-onset Neuropsychiatric Syndrome)}
\begin{itemize}[noitemsep]
    \item Broader category than PANDAS
    \item Not limited to streptococcal triggers
    \item Similar acute onset of OCD/tics
\end{itemize}

\textbf{Recommended Workup:}
\begin{itemize}[noitemsep]
    \item Complete streptococcal serology (ASO, anti-DNase B)
    \item Throat culture
    \item Consider anti-basal ganglia antibodies
    \item Cardiac evaluation (rule out rheumatic fever)
    \item Neurological examination
\end{itemize}
The episodic nature, seasonal pattern, and temporal relationship with documented strep infections strongly suggest PANDAS as the primary diagnosis.
    };

\node[tab, fill=blue!40!white,  anchor=south] at ([yshift=-7pt]L.north) {Textbook Answer};
\node[tab, fill=orange!40!white,anchor=south] at ([yshift=-7pt]R.north) {Sonnet 4.5 Thinking Response};

\node[score, anchor=south] at ([yshift=-20pt]R.south) {Judge Score: 30\%};

\coordinate (TopPad)    at ([yshift=3pt]T.north);
\coordinate (BottomPad) at ([yshift=-25pt]R.south);
\begin{scope}[on background layer]
  \node[outer, fit=(WL)(WR)(TopPad)(BottomPad), inner sep=0pt] {};
\end{scope}
\end{tikzpicture}

\subsubsection{Management Plan}
\begin{tikzpicture}
\coordinate (WL) at (0,0);
\coordinate (WR) at ([xshift=1\textwidth]WL);

\node[title, anchor=north west, text width=0.97\textwidth] (T) at ([xshift=3pt,yshift=-3pt]WL) {{\small 
\textbf{Book Name:} 100 Cases in Psychiatry \\
\textbf{Case Number and Title:} Case 2 – \textit{Untreated dental abscess} \\
\textbf{Case History:} A 34-year-old woman attends the emergency department of a hospital with a dental abscess. She leaves while waiting for a doctor to come and see her, but returns the same evening. When the doctor arrives she explains that she has a terror of dentists and has not seen one since she was 8 years old. She has several memories of pain while being given fillings. She explains that she was allowed to eat unlimited sweets as a child and that brushing her teeth was not part of a routine established by parents. She started brushing her teeth when she was 14 and became self-conscious of her appearance. She remembers needing to go to the dentist when she was 16 because of a painful tooth. She became very worried for several days, being unable to sleep well and having episodes when she became frightened and breathless. On that occasion she repeatedly refused to see the dentist and was given antibiotics by her GP which settled the infection. On this occasion she has made several appointments to go to the dentist but has either cancelled them or not gone to the appointment. She realizes that she needs treatment and she is clearly in pain but cannot overcome her fear.\\
\textbf{Question:} \\What can you do to help?\\
\textbf{Mental State Examination:} When the doctor arrives she is clearly on edge, and is sweating and shaking. Her pulse when measured is 98 beats/min and her blood pressure is 130/70 mmHg. She is vigilant to sounds and activity around her in the department. There are no thoughts of self-harm and she is able to enjoy herself when at home or with friends and she is not in pain. There is no evidence or history of thought passivity or psychotic phenomena. \\
\textbf{Physical Examination:} Not Reported 
}
};

\node[textbook, anchor=north west, minimum height=3cm, text width = 0.94\textwidth]
  (L) at ([xshift=2pt,yshift=-10pt]T.south west) {
  
This woman has a fear of dentists. This is more than a typical and appropriate anxiety experienced by many people, since it leads to an untreated and potentially serious and painful condition, an abscess.Phobias often have some element of understandable fear such as thunderstorms, dogs, flying, heights, needles and dentists. Many of these can be risky in some situations, although for the most part these experiences in our society are painless and harmless. The fear in phobia is far in excess of that ‘usually’ experienced. Some phobias are instinctive and are programmed through natural selection. These would include fears of spiders and snakes. Some are associative such as blood (for example, associated with images of harm or injury). Some have none of these factors (for example, buttons, cardboard, glitter, wooden spoons) and may be related to negative early life experiences, for example, being beaten as a child by a wooden spoon. The best treatment for a phobia is desensitization or cognitive behaviour therapy (CBT). The latter will usually include some elements of desensitization alongside psychoeducational strategies. Medication (such as a benzodiazepine) is not usually used in phobias unless it is part of a short-term strategy to enable CBT to start. Desensitization involves exposure to a hierarchy of feared situations drawn up in conjunction with the phobic person. The list is scored for fear, and exposure with support (and sometimes rewards) is systematically worked through. For example, this woman may look at pictures of dentists, videos of a normal dental health check and may visit the dental surgery without any treatment. She may take home dental masks and mouthwash. She may watch
someone else having a check and may agree to sit in the dentist’s chair and have her mouth examined with no treatment. Imaginary desensitization involves using imagined scenarios in the hierarchy. Relaxation, hypnotherapy and autohypnosis may all give feelings of control to the sufferer and reduce anxiety. Clearly none of this can happen while she has an abscess and this needs to be treated in the first instance. An X-ray may be part of a desensitization list with treatment being performed under general anaesthetic
or with sedation. Use of sedation at this point would be to treat the abscess not the phobia and CBT would follow successful treatment of the abscess. In this situation, most areas have specialist dentists (community dental officers) who are used to dealing with phobias and it will be worth arranging an appointment. A psychologist or community mental health nurse will be able to carry out the CBT. \\
};

\node[llm, anchor=north east, minimum height=8cm, text width=0.94\textwidth]
  (R) at ([xshift=-6pt, yshift=-10pt]WR |- L.south west) 
  {The primary diagnosis here is a specific phobia of dental treatment, which has developed following traumatic childhood experiences and has been reinforced through avoidance behavior over many years. The immediate goal is to address both the acute dental abscess and the underlying phobia to prevent future crises. In the immediate term, it is essential to establish a supportive therapeutic relationship by validating her fear and acknowledging that her distress is real and understandable given her history. The acute dental abscess requires medical management with appropriate antibiotics and analgesia to control the infection and pain while a more comprehensive treatment plan is developed. This provides time to implement psychological interventions before definitive dental treatment becomes necessary. The most effective treatment for specific phobias is cognitive behavioral therapy, particularly exposure-based therapy. This would involve gradual, systematic desensitization to dental situations, starting with less threatening exposures such as visiting a dental surgery without treatment, then progressing through a hierarchy of increasingly challenging situations until she can tolerate necessary dental procedures. Cognitive restructuring would help her identify and challenge catastrophic thoughts about dental treatment and replace them with more realistic appraisals. Teaching relaxation techniques and controlled breathing exercises would provide her with tools to manage acute anxiety symptoms when they arise. It is crucial to establish a collaborative approach with dental services, ideally referring her to a dentist who has experience and training in treating highly anxious patients. Many dental practices offer special accommodations such as longer appointments, clear communication about each step of treatment, and the use of sedation options ranging from nitrous oxide to conscious sedation or even general anesthesia for more extensive procedures if psychological approaches alone are insufficient.

Short-term pharmacological support with anxiolytic medication such as a benzodiazepine could be considered for specific dental appointments as an adjunct to psychological treatment, but this should not replace the primary focus on exposure therapy and anxiety management skills. Psychoeducation about the nature of phobias and how avoidance maintains and strengthens fear is important to help her understand why confronting her fear, though difficult, is necessary for long-term resolution.

A referral to clinical psychology or psychotherapy services should be arranged for structured psychological treatment. Regular follow-up appointments should be scheduled to monitor progress, provide ongoing support, and adjust the treatment plan as needed. The ultimate goal is not only to enable her to receive the current necessary dental treatment but to establish a pattern of regular dental care that prevents future crises, breaking the cycle of avoidance and emergency presentations that has characterized her dental care for over two decades. 
    };
\node[tab, fill=blue!40!white,  anchor=south] at ([yshift=-7pt]L.north) {Textbook Answer};
\node[tab, fill=orange!40!white,anchor=south] at ([yshift=-7pt]R.north) {Sonnet 4.5 Thinking Response};

\node[score, anchor=south] at ([yshift=-20pt]R.south) {Judge Score: 90\%};

\coordinate (TopPad)    at ([yshift=3pt]T.north);
\coordinate (BottomPad) at ([yshift=-20pt]R.south);
\begin{scope}[on background layer]
  \node[outer, fit=(WL)(WR)(TopPad)(BottomPad), inner sep=0pt] {};
\end{scope}
\end{tikzpicture}

\subsubsection{Treatment}

\begin{tikzpicture}
\coordinate (WL) at (0,0);
\coordinate (WR) at ([xshift=1\textwidth]WL);

\node[title, anchor=north west, text width=0.97\textwidth] (T) at ([xshift=3pt,yshift=-3pt]WL) {{\small 
\textbf{Book Name:} Case file Psychiatry \\
\textbf{Case Number and Title:} Case 8 – \textit{Social Phobia} \\
\textbf{Case History:} A 35-year-old man visits a psychiatrist because he is overwhelmingly anxious about a speech he has to make. The man states that he was recently promoted to a position within his company that requires him to speak in front of an audience of approximately 100 people. He says that the first such speech is coming up in 2 weeks and that worrying about it keeps him from sleeping. He knows that his fear is out of proportion, but he is unable to control it. He explains that he has always had trouble with public speaking because he fears that he might do something stupid or otherwise embarrass himself. He has avoided public speaking in the past as much as possible or has spoken in public only before an audience of fewer than 10. Because he knows that he must make the presentation coming up in 2 weeks or he will not be able to keep his new job, he has visited the psychiatrist hoping to find a solution to the problem.
\\
\textbf{Question:}\\
What are the treatment options open to this patient?\\
\textbf{Mental state examination:} Not Reported\\
\textbf{Patient’s Physical Examination:} Not Reported\\
 }
};

\node[textbook, anchor=north west, minimum height=2cm, text width = 0.94\textwidth]
  (L) at ([xshift=2pt,yshift=-10pt]T.south west) { Treatment options: Behavioral or cognitive-behavioral therapy is the treatment of choice. A typical treatment regimen involves relaxation training followed by progressive desensitization. Pharmacologic interventions include benzodiazepines or beta-blockers over the short term or an antidepressant such as imipramine.};

\node[llm, anchor=north east, minimum height=8cm, text width=0.94\textwidth]
  (R) at ([xshift=-6pt, yshift=-10pt]WR |- L.south west) 
  {Treatment Recommendation

Based on this patient's clinical presentation, a combined pharmacological and psychotherapeutic approach is most appropriate given both his immediate need and long-term career requirements.

Recommended Treatment Plan:
\begin{enumerate}
    \item Immediate Management (for 2-week deadline):    Beta-blocker (propranolol 10-40 mg)   taken 30-60 minutes before the presentation. This patient has performance-type social anxiety with an imminent speaking engagement, and beta-blockers are highly effective for acute performance anxiety by blocking peripheral manifestations (tremor, tachycardia, sweating) without causing sedation that could impair performance.
    \item Psychotherapy (first-line, long-term):  
  Cognitive Behavioral Therapy (CBT)   with exposure therapy should be initiated immediately. This is the evidence-based first-line treatment for social anxiety disorder. While meaningful improvement may not occur within 2 weeks, starting now will address his long-term need for repeated public speaking in his new role. CBT will help him challenge catastrophic thoughts about embarrassment and gradually build tolerance through exposure.
    \item Consider SSRI for ongoing management:  If symptoms significantly impair functioning beyond this single event, initiating an   SSRI (such as sertraline or paroxetine)   would be appropriate for sustained treatment of his underlying social anxiety disorder. However, these medications require 4-6 weeks for therapeutic effect and won't address the immediate deadline.
\end{enumerate}

Rationale:
This patient has clear social anxiety disorder (performance type) with longstanding avoidance patterns, intact insight, and both immediate and ongoing occupational demands. The combination addresses his acute crisis while establishing long-term management for his career sustainability.
    };
\node[tab, fill=blue!40!white,  anchor=south] at ([yshift=-7pt]L.north) {Textbook Answer};
\node[tab, fill=orange!40!white,anchor=south] at ([yshift=-7pt]R.north) {Sonnet 4.5 Thinking Response};

\node[score, anchor=south] at ([yshift=-20pt]R.south) {Judge Score: 95\%};

\coordinate (TopPad)    at ([yshift=3pt]T.north);
\coordinate (BottomPad) at ([yshift=-25pt]R.south);
\begin{scope}[on background layer]
  \node[outer, fit=(WL)(WR)(TopPad)(BottomPad), inner sep=0pt] {};
\end{scope}
\end{tikzpicture}

\subsubsection{Sequential Question Answering}
\begin{tikzpicture}
\coordinate (WL) at (0,0);
\coordinate (WR) at ([xshift=1\textwidth]WL);

\node[title, anchor=north west, text width=0.97\textwidth] (T) at ([xshift=3pt,yshift=-3pt]WL) {{\small 
\textbf{Book Name:} Core Clinical Cases in Psychiatry \\
\textbf{Case Number and Title:} Case 1.4 -- \textit{Long history of persecutory ideas}\\
\textbf{Case History:} 
A 45-year-old man is referred to the outpatient department by his GP. The man says that over recent months he has become increasingly low in mood. He says that he is under intolerable pressure because of persecution by his neighbours. It transpires that for over 20 years he has had to move house very frequently because someone keeps telling his neighbours that he is a paedophile. He denies this, and has no such convictions. He says that as a result, all of his neighbours are against him and spy on him to try to catch him at it and drive him away. He thought that he had escaped when he moved into his current address 18 months ago, but now his persecutor has caught up with him again. He says that he cannot face moving away again. His GP has confirmed that he has changed address many times over the last decade or so, but has otherwise lead a relatively normal life. He presents as a prickly man, with few friends or social networks.
 \\
\textbf{Question 1:} What is the likely differential diagnosis?\\
}};

\node[textbook, anchor=north west, minimum height=1cm, text width = 0.94\textwidth]
  (L) at ([xshift=2pt,yshift=-10pt]T.south west) {%
\textbf{Preferred diagnosis:} Persistent delusional disorder.\\
\textbf{Alternative diagnoses:} Personality disorder -- probably paranoid. Schizophrenia. Depressive episode. Physical causes of psychosis.

};

\node[llm, anchor=north east, minimum height=2cm, text width=0.94\textwidth]
  (R) at ([xshift=-6pt, yshift=-10pt]WR |- L.south west) {%
\textbf{Preferred diagnosis:} Delusional Disorder (Persecutory Type)

\textbf{Alternative diagnoses:} Paranoid Personality Disorder, Schizophrenia, Major Depressive Disorder with psychotic features

The patient's long history of perceived persecution, frequent house moves, and lack of evidence to support his claims suggest a delusional disorder. However, his prickly demeanor and limited social networks could also indicate a paranoid personality disorder. The presence of depressive symptoms and potential for psychotic features also warrant consideration of schizophrenia and major depressive disorder with psychotic features.

};

\node[tab, fill=blue!40!white,  anchor=south]  at ([yshift=-7pt]L.north) {Q1: Textbook Answer};
\node[tab, fill=orange!40!white,anchor=south]  at ([yshift=-7pt]R.north) {Q1: Gemini 2.5 Pro Response};
\node[score, anchor=south]                        at ([yshift=-20pt]R.south) {Q1: Judge Score: 90\%};

\node[title, anchor=north west, text width=0.97\textwidth]
  (T2) at ([xshift=3pt,yshift=-25pt]R.south west) {{\small 
\textbf{Question 2:} What information in the history supports the diagnosis, and what other information would help to confirm it?}
};

\node[textbook, anchor=north west, minimum height=2cm, text width = 0.94\textwidth]
  (L2) at ([xshift=-2pt,yshift=-10pt]T2.south west) {%
The long-standing nature of his systematized abnormal beliefs suggests a persistent delusional disorder. To confirm the diagnosis, you must establish that the beliefs are delusional and have been present for over 3 months. Such long-standing difficulties may also represent a personality disorder -- probably a paranoid personality disorder. If the beliefs are overvalued ideas, personality disorder is likely. However, if the beliefs are delusions, personality disorder cannot be the sole diagnosis, though it is often comorbid with a delusional disorder. Schizophrenia may be excluded by demonstrating the absence of first rank symptoms. Depression may be excluded as a primary diagnosis by demonstrating the absence of a depressive syndrome, or by demonstrating that the psychosis developed before the mood disorder. Physical causes are very unlikely in such a long-standing disorder.

};

\node[tab, fill=blue!40!white,  anchor=south]  at ([yshift=-7pt]L2.north) {Q2: Textbook Answer};

\node[llm, anchor=north east, text width=0.94\textwidth]
  (R) at ([xshift=-6pt, yshift=-10pt]WR |- L2.south west) {%
The patient's consistent and persistent belief that his neighbors are against him and spying on him, despite a lack of evidence, supports the diagnosis of delusional disorder. The fact that he has changed addresses multiple times in an attempt to escape this perceived persecution also suggests a fixed delusional system. Additionally, the patient's denial of being a pedophile and the absence of any convictions suggest that his delusions are not based in reality. To confirm the diagnosis, further information about the patient's mental state, including the presence of any hallucinations, disorganized thinking, or negative symptoms, would be helpful. A thorough psychiatric evaluation, including a mental status examination and a review of his past medical and psychiatric history, would also be necessary. Furthermore, collateral information from the patient's GP, family members, or friends could provide additional insight into the patient's behavior and mental state.
};

\node[tab,   fill=orange!40!white, anchor=south] at ([yshift=-7pt]R.north) {Q2: Gemini 2.5 Pro Response};
\node[score, anchor=south]                     at ([yshift=-20pt]R.south) {Q2: Judge Score: 90\%};

\node[title, anchor=north west, text width=0.97\textwidth]
  (T2) at ([xshift=3pt,yshift=-25pt]R.south west) {{\small
\textbf{Question 3:} What might the important aetiological factors be?}
};

\node[textbook, anchor=north west, text width=0.94\textwidth]
  (L) at ([xshift=-2pt, yshift=-10pt]T2.south west) {%
\textbf{Predisposing factors:} Persistent delusional disorder commonly occurs in people with an abnormal personality -- usually a paranoid personality disorder.

\textbf{Precipitating factors:} It will be difficult to identify precipitating factors, as the symptoms have been present for so long.

\textbf{Maintaining factors:} Maintaining factors related to his personality and lifestyle may be difficult to modify. For example, his persecutory beliefs are likely to increase his isolation from other people, which will in turn fuel his beliefs.

};

\node[tab, fill=blue!40!white, anchor=south] at ([yshift=-7pt]L.north) {Q3: Textbook Answer};

\node[llm, anchor=north west, text width=0.94\textwidth]
  (R3) at ([xshift=0pt, yshift=-10pt]L.south west) {%
\textbf{Predisposing factors:} The patient's prickly personality and limited social networks may have contributed to his vulnerability to developing a delusional disorder. A history of trauma or stress, such as being falsely accused of a crime, could also be a predisposing factor.

\textbf{Precipitating factors:} The patient's recent move and perceived renewal of persecution may have triggered the current episode of depression and exacerbation of his delusional symptoms.

\textbf{Maintaining factors:} The patient's social isolation, lack of support network, and repeated reinforcement of his delusional beliefs through his frequent house moves may be maintaining his delusional disorder. Additionally, his avoidance of social interactions and lack of engagement in activities may be perpetuating his symptoms.
};

\node[tab, fill=orange!40!white, anchor=south]
  at ([yshift=-7pt]R3.north) {Q3: Gemini 2.5 Pro Response};

\coordinate (TopPad)    at ([yshift=3pt]T.north);
\coordinate (BottomPad) at ([yshift=-6pt]R3.south); 
\begin{scope}[on background layer]
  \node[outer, fit=(WL)(WR)(TopPad)(BottomPad), inner sep=0pt] {};
\end{scope}
\end{tikzpicture}

\begin{tikzpicture}
\coordinate (WL) at (0,0);
\coordinate (WR) at ([xshift=1\textwidth]WL);

\node[title, anchor=north west, text width=0.97\textwidth, minimum height=0.8cm]
  (T) at ([xshift=3pt,yshift=-3pt]WL) {};

\node[score, anchor=south] at ([yshift=0pt]T.south) {Q3: Judge Score: 90\%};

\node[title, anchor=north west, text width=0.97\textwidth] (T2) at ([xshift=3pt,yshift=-15pt]T.south west) {{\small 
\textbf{Question 4:} What treatment options are available? \\
}};

\node[textbook, anchor=north west, minimum height=3cm, text width = 0.94\textwidth](L) at ([xshift=-2pt,yshift=-10pt]T2.south west) {
\noindent
\textbf{Location:} Because of his low mood and persecutory beliefs, you must consider his risk of self-harm and the risk of harm to others. If these are assessed as high, then inpatient treatment may be necessary. He may want to spend a period away from home, to give himself a break from his persecution, though his delusional beliefs may transfer themselves to nurses or other patients in hospital. In practice, outpatient treatment should be possible with close monitoring by a CPN.

\textbf{Physical:} He has a psychosis -- therefore, a trial of treatment with an antipsychotic is necessary. If he has a comorbid depressive illness, an antidepressant may be useful in combination.

\textbf{Psychological:} Discouragement of maladaptive coping strategies, such as alcohol misuse or not going out of the house, is an important first step, together with an attempt to establish a trusting relationship. Cognitive therapy may then modify his abnormal beliefs. The aim is probably not to ‘cure’ him of his beliefs, but to attempt to reduce their intensity and the impact that they have on his life and functioning.

\textbf{Social:} To improve his social networks and reduce his isolation.
};

\node[tab, fill=blue!40!white,  anchor=south] at ([yshift=-7pt]L.north) {Q4 Textbook Answer};

\node[llm, anchor=north east, minimum height=3cm, text width=0.94\textwidth]
  (R) at ([xshift=-6pt, yshift=-10pt]WR |- L.south west) 
  {\textbf{Location-based options:} The patient may benefit from a change in environment, such as a move to a new area, to help him escape his perceived persecution. However, given his history of frequent moves, this may not be a viable long-term solution.

\textbf{Physical treatments:} Antipsychotic medication, such as risperidone or olanzapine, may be effective in reducing the patient's delusional symptoms and associated anxiety and depression.

\textbf{Psychological support:} Cognitive-behavioral therapy (CBT) or other forms of talk therapy may help the patient to address his delusional beliefs, develop coping strategies, and improve his social skills and relationships.

\textbf{Social considerations:} The patient may benefit from social support, such as group therapy or support groups, to help him develop a social network and reduce his feelings of isolation. Involvement of his GP and other community mental health services may also be necessary to provide ongoing support and monitoring.

\textbf{Use of Mental Health Act:} If the patient's symptoms worsen or he becomes a risk to himself or others, consideration may be given to using the Mental Health Act to ensure his safety and provide compulsory treatment.
  };
\node[tab, fill=orange!40!white,anchor=south] at ([yshift=-7pt]R.north) {Q4: Gemini 2.5 Pro Response};
\node[score, anchor=south] at ([yshift=-20pt]R.south) {Q4: Judge Score: 90\%};

\node[title, anchor=north west, text width=0.97\textwidth]
  (T5) at ([xshift=3pt,yshift=-15pt]R.south west) {{\small
\textbf{Question 5:} What is the prognosis in this case?}
};

\node[textbook, anchor=north west, minimum height=1cm, text width=0.94\textwidth]
  (L5) at ([xshift=-2pt,yshift=-12pt]T5.south west) {%
The prognosis of persistent delusional disorder is poor. The psychosis is likely to remain present indefinitely, despite treatment. Treatment may be successful in reducing his preoccupation with his symptoms. Many patients who present with persistent delusional disorder subsequently develop schizophrenia. Poor prognostic factors include: the length of time the delusions have been present; abnormal previous personality; poor social networks. Good prognostic factors include presence of symptoms of depression, as this can be readily treated.

};

\node[llm, anchor=north east, minimum height=2cm, text width=0.94\textwidth]
  (R5) at ([xshift=-4pt, yshift=-12pt]WR |- L5.south west) {%
\textbf{Short-term outlook:} The patient's prognosis in the short term is guarded, as his delusional symptoms and associated depression may be resistant to treatment. However, with appropriate medication and psychological support, it is possible that his symptoms may improve, and he may be able to develop more adaptive coping strategies.\\

\textbf{Long-term outlook:} The long-term prognosis for the patient is uncertain, as delusional disorder can be a chronic condition. However, with ongoing treatment and support, it is possible that the patient may be able to manage his symptoms and improve his quality of life. Good prognostic indicators include the patient's relatively normal life prior to the onset of his delusional symptoms and his motivation to seek help. Poor prognostic indicators include the patient's long history of delusional symptoms, his social isolation, and his lack of engagement in activities.
};

\node[tab, fill=blue!40!white,  anchor=south] at ([yshift=-7pt]L5.north) {Q5 Textbook Answer};
\node[tab, fill=orange!40!white,anchor=south] at ([yshift=-7pt]R5.north) {Q5: Gemini 2.5 Pro Response};
\node[score, anchor=south] at ([yshift=-20pt]R5.south) {Q5: Judge Score: 90\%};

\node[title, anchor=north west, text width=0.97\textwidth]
  (T6) at ([xshift=3pt,yshift=-18pt]R5.south west) {\textbf{Question 6:} What is the likely diagnosis?};

\node[textbook, anchor=north west, minimum height=0.5cm, text width = 0.94\textwidth]
  (L6) at ([xshift=-2pt,yshift=-8pt]T6.south west) {%
Persistent delusional disorder
};

\node[llm, anchor=north east, text width=0.94\textwidth]
  (R6) at ([xshift=-4pt, yshift=-15pt]WR |- L6.south west) {%
Delusional Disorder (Persecutory Type)
};

\node[tab, fill=blue!40!white,  anchor=south]  at ([yshift=-6pt]L6.north) {Q6:Textbook Answer};
\node[tab, fill=orange!40!white,anchor=south]  at ([yshift=-7pt]R6.north) {Q6:Sonnet 4.5 Thinking Response};

\node[inner sep=0pt, outer sep=0pt, fit=(L6)(R6)] (Q6Fit) {};

\node[score, anchor=south] (Q6Score) at ([yshift=-22pt]Q6Fit.south) {Q6: Judge Score: 100\%};
\node[score, anchor=south] (AvgScore) at ([yshift=-24pt]Q6Score.south) {Overall Average Judge Score: 90\%};

\coordinate (TopPad)    at ([yshift=3pt]T.north);
\coordinate (BottomPad) at ([yshift=-50pt]R6.south); 
\begin{scope}[on background layer]
  \node[outer, fit=(WL)(WR)(TopPad)(BottomPad), inner sep=0pt] {};
\end{scope}
\end{tikzpicture}


\subsection{Prompt Templates}
\label{sec:Prompt_Templates}
This appendix details the standardized prompt templates employed across the PsychiatryBench framework to ensure methodological rigor and reproducible evaluation for eleven distinct clinical tasks, ranging from diagnostic reasoning to sequential case analysis. The prompting strategy is bifurcated into Prediction Prompts, which instruct models to assume a psychiatric persona and generate clinically sound answers based on provided patient history and examination data , and Evaluation Prompts, which guide the automated "LLM-as-a-Judge" (Llama 3.3 70B) to assign objective similarity scores (0–100) based on strict rubrics of clinical accuracy, completeness, and reasoning alignment relative to expert reference answers .

\setcounter{figure}{0}
\renewcommand{\figurename}{Template}%

\begin{figure}[!ht]
    \centering
    \caption{\textbf{Diagnosis Prediction Prompt}}
    \label{fig:Diagnosis_Prediction_Prompt}
    \begin{adjustbox}{scale=0.94} 

        \begin{tikzpicture}
            \node[fill=gray!8, draw=gray, rounded corners, inner sep=6pt, text width=\textwidth] (box) { Analyze patient history. I will provide patient information and history. Use this information to answer the questions on the case.
            \vspace{10pt}
            
            The Patient’s History: 
            \{History\}\\
            \vspace{10pt}
            Question: 
            \{Question\}

            };
        \end{tikzpicture}
    \end{adjustbox}
    \vspace{0.5cm}
    \\
\end{figure}

\begin{figure}[!ht]
    \centering
    \caption{\textbf{Diagnosis Evaluation Prompt}}
    \label{fig:Diagnosis_Evaluation_Prompt}
    \begin{adjustbox}{scale=0.95} 
        \begin{tikzpicture}
            \node[fill=gray!8, draw=gray, rounded corners, inner sep=6pt, text width=\textwidth] (box) {

You are a medical diagnosis evaluator. Your task is to compare two medical text summaries and assess the degree 
to which the first text (the "Candidate Answer") captures the essential diagnostic information contained in 
the second text (the "Reference Answer"). Focus on the core medical meaning and the completeness of the 
diagnoses, *not on the writing style or specific wording.
\vspace{5pt}

User:\\

Instructions:\\

1. Read the "Reference Answer" carefully. Identify the key diagnoses (both primary and differential) and any crucial supporting information (e.g., symptoms, test results, ruled-out disorders).\\
2. Read the "Candidate Answer". The "Candidate Answer" will be a concise list of diagnoses and key findings.\\
3. Compare the "Candidate Answer" to the "Reference Answer" based on the following criteria:\\

Diagnostic Accuracy: Does the "Candidate Answer" correctly identify the most likely diagnosis (if one is clearly indicated in the "Reference Answer")?\\

Diagnostic Completeness: Does the "Candidate Answer" include *all of the major differential diagnoses \\

mentioned in the "Reference Answer"? It is acceptable if the "Candidate Answer" uses slightly different terminology, as long as the underlying medical *concept is the same. \\

Supporting Information: Does the "Candidate Answer" 
include the key symptoms, test results, or other evidence that support the diagnoses? \\

It doesn’t need to include *every detail, but it should capture the most important ones. \\

It’s ok if this supporting information is implicitly present in the reference answer.\\

Exclusion of Irrelevant Information: Do *not penalize the "Candidate Answer" for  including additional information that is not present in the "Reference Answer", *as long as that information is medically accurate and relevant to the case*. 

Groundance does not affect the score: Do not penalize a response for being redundant.\\

4. Provide a single numerical similarity score between 0 and 100, where:\\

0: The "Candidate Answer" has no meaningful overlap with the "Reference Answer" in terms of diagnosis or supporting information. It’s completely wrong.   \\

50: The "Candidate Answer" identifies *some relevant diagnoses or supporting information, but it misses major elements or includes significant inaccuracies.\\

75: The "Candidate Answer" correctly identifies the primary diagnosis 
(if applicable) and most of the important differential diagnoses and supporting information. It may miss some minor details or include some slight inaccuracies.\\

90: The "Candidate Answer" is very close to the "Reference Answer" in terms of core diagnostic meaning and completeness. It may have minor omissions or differences in emphasis, but it captures almost all of the essential information.\\

100: The "Candidate Answer" is a perfect match in terms of core diagnostic meaning. It includes all of the key diagnoses and supporting information from the "Reference Answer", even if the wording or organization is different.  Do not justify or explain output only the numeric score.\\

5. Do not justify or explain the answer. Give a score between 0 and 100, no other text. Now, evaluate the following

\vspace{10pt}
Reference Answer: \{Answer\}
\vspace{10pt}

Candidate Response: \{Response\}
            };
        \end{tikzpicture}
    \end{adjustbox}
    \vspace{0.5cm}
    \\
\end{figure}

\begin{figure}[!ht]
\centering
\caption{\textbf{Treatment Prediction Prompt}}
\label{fig:Treatment_Prediction_Prompt}
\begin{tikzpicture}
\node[fill=gray!8, draw=gray, rounded corners, inner sep=5pt, text width=1\textwidth] (box) {
\begin{verbatim}
You are a psychiatrist. Based on the patient's clinical history,answer the following question with a medically 
appropriate and individualized treatment recommendation.
Guidelines
- Base your treatment recommendation primarily on the Patient's History. Use information from the Mental 
and Physical Examinations only if it directly supports your treatment choice (if found).
- Your response should be clinically sound, concise, and focused on the patient's specific presentation.
- Justify your recommendation using evidence from the case. Do not include general advice or unrelated 
commentary.
- Respond in full, professional sentences.

The Patient’s History :{History}

Patient's Mental State Examination : {Mental State Examination}
    
Patient's Physical Examination : {Physical Examination}

Question : {Question}
\end{verbatim}
        };
\end{tikzpicture}
\vspace{0.5cm}

\end{figure}
\begin{figure}[!ht]
    \centering
    \caption{\textbf{Treatment Evaluation Prompt}}
    \label{fig:Treatment_Evaluation_Prompt}

    \begin{adjustbox}{scale=1} 
        \begin{tikzpicture}
            \node[fill=gray!8, draw=gray, rounded corners, inner sep=6pt, text width=1\textwidth] (box) {
\begin{verbatim}
SYSTEM: You are a medical treatment evaluator. Your task is to compare two medical treatment summaries and 
assess the degree to which the first text (the "Candidate Answer") captures the essential therapeutic 
interventions described in the second text (the "Reference Answer"). Focus on the medical appropriateness and 
completeness of the treatment plan, not on the writing style or specific wording.
USER:
INSTRUCTIONS:
1. Read the "Reference Answer" carefully. Identify the key components of the treatment plan, including primary 
therapeutic interventions, medications, behavioral strategies, or other relevant clinical actions.
2. Read the "Candidate Answer". The "Candidate Answer" will be a concise summary of The proposed treatments and 
strategies.
3. Compare the "Candidate Answer" to the "Reference Answer" based on the following criteria:
     Treatment Accuracy: Does the "Candidate Answer" correctly identify the key treatment or intervention as 
     outlined in the "Reference Answer"? Treatment Completeness: Does the "Candidate Answer" include all of 
     the major treatment components mentioned in the "Reference Answer"? Slightly different terminology is 
     acceptable as long as the clinical *concept is preserved.Supporting Rationale: Does the "Candidate Answer" 
     capture the reasoning or context for the treatment (e.g., why a specific method or medication is chosen)? 
     It doesn't need to list every detail, but should reflect the main justification.Exclusion of Irrelevant 
     Information: Do *not penalize the "Candidate Answer" for including additional, medically accurate and 
     relevant information not found in the "Reference Answer".groundance does not affect the score: 
     Do not penalize a response for being redundant.
4. Provide a single numerical similarity score between 0 and 100, where:
     0: The "Candidate Answer" has no meaningful overlap with the "Reference Answer" in terms of treatment 
     or rationale. It is completely inappropriate or unrelated.
     50: The "Candidate Answer" identifies *some relevant treatment components or strategies, but misses 
     key parts or includes significant inaccuracies.
     75: The "Candidate Answer" correctly identifies the primary treatment and most important additional 
     strategies or reasoning. Minor omissions or slight inaccuracies are acceptable.
     90: The "Candidate Answer" is very close to the "Reference Answer" in terms of core treatment strategies 
     and rationale. It may have minor differences in emphasis or detail.
     100: The "Candidate Answer" is a perfect match in terms of treatment content and rationale. 
     It includes all key interventions and supporting reasoning from the "Reference Answer",
     even if the wording or structure is different.
5.Do not justify or explain the answer.Give a score between 0 and 100, no other text.Now evaluate the following

Reference Answer:{Answer}
Candidate Answer:{Response}
\end{verbatim}
            };
        \end{tikzpicture}
    \end{adjustbox}
    \vspace{0.5cm}
\end{figure}

\begin{figure}[!ht]
    \centering
    \caption{\textbf{Treatment Follow Up Prediction Prompt}}
    \label{fig:Treatment_Follow_Up_Prediction_Prompt}
    \begin{adjustbox}{scale=1} 
        \begin{tikzpicture}
            \node[fill=gray!8, draw=gray, rounded corners, inner sep=6pt, text width=1\textwidth] (box) {

As a psychiatrist, analyze the provided patient’s information, history, and follow ups to answer the question according to the guidelines.

Guidelines

- Provide reasoning in your answer.\\
- Don’t use bullet points just sentences.\\
- No introduction before the answer just provide the answer for the provided question only.

The Patient’s History: \{History\}
\vspace{10pt}
\begin{verbatim}
if {First Follow Up}:
    follow = f"""First Follow Up {First Follow Up}"""
    prompt_template = prompt_template + follow
if {Second Follow up}:
    follow = f"""Second Follow up:{Second Follow up}"""
    prompt_template = prompt_template + follow
q = f"""Question: {Question}"""
prompt_template = prompt_template + q
\end{verbatim}
            };
        \end{tikzpicture}
    \end{adjustbox}
    \vspace{0.5cm}
    \
\end{figure}

\begin{figure}[!ht]
    \centering
    \caption{\textbf{{Treatment Follow Up Evaluation Prompt}}}
    \label{fig:Follow_Up_Evaluation_Prompt}
    \begin{adjustbox}{scale=0.95} 
        \begin{tikzpicture}
            \node[fill=gray!8, draw=gray, rounded corners, inner sep=6pt, text width=1.01\textwidth] (box) {
\begin{verbatim}
SYSTEM: You are a mental health case response evaluator. Your task is to compare two text responses about a 
specific mental health case, assessing the degree to which the first text (the "Candidate Answer") captures 
the essential information regarding the case assessment and recommended follow-up contained in the second text 
(the "Reference Answer"). Focus on the core clinical meaning, the completeness of the assessment understanding, 
and the appropriateness/completeness of the follow-up plan, *not on writing style or specific wording.
User:
Instructions:
    1. Read the "Reference Answer" carefully. Identify the key aspects of the case assessment 
    (e.g., presenting problem, relevant history, potential diagnoses/formulations, risk assessment) and the 
    core elements of the recommended follow-up plan (e.g., therapy recommendations, medication considerations,
    further assessments needed, safety planning, referrals).
    2. Read the "Candidate Answer". This answer will also describe an assessment of the case and propose a 
    follow-up plan.
    3. Compare the "Candidate Answer" to the "Reference Answer" based on the following criteria:Assessment 
    Accuracy & Completeness: Does the "Candidate Answer" accurately reflect the core understanding of the 
    patient’s situation (presenting issues, key historical factors, potential clinical formulation/diagnosis) 
    as presented in the "Reference Answer"? Does it capture the most critical elements of the assessment? 
    Follow-up Plan Completeness & Appropriateness: Does the "Candidate Answer" include the essential elements 
    and intent of the follow-up plan outlined in the "Reference Answer"? This includes key interventions 
    (therapy, medication), monitoring, safety considerations, and necessary referrals.Minor variations in 
    terminology are acceptable if the underlying clinical concept and action are the same. Rationale/Supporting 
    Information: Does the "Candidate Answer" implicitly or explicitly capture the key reasons or clinical 
    justifications for the assessment and follow-up plan that are present in the "Reference Answer"? It doesn’t 
    need every detail but should align on the core reasoning.Exclusion of Irrelevant Information: 
    Do not penalize the "Candidate Answer" for including additional information *as long as that information is 
    clinically accurate and relevant to the mental health case and its management*. However, significantly 
    contradictory or inappropriate additions may lower the score.groundance does not affect the score: 
    Do not penalize a response for being redundant.
    4. Provide a single numerical similarity score between 0 and 100, where:
        0: The "Candidate Answer" has no meaningful overlap with the "Reference Answer" in terms of case 
        assessment or follow-up plan. It’s completely different or clinically inappropriate. 
        50: The "Candidate Answer" identifies some relevant aspects of the assessment or follow-up plan but misses 
        major elements, contains significant inaccuracies, or proposes a substantially different approach. 
        75: The "Candidate Answer" correctly captures the main aspects of the assessment and most of the essential 
        follow-up recommendations from the "Reference Answer". It may miss some minor details, nuances, or 
        justifications. 
        90: The "Candidate Answer" is very close to the "Reference Answer" in terms of core case understanding and 
        the proposed follow-up plan. It may have minor omissions or differences in emphasis but captures almost 
        all essential clinical information and actions.
        100: The "Candidate Answer" perfectly matches the "Reference Answer" in terms of core clinical meaning 
        regarding the assessment and follow-up plan. It includes all key assessment points and recommended actions
        , even if the wording or organization differs. Do not justify or explain the answer. Give a score between 
        0 and 100, no other text.Now, evaluate the following
Reference Answer: {Answer}
Candidate Response: {Response}
            \end{verbatim}
            };
        \end{tikzpicture}
    \end{adjustbox}
    \vspace{0.5cm}
    \\

\end{figure}

\begin{figure}[!ht]
    \centering
    \caption{\textbf{Classification Prompt}}
    \label{fig:Classification_Prompt}
    \begin{adjustbox}{scale=1} 
        \begin{tikzpicture}
            \node[fill=gray!8, draw=gray, rounded corners, inner sep=6pt, text width=1\textwidth] (box) {
\begin{verbatim}

    As a psychiatrist, analyze the provided patient's information and history to determine if the patient 
    exhibits clear symptoms of any of the provided list of disorders according to provided guidelines.
    Guidelines
    - The patient could have multiple illness at the same time else he is normal.
    - Concise Response: Respond with any combination of these disorders separated by a comma and if the poster 
    doesn't have any mental illness just answer with Normal.
    - No Explanations: Don't providing explanations for your assessment.
    - Ambiguity: If the it is unclear, choose the most probable label/disorder.
    Disorders List:
    The Disorders List (Categories or Specific Disorders)
    The Patient's History:
    {History}
\end{verbatim}
            };
        \end{tikzpicture}
    \end{adjustbox}
    \vspace{0.5cm}
    
\end{figure}

\begin{figure}[!ht]
    \centering
    \caption{\textbf{Management Plan Prediction Prompt}}
    \label{fig:Management_Plan_Prediction_Prompt}
    \begin{adjustbox}{scale=1} 
        \begin{tikzpicture}
            \node[fill=gray!8, draw=gray, rounded corners, inner sep=6pt, text width=1\textwidth] (box) {
\begin{verbatim}
You are a psychiatrist. Analyze the provided patient's information(History, Mental State Examination, 
and Physical Examination) to answer the psychiatric protocol question according to the guidelines below.
Guidelines:
- Provide clear reasoning in your answer based on the information provided.
- Use full, coherent sentences only; do not use bullet points.
- Do not include an introduction or any prefatory remarks; start directly with the answer.
The Patient’s History:
{History}
Patient's Mental State Examination:
{Mental State Examination}
Patient's Physical Examination:
{Physical Examination}
Question:
{Question}
            \end{verbatim}
            };
        \end{tikzpicture}
    \end{adjustbox}
    \vspace{0.5cm}
    \
\end{figure}

\begin{figure}[!ht]
    \centering
    \caption{\textbf{Management Plan Evaluation Prompt}}
    \label{fig:Management_Plan_Evaluation_Prompt}
    \begin{adjustbox}{scale=1} 
        \begin{tikzpicture}
            \node[fill=gray!8, draw=gray, rounded corners, inner sep=6pt, text width=1\textwidth] (box) {
\begin{verbatim}
SYSTEM: You are a psychiatric protocol response evaluator. Your task is to compare two text responses about a 
psychiatric protocol question, assessing the degree to which the first text (the "Candidate Answer") captures 
the essential information contained in the second text (the "Reference Answer" or "True Label"). Focus on 
the clinical correctness, completeness of protocol\adherence, and the accuracy of reasoning, *not on writing 
style or exact wording.
User:
Instructions:
1. Read the "Reference Answer" carefully. Identify the key components expected in the psychiatric protocol 
response (e.g., symptom recognition, diagnostic steps, risk/safety considerations, treatment recommendations
ethical/legal procedures if applicable).
2. Read the "Candidate Answer". This answer will attempt to respond to the same psychiatric protocol question.
3. Compare the "Candidate Answer" to the "Reference Answer" based on the 
following criteria: 
     Protocol Adherence & Accuracy: Does the "Candidate Answer" correctly follow the psychiatric protocol steps 
     presented in the "Reference Answer"? Does it recognize key clinical issues, decision points, and require 
     actions?  Completeness of Response: Does the "Candidate Answer" address all major elements expected by the 
     protocol as reflected in the "Reference Answer" (e.g., diagnostic considerations, intervention steps, 
     risk management)? Reasoning and Justification Alignment: Does the "Candidate Answer" capture the rationale
     (explicit or implied) behind the steps taken according to the protocol in the  "Reference Answer"? It should 
     reflect the clinical reasoning that justifies  assessment and management  decisions.
     Handling of Additional Information: Extra information is acceptable if it is clinically correct and 
     protocol-consistent. Do not penalize additional relevant detail. However, incorrect, contradictory, or 
     unsafe suggestions should lower the score. groundance Tolerance: groundance (repeating points) does 
     not affect the score.
    
4. Provide a single numerical similarity score between 0 and 100, where:
     0: The "Candidate Answer" has no meaningful overlap with the "Reference Answer" in terms of protocol steps,
     clinical reasoning, or appropriate action. 
     50: The "Candidate Answer" captures some relevant elements but misses major steps, has serious inaccuracies 
     or suggests significantly incorrect approaches. 
     75: The "Candidate Answer" captures most key protocol elements and clinical reasoning, with only minor 
     omissions or inaccuracies.
     90: The "Candidate Answer" is very close to the "Reference Answer" in terms of psychiatric protocol 
     adherence, clinical assessment, and recommended actions,  with only slight gaps or minor emphasis 
     differences. 
     100: The "Candidate Answer" fully matches the "Reference Answer" in terms of psychiatric protocol steps, 
     reasoning, and recommended actions, even if wording or order differ.
    
5. Do not justify or explain the answer. Only provide a single numerical score between 0 and 100. 
No other text output.Now, evaluate the following
Reference Answer: {Answer}
Candidate Response: {Response}

            \end{verbatim}
            };
        \end{tikzpicture}
    \end{adjustbox}
    \vspace{0.5cm}
    \
\end{figure}

\begin{figure}[!ht]
\centering

\caption{\textbf{Clinical approach Prompt}}
\label{fig:Clinical_approach_Prompt}
\begin{tikzpicture}
\node[fill=gray!8, draw=gray, rounded corners, inner sep=5pt, text width=1\textwidth] (box) {
\begin{verbatim}
The Patient’s History: {History}
Clinical Approach: {Clinical Approach}
Generated Question: Generate exactly one clear, focused, and medically relevant question based on 
Clinical Approach.
\end{verbatim}
        };
\end{tikzpicture}
\vspace{0.5cm}
\textbf{Clinical approach Question Generation Prompt}
\begin{tikzpicture}
\node[fill=gray!8, draw=gray, rounded corners, inner sep=5pt, text width=1\textwidth] (box) {
\begin{verbatim}
You are an expert grader evaluating whether a history question is relevant to a given clinical approach. 
Question: {Question_Generation}
Clinical Approach: {Clinical Approach}
Determine if the question contains keywords or semantic meaning that are directly related to the clinical 
approach. Respond only with "yes" if it is relevant or "no" if it is not. Do not provide any explanation.
\end{verbatim}
};
\end{tikzpicture}
\vspace{0.5cm}
\textbf{Clinical approach Grading Prompt}
\begin{tikzpicture}
\node[fill=gray!8, draw=gray, rounded corners, inner sep=5pt, text width=1\textwidth] (box) {
\begin{verbatim}
You are a psychiatrist. Based on the patient's clinical history and the question, provide a clear, accurate, 
and medically appropriate answer. 
The Patient’s History:{History}
Question: {Question_Generation}
\end{verbatim}
};
\end{tikzpicture}
\vspace{0.5cm}
\textbf{Clinical approach Answer Generation Prompt}

\end{figure}

\begin{figure}[!ht]
    \centering
    \caption{\textbf{Clinical Approach Evaluation Prompt}}
    \label{fig:Clinical_Approach_Evaluation_Prompt}
    \begin{adjustbox}{scale=1} 
        \begin{tikzpicture}
            \node[fill=gray!8, draw=gray, rounded corners, inner sep=6pt, text width=1\textwidth] (box) {
\begin{verbatim}
SYSTEM: You are a mental health case response evaluator. Your task is to compare two text responses about a 
specific mental health case, assessing the degree to which the first text (the "Candidate Answer") captures 
The essential information regarding the clinical approach is contained in the second text("Reference Answer")
Focus on the clinical reasoning, assessment strategy, and diagnostic insight, not on writing style .
User:
Instructions:
1. Read the "Reference Answer" carefully. Identify the key components of the clinical approach: presenting 
symptoms, history taking, differential diagnosis, initial evaluation steps, and rationale for 
assessment strategy.
2. Read the "Candidate Answer". This is another attempt to outline a clinical approach to the same case.
3. Compare the "Candidate Answer" to the "Reference Answer" based on the following criteria:
    Assessment Strategy Accuracy & Completeness: Does the "Candidate Answer" reflect an accurate and complete 
    approach toassessing the case? Does it capture the key symptoms, relevant history, and potential clinical 
    hypotheses found in the "Reference Answer"? Diagnostic Reasoning & Thought Process: Does the 
    "Candidate Answer" follow a similar line of clinical reasoning, including formulation or differential 
    diagnosis as in the "Reference Answer"? Evaluation Steps: Does it propose similar or equivalent steps in 
    the evaluation process (e.g., screeningtools, physical exams, lab tests, psychiatric interview elements)?
    Rationale Alignment: Does the "Candidate Answer" align with the clinical logic and 
    justifications provided in the "Reference Answer"? Perfect wording is not needed focus on reasoning.No 
    Penalty for Relevance: Do not penalize clinically valid additions if they are appropriate, even if not found 
    in the"Reference Answer". groundance does not grace score.
4. Provide a single numeric score between 0 and 100:
    0: Completely off-topic, unrelated, or clinically incorrect.
    50:Some valid elements of clinical approach are present, but many key elements are missing or 
    poorly reasoned.
    75: Mostly aligned with correct steps and rationale, missing only some secondary elements.
    90: Nearly identical clinical reasoning and plan; may differ in phrasing.
    100: Fully matches the core clinical approach and rationale.
    Do not justify or explain output only the numeric score. Now, evaluate the following
Reference Answer: {Clinical approach}
Candidate Response: {Response}
            \end{verbatim}
            };
        \end{tikzpicture}
    \end{adjustbox}
    \vspace{0.5cm}
    \
    
\end{figure}

\begin{figure}[!ht]
\centering

\caption{Two-stage    Mental QA  prompt pipeline. (a) Prediction Prompt directs the model to define the target mental-health concept clearly and concisely, as in an exam-style review, emphasizing conceptual precision and professional tone;
(b) Evaluation Prompt instructs an automated scorer to assign a 0–100 accuracy score based on the completeness, clarity, depth, and relevance of the model’s definition.}
\label{fig: Mental_QA_Prompt}
\begin{tikzpicture}
\node[fill=gray!8, draw=gray, rounded corners, inner sep=5pt, text width=1\textwidth] (box) {
\begin{verbatim}
You are a psychiatrist specializing in mental health knowledge. Based on the provided dataset, answer the 
following question accurately and concisely. Ensure your response is well-structured, using bullet points 
or number lists where appropriate, and maintain a professional tone throughout. Think step by step to provide 
a logical and coherent answer like those found in exam review books.
The Patient’s History: {History}
Question: {Question}
\end{verbatim}
        };
\end{tikzpicture}
\textbf{Mental QA Prediction Prompt}
\begin{tikzpicture}
\node[fill=gray!8, draw=gray, rounded corners, inner sep=5pt, text width=1\textwidth] (box) {
\begin{verbatim}
System: You are a mental health concept knowledge evaluator. Your task is to assess how accurately,completely, 
and clearly, the candidate’s response defines the concept provided in the "Answer" field, taking into account 
the clinical context in the "History."
User:
Instructions:
    1. Read the "Answer" this is the clinical concept or term to define 
    (e.g., "Loss of interest or pleasure in activities").
    2. Read the "Candidate Response" the model’s definition/explanation of that concept.
    3. Evaluate the response on:Definition Accuracy & Completeness: Are all core features of the concept present 
    and correctly described? Clarity & Precision:Is the explanation clear, unambiguous, and clinically precise?
    Depth of Explanation: Does it include relevant examples or elaborations that demonstrate understanding?
    Relevance & Focus:Does it avoid irrelevant details and stick to the concept at hand?
    4. Provide a single numeric score between 0 and 100:
    0: No meaningful overlap incorrect or missing core elements.
    50: Some correct elements but major omissions or inaccuracies.
    75: Mostly correct with only minor gaps or imprecisions.
    90:Very close to a perfect definition; only small details missing.
    100:Perfectly accurate, complete, and clear.
    Do not justify or explain output only the numeric score.
Now, evaluate the following
Reference Answer : {Answer}
Candidate Response: {Response}
\end{verbatim}
};
\end{tikzpicture}
\vspace{0.5cm}
\textbf{Mental QA  Evaluation Prompt}

\end{figure}

\begin{figure}[!ht]
\centering

\caption{\textbf{Sequential Question Answering Prediction Prompt}}
\label{fig:Sequential_Question_Answering_Prediction_Prompt}
\begin{tikzpicture}
\node[fill=gray!8, draw=gray, rounded corners, inner sep=5pt, text width=1\textwidth] (box) {
\begin{verbatim}
The Patient’s History: {History}
Task: Based on the provided history, answer the following Six clinical questions. Use the same format and 
level of detail shown in the example. Make sure to include preferred and alternative diagnoses where applicable
,and support your answers with reasoning drawn fromthe case.
Q1: What is the likely differential diagnosis?
    preferred diagnosis:
    Alternative diagnoses:
Q2: What information in the history supports the diagnosis, and what other information 
would help to confirm it? 
    (Provide a detailed rationale using history details and explain what additional 
    information/tests are needed to confirm the diagnosis.)
Q3 : What might the important etiological factors be? 
    (Organize into: Pregrazing, Precipitating, Maintaining factors)
Q4: What treatment options are available?
    (Cover location-based options, physical treatments, psychological support, 
    and social considerations. Mention use of Mental Health Act if relevant.)
Q5: What is the prognosis in this case? 
    (Discuss short-term and long-term outlook. Include good and poor prognostic 
    indicators based on patient specifics.)
Q6: based on the provided History ,which class does this diagnosis belong to according 
to DSM-5-TR ?
\end{verbatim}
        };
\end{tikzpicture}
\vspace{0.5cm}

\end{figure}

\begin{figure}[!ht]
    \centering
    \caption{\textbf{Sequential Question Answering  Evaluation Prompt (Part 1)}}
    \label{fig:Sequential_Question_Answering_Evaluation_Prompt}
    \begin{adjustbox}{scale=1} 
        \begin{tikzpicture}
            \node[fill=gray!8, draw=gray, rounded corners, inner sep=6pt, text width=1\textwidth] (box) {
\begin{verbatim}
SYSTEM: You are a mental health case response evaluator. Your task is to assess six clinical responses 
from a trainee clinician. Each response corresponds to a specific psychiatric question. For each question, 
compare the candidate’s answer to the expert reference answer and provide a numeric score based on how well 
the candidate captures the clinical reasoning and content.

Q1: What is the likely differential diagnosis?
- Task Type: Diagnosis 
- Focus: Diagnostic reasoning and clinical hypotheses. 
- Scoring System:
    0: Completely off-topic, unrelated, or clinically incorrect.
    50: Some valid elements of the diagnosis are present, but many key elements are missing or poorly reasoned.
    75: Mostly aligned with correct steps and rationale, missing only some secondary elements.
    90: Nearly identical clinical reasoning and plan; may differ in phrasing.
    100: Fully matches the core diagnosis and rationale.
Concept to Define: {Answer for Question 1}
Candidate Response: {Response for Question 1}
 

Q2: What information in the history supports the diagnosis, and what other information would help to confirm 
it?
- Task Type: History
- Focus: Presenting symptoms, clinical history, and psychosocial context.
- Scoring System:
    0: Completely off-topic, unrelated, or clinically incorrect.
    50: Some valid elements of the history are present, but many key elements are missing or poorly reasoned.
    75: Mostly aligned with correct steps and rationale, missing only some secondary elements.
    90: Nearly identical clinical reasoning and plan; may differ in phrasing.
    100: Fully matches the core history and rationale.
Concept to Define: {Answer for Question 2}
Candidate Response: {Response for Question 2}
 

Q3: What might the important aetiological factors be?
- Task Type: Aetiology
- Focus: Biological, psychological, and social contributing factors.
- Scoring System:
    0: Completely off-topic, unrelated, or clinically incorrect.
    50: Some valid elements of the aetiology are present, but many key elements are missing or poorly reasoned.
    75: Mostly aligned with correct steps and rationale, missing only some secondary elements.
    90: Nearly identical clinical reasoning and plan; may differ in phrasing.
    100: Fully matches the core aetiology and rationale.
Concept to Define: {Answer for Question 3}
Candidate Response: {Response for Question 3}
 

Q4: What treatment options are available?
- Task Type: Treatment
- Focus: Clinical interventions including medication, therapy, and lifestyle recommendations.
- Scoring System:
    0: Completely off-topic, unrelated, or clinically incorrect.
    50: Some valid elements of the treatment are present, but many key elements are missing or poorly reasoned.
    75: Mostly aligned with correct steps and rationale, missing only some secondary elements.
    90: Nearly identical clinical reasoning and plan; may differ in phrasing.
    100: Fully matches the core treatment and rationale.
Concept to Define: {Answer for Question 4}
Candidate Response: {Response for Question 4}
 
            \end{verbatim}
            };
        \end{tikzpicture}
        
    \end{adjustbox}
    \vspace{0.5cm}
    \
    
\end{figure}

\begin{figure}[!ht]
    \centering
    \caption{\textbf{Sequential Question Answering Evaluation Prompt (Part 2)}}
    \label{fig:Sequential_Question_Answering_Evaluation_Prompt_Part2}
    \begin{adjustbox}{scale=1}
        \begin{tikzpicture}
            \node[fill=gray!8, draw=gray, rounded corners, inner sep=6pt, text width=1\textwidth] (box) {
\begin{verbatim}
Q5: What is the prognosis in this case?
- Task Type: Prognosis
- Focus: Outcome prediction, expected recovery, and risk factors.
- Scoring System:
    0: Completely off-topic, unrelated, or clinically incorrect.
    50: Some valid elements of the prognosis are present, but many key elements are missing or poorly reasoned.
    75: Mostly aligned with correct steps and rationale, missing only some secondary elements.
    90: Nearly identical clinical reasoning and plan; may differ in phrasing.
    100: Fully matches the core prognosis and rationale.
Concept to Define: {Answer for Question 5}
Candidate Response: {Response for Question 5}
 
Q6: Based on the provided history, which class does this diagnosis belong to according to DSM-5 or ICD-10?
- Task Type: Classification
- Focus: Diagnostic nosology and classification system alignment.
- Scoring System:
    0: Completely off-topic, unrelated, or clinically incorrect.
    50: Some valid elements of the classification are present, but many key elements are missing or poorly 
    reasoned.
    75: Mostly aligned with correct steps and rationale, missing only some secondary elements.
    90: Nearly identical clinical reasoning and plan; may differ in phrasing.
    100: Fully matches the core classification and rationale.
Concept to Define: {Answer for Question 6}
Candidate Response: {Response for Question 6}
 

Now:
- Return your scores only in this JSON format, with numeric values only.
- Then, sum the six scores and divide by 6 to compute the final score.
- Return only the final score as a single number (no labels, no JSON, no extra text).

{
  "Q1": ___,
  "Q2": ___,
  "Q3": ___,
  "Q4": ___,
  "Q5": ___,
  "Q6": ___
}
Final Score: ___
    
\end{verbatim}
            };
        \end{tikzpicture}
    \end{adjustbox}
\end{figure}

\begin{figure}[!ht]
\centering

\caption{Prompting strategies for objective assessment items. (a) MCQ Prompt 1 directs the model to choose one answer without explanation;
(b) EMI Prompt 2 blends scenario-based extended matching items (EMIs) with standard MCQs, requiring only the answer letters.}
\label{fig:Multiple_choice_question}
\begin{tikzpicture}
\node[fill=gray!8, draw=gray, rounded corners, inner sep=5pt, text width=1\textwidth] (box) {
\begin{verbatim}
Answer the question using one of the given choices.Do not provide any explanation for your choice. 
Indicate your choice with a single English letter.Provide only one answer; Multiple answers could be provided
The Patient’s History:{History}
Question:{Question}
Choices:{Choices}
\end{verbatim}
        };
\end{tikzpicture}
\vspace{0.5cm}
\textbf{MCQ Prompt 1}
\begin{tikzpicture}
\node[fill=gray!8, draw=gray, rounded corners, inner sep=5pt, text width=1\textwidth] (box) {
\begin{verbatim}
Analyze the multiple-choice question (MCQ) and select the best answer.Assess each option logically based on 
relevant psychological history.Do not provide any explanation for your choice. Indicate your choice with 
a single English letter. (Provide only one answer; Multiple answers could be provided.)
The Patient’s History:{History}
Question:{Question}
Choices:{Choices}
\end{verbatim}
};
\end{tikzpicture}
\vspace{0.5cm}
\textbf{EMI Prompt 1}
\end{figure}

\begin{figure}[p]
  \centering
    \caption{Point-based similarity EMI prompting templates: (a) a straightforward "header-required" template that simply presents the fields.}

  \label{fig:point_based_similarity_prompt}

  \begin{subfigure}{\textwidth}
    \centering
    \begin{adjustbox}{scale=1} 
      \begin{tikzpicture}
        \node[
          fill=gray!8,
          draw=gray,
          rounded corners,
          inner sep=5pt,
          text width=\textwidth
        ] {
\begin{verbatim}
Answer the question using one of the given choices.Do not provide any explanation for your choice. Indicate
your choice with a single English letter. (Provide only one answer; Multiple answers could be provided.)
Header:{Header}
Required:{Required}
Question:{Question}
Choices:{Choices}
\end{verbatim}
        };
      \end{tikzpicture}
    \end{adjustbox}
    \caption{EMI Separated Prompt 1}
    \label{fig:emi_separated1}
  \end{subfigure}


\end{figure}

\subsection {Classification Mapping}
\label{subsec:Class}
This subsection details the standardized diagnostic taxonomy employed for the classification task, which was explicitly derived from the DSM-5-TR to ensure terminological precision. To facilitate rigorous multi-label evaluation, two distinct reference lists were curated and validated by a licensed psychiatrist : a granular list of Specific Disorders ~\ref{fig:Specific_Disorders_List}, encompassing detailed diagnoses such as "Bipolar I Disorder," and a high-level list of Disorder Categories ~\ref{fig:Categories_Disorder_List}, which groups conditions into broader classes like "Depressive Disorders." These mappings serve as the ground truth for evaluation, requiring models to align their outputs with authoritative psychiatric nosology rather than relying on ambiguous or unstructured free-text descriptions
\begin{figure}[!ht]
    \centering
    \caption{\textbf{Specific Disorders List}}
    \label{fig:Specific_Disorders_List}
    \begin{adjustbox}{scale=1} 
        \begin{tikzpicture}
            \node[fill=gray!8, draw=gray, rounded corners, inner sep=6pt, text width=1.02\textwidth] (box) {
\begin{verbatim}
[ "Intellectual Developmental Disorder (Intellectual disability)",
"Global Developmental Delay", "Unspecified Intellectual Developmental Disorder",
"Language Disorder", "Speech Sound Disorder", 
"Childhood-Onset Fluency Disorder (Stuttering)", 
"Social (Pragmatic) Communication Disorder", "Unspecified Communication Disorder",
"Autism Spectrum Disorder", "Attention-Deficit/Hyperactivity Disorder",
"Other Specified Attention-Deficit/Hyperactivity Disorder",
"Unspecified Attention-Deficit/Hyperactivity Disorder","Specific Learning Disorder",
"Developmental Coordination Disorder", "Stereotypic Movement Disorder",
"Substance-Induced Repetitive Behaviors", "Tourette’s Disorder",
"Persistent (Chronic) Motor or Vocal Tic Disorder", "Provisional Tic Disorder",
"Other specified Tic disorder", "Unspecified Tic Disorder",
"Other specified Neurodevelopmental disorders", 
"Unspecified neurodevelopmental disorders","Delusional Disorder", 
"Brief Psychotic Disorder", 
"Schizophreniform Disorder", "Schizophrenia", "Schizoaffective Disorder", 
"Substance/Medication-Induced Psychotic Disorder",
"Psychotic disorder due to another medical disorders",
"Catatonia associated with another mental disorder",
"Catatonia associated with another medical disorder", "Unspecified Catatonia",
"Other Specified Schizophrenia Spectrum and Other Psychotic Disorder",
"Unspecified Schizophrenia Spectrum and Other Psychotic Disorder", "Bipolar I Disorder", 
"Bipolar II Disorder", "Cyclothymic Disorder",
"Substance/Medication-Induced Bipolar and Related Disorder",
"Bipolar and Related Disorder Due to Another Medical disorders",
"Other Specified Bipolar and Related Disorder", 
"Unspecified Bipolar and Related Disorder", 
"Unspecified Mood Disorder", "Disruptive Mood Dysregulation Disorder", 
"Major Depressive Disorder", "Persistent Depressive Disorder", 
"Premenstrual Dysphoric Disorder", "Substance/Medication-Induced Depressive Disorder",
"Depressive Disorder Due to Another Medical disorders",
"Other Specified Depressive Disorder", "Unspecified Depressive Disorder",
"Separation Anxiety Disorder", "Selective Mutism", "Specific Phobia",
"Social Anxiety Disorder", 
"Panic Disorder", "Agoraphobia","Generalized Anxiety Disorder", 
"Substance/Medication-Induced Anxiety Disorder",
"Anxiety Disorder Due to Another Medical disorders", "Other Specified Anxiety Disorder", 
"Unspecified Anxiety Disorder", "Obsessive-Compulsive Disorder",
"Body Dysmorphic Disorder", 
"Hoarding Disorder","Trichotillomania (Hair-Pulling Disorder)",
"Excoriation (Skin-Picking) Disorder",
"Substance/Medication-Induced Obsessive-Compulsive and Related Disorder",
"Obsessive-Compulsive and Related Disorder Due to Another Medical disorders",
"Other Specified Obsessive-Compulsive and Related Disorder",
"Unspecified Obsessive-Compulsive and Related Disorder", "Reactive Attachment Disorder", 
"Disinhibited Social Engagement Disorder", "Posttraumatic Stress Disorder", 
"Acute Stress Disorder", "Adjustment Disorders", "Prolonged Grief Disorder", 
"Other Specified Trauma- and Stressor-Related Disorder",
"Unspecified Trauma- and Stressor-Related Disorder", "Dissociative Identity Disorder",
"Dissociative Amnesia", "Depersonalization/Derealization Disorder",
"Other Specified Dissociative Disorder", "Unspecified Dissociative Disorder",
"Somatic Symptom Disorder", "Illness Anxiety Disorder",
"Functional Neurological Symptom Disorder (Conversion Disorder)",
"Psychological Factors Affecting Other Medical disorders", "Factitious Disorder",
"Other Specified Somatic Symptom and Related Disorders",
"Unspecified Somatic Symptom and Related Disorder", "Pica", "Rumination Disorder",
\end{verbatim}
            };
        \end{tikzpicture}
    \end{adjustbox}
    \vspace{0.5cm}
    \
    % \textbf{Evaluation Prompt}
    % \caption{}
\end{figure}
\begin{figure}[!ht]
    \centering
    \begin{adjustbox}{scale=1} % Adjust scale as needed
        \begin{tikzpicture}
            \node[fill=gray!8, draw=gray, rounded corners, inner sep=6pt, text width=1\textwidth] (box) {
\begin{verbatim}
"Avoidant/Restrictive Food Intake Disorder", "Anorexia Nervosa", "Bulimia Nervosa",
"Binge-eating disorder", "Other Specified Feeding or eating disorder",
"Unspecified Feeding or eating disorder", "Enuresis", "Encopresis",
"Other Specified Elimination Disorder", "Unspecified Elimination Disorder",
"Insomnia Disorder", "Hypersomnolence Disorder", "Narcolepsy",
"Obstructive Sleep Apnea Hypopnea", "Central Sleep Apnea",
"Sleep-Related Hypoventilation", "Delayed Sleep Phase Type",
"Advanced Sleep Phase Type", "Irregular Sleep-Wake Type",
"Non-24-Hour Sleep-Wake Type", "Shift Work Type", "Parasomnias",
"Non-Rapid Eye Movement Sleep Arousal Disorders", "Nightmare Disorder",
"Rapid Eye Movement Sleep Behavior Disorder", "Restless Legs Syndrome",
"Substance/Medication-Induced Sleep Disorder", "Other Specified Insomnia Disorder",
"Unspecified Insomnia Disorder", "Other Specified Hypersomnolence Disorder",
"Unspecified Hypersomnolence Disorder", "Other Specified Sleep-Wake Disorder",
"Unspecified Sleep-Wake Disorder", "Delayed Ejaculation", "Erectile Disorder",
"Female Orgasmic Disorder", "Female Sexual Interest/Arousal Disorder",
"Genito-Pelvic Pain/Penetration Disorder", "Male Hypoactive Sexual Desire Disorder",
"Premature (Early) Ejaculation", "Substance/Medication-Induced Sexual Dysfunction",
"Other Specified Sexual Dysfunction", "Unspecified Sexual Dysfunction",
"Other Specified Gender Dysphoria", "Unspecified Gender Dysphoria",
"Oppositional Defiant Disorder","Intermittent Explosive Disorder","Conduct Disorder",
"Pyromania", "Kleptomania",
"Other Specified Disruptive, Impulse-Control, and Conduct Disorder",
"Unspecified Disruptive, Impulse-Control, and Conduct Disorder",
"Alcohol Use Disorder", "Alcohol Intoxication", "Alcohol Withdrawal",
"Alcohol-induced Mental disorders", "Unspecified Alcohol-Related Disorder",
"Caffeine Intoxication", "Caffeine Withdrawal", "Caffeine-Induced Mental Disorders",
"Unspecified Caffeine-Related Disorder", "Cannabis Use Disorder",
"Cannabis Intoxication", "Cannabis Withdrawal", "Cannabis-Induced Mental Disorders",
"Unspecified Cannabis-Related Disorder", "Phencyclidine Use Disorder",
"Other Hallucinogen Use Disorder", "Phencyclidine Intoxication",
"Other Hallucinogen Intoxication", "Hallucinogen Persisting Perception Disorder",
"Phencyclidine-Induced Mental Disorders", "Hallucinogen-Induced Mental Disorders",
"Unspecified Phencyclidine-Related Disorder",
"Unspecified Hallucinogen-Related Disorder", "Inhalant Use Disorder",
"Inhalant Intoxication", "Inhalant-Induced Mental Disorders",
"Unspecified Inhalant-Related Disorder","Opioid Use Disorder","Opioid Intoxication",
"Opioid Withdrawal", "Opioid-Induced Mental Disorders",
"Unspecified Opioid-Related Disorder",
"Sedative, Hypnotic, or Anxiolytic Use Disorder",
"Sedative, Hypnotic, or Anxiolytic Intoxication",
"Sedative, Hypnotic, or Anxiolytic Withdrawal",
"Sedative-, Hypnotic-, or Anxiolytic-Induced Mental Disorders",
"Unspecified Sedative-, Hypnotic-, or Anxiolytic- ]
\end{verbatim}
            };
        \end{tikzpicture}
    \end{adjustbox}
    \vspace{0.5cm}
    
\end{figure}

\begin{figure}[!ht]
\caption{\textbf{Disorder Categories List}}
    \label{fig:Categories_Disorder_List}

    \centering
    \begin{adjustbox}{scale=1} 
        \begin{tikzpicture}
            \node[fill=gray!8, draw=gray, rounded corners, inner sep=6pt, text width=1\textwidth] (box) {
\begin{verbatim}
["Intellectual Developmental Disorders", "Communication Disorders", "Autism Spectrum Disorder",
"Attention-Deficit/Hyperactivity Disorder", "Specific Learning Disorder", "Motor Disorders", 
"Tic Disorders","Schizophrenia Spectrum and Other Psychotic Disorders", "Bipolar and Related Disorders", 
"Depressive Disorders", "Anxiety Disorders","Obsessive-Compulsive and Related Disorders", 
"Trauma- and Stressor-Related Disorders", "Dissociative Disorders", "Somatic Symptom and Related Disorders", 
"Feeding and Eating Disorders", "Elimination Disorders", "Sleep-Wake Disorders", "Parasomnias",
"Sexual Dysfunctions","Breathing-Related Sleep Disorders", "Gender Dysphoria", "Disruptive, 
Impulse-Control, and Conduct Disorders", "Substance-Related and Addictive Disorders",
"Neurocognitive Disorders", "Personality Disorders", "Medication-Induced Movement Disorders", 
"Paraphilic Disorders"]
\end{verbatim}
            };
        \end{tikzpicture}
    \end{adjustbox}
    \vspace{0.5cm}
    
\end{figure}

\begin{figure}[!ht]
    \centering
    \caption{Prompt generated by GPT-4.5 for point-based similarity scoring of clinical reasoning responses.}

    \begin{adjustbox}{scale=1} 
        \begin{tikzpicture}
            \node[fill=gray!8, draw=gray, rounded corners, inner sep=6pt, text width=1\textwidth] (box) {
\begin{verbatim}
I will provide two texts: Text A (LLM’s response) and Text B (True answer). Evaluate their similarity based 
strictly on the presence and accuracy of key diagnoses, differential diagnoses, and clinical reasoning. 
Follow these guidelines:
    • Assign a similarity score between 0 and 100:
        - 0: Text A does not contain the correct diagnoses or significantly misrepresents key clinical reasoning 
        compared to Text B.
        - 50: Text A identifies some key diagnoses but either misses important differentials or partially 
        misrepresents clinical reasoning compared to Text B.
        - 100 means Text A correctly captures all essential diagnoses, 
        differential diagnoses,and clinical reasoning as stated in Text B.
    • Use the following criteria to guide your scoring:
        - Presence of the correct main diagnosis (40 points).
        - Inclusion and accuracy of differential diagnoses (30 points).
        - Accuracy and completeness of clinical reasoning and justification (30 points).
    • Provide ONLY the numeric similarity score (0-100) without any additional explanation or justification.
\end{verbatim}
            };
        \end{tikzpicture}
    \end{adjustbox}
    \vspace{0.5cm}
    \label{fig:Prompt-generated-GPT-4.5}
\end{figure}

\begin{figure}[!ht]
    \centering
    \caption{Prompt generated by Gemini for Llama 3 70B evaluation.}
    \label{fig:generated-Gemini-for-Llama-evaluation.}
    \begin{adjustbox}{scale=0.85} 
        \begin{tikzpicture}
            \node[fill=gray!8, draw=gray, rounded corners, inner sep=6pt, text width=1\textwidth] (box) {
\begin{verbatim}
SYSTEM: You are a medical diagnosis evaluator. Your task is to compare two medical text summaries and assess 
the degree to which the first text (the "Candidate Answer") captures the essential diagnostic information 
contained in the second text (the "Reference Answer"). Focus on the core medical meaning and the completeness 
of the diagnoses, not on the writing style or specific wording.
USER: 
INSTRUCTIONS:
1. Read the "Reference Answer" carefully. Identify the key diagnoses (both primary and differential) and any 
crucial supporting information (e.g., symptoms, test results, ruled-out disorders).
2. Read the "Candidate Answer". The "Candidate Answer" will be a concise list of diagnoses and key findings.
3. Compare the "Candidate Answer" to the "Reference Answer" based on the following criteria:
    • Diagnostic Accuracy: Does the "Candidate Answer" correctly identify the most likely diagnosis 
    (if one is clearly indicated in the "Reference Answer")?
    • Diagnostic Completeness: Does the "Candidate Answer" include all of the major differential diagnoses 
    mentioned in the "Reference Answer"? It is acceptable if the "Candidate Answer" uses slightly different 
    terminology as long as the underlying medical concept is the same.
    • Supporting Information: Does the "Candidate Answer" include the key symptoms, test results, 
    or other evidence that support the diagnoses? It doesn’t need to include every detail, but it should capture 
    the most important ones.It’s ok if this supporting information is implicitly present in the reference answer
    • Exclusion of Irrelevant Information: 
    Do not penalize the "Candidate Answer" for including additional 
    information that is not present in the "Reference Answer", as long as that information is medically accurate 
    and relevant to the case.
    • groundance does not affect the score:Do not penalize a response for being redundant
4. Provide a single numerical similarity score between 0 and 100, where:
    • 0: The "Candidate Answer" has no meaningful overlap with the "Reference Answer" in terms of diagnosis or 
    supporting information. It’s completely wrong.
    • 50: The "Candidate Answer" identifies some relevant diagnoses or supporting information, but it misses 
    major elements or includes significant inaccuracies.
    • 75: The "Candidate Answer" correctly identifies the primary diagnosis 
    (if applicable) and most of the important differential diagnoses and supporting information. It may miss 
    some minor details or include some slight inaccuracies.
    • 90: The "Candidate Answer" is very close to the "Reference Answer" in terms of core diagnostic meaning and 
    completeness. It may have minor omissions or differences in emphasis, but it captures almost all of 
    the essential information.
    • 100: The "Candidate Answer" is a perfect match in terms of core diagnostic meaning. It includes all of 
    the key diagnoses and supporting information from the "Reference Answer",even if the wording 
    or organization is different.
5. Do not justify or explain the answer. Give a score between 0 and 100, no other text.
Here is an example for better understanding:
Reference Answer:
Differential diagnosis: 1. Delirium 2. Substance intoxication 3.Substance withdrawal
delirium 4. Mental retardation
Candidate Answer:
- Primary Diagnosis: Alzheimer’s_Disease
- Key Findings: Progressive memory loss, disorientation, difficulty with ADLs.
- Differential Diagnoses: Vascular_Dementia, 
Lewy_Body_Dementia, Frontotemporal_Dementia
ASSISTANT: 75
Here is another example:
Reference Answer:
Alcohol Use Disorder (Severe)
Candidate Answer:
- Primary Diagnosis: Alcohol_Use_Disorder (Mild)
- Key Findings: History of heavy alcohol use, withdrawal_symptoms.
ASSISTANT: 50
Now, evaluate the following: 
Reference Answer : {Answer}
Candidate Response: {Response}
            \end{verbatim}
            };
        \end{tikzpicture}
    \end{adjustbox}
    \vspace{0.5cm}
    
\end{figure}

\end{document}